\documentclass[sn-mathphys,Numbered]{sn-jnl-arxiv}

\usepackage{natbib}
\usepackage{graphicx}%
\usepackage{multirow}%
\usepackage{amsmath,amssymb,amsfonts}%
\usepackage{amsthm}%
\usepackage{mathrsfs}%
\usepackage{xcolor}%
\usepackage{textcomp}%
\usepackage{manyfoot}%
\usepackage{booktabs}%
\usepackage{algorithm}%
\usepackage{algorithmicx}%
\usepackage{algpseudocode}%
\usepackage{listings}%
\usepackage[nobottomtitles]{titlesec}

\usepackage[ampersand]{easylist}
\ListProperties(Hide2=1,Hide3=2,Progressive*=.5cm,Numbers3=l, Numbers4=r,FinalMark3={)})
\usepackage{etoolbox}
\AtBeginEnvironment{easylist}{\ListProperties(Start1=1)}

\usepackage{xcolor}
\newcommand{\blank}{\underline{\hspace{15pt}}}

\definecolor{cblue}{RGB}{8, 85, 153}

\usepackage[capitalise]{cleveref}
\usepackage{float}
\usepackage{placeins} 
\usepackage{adjustbox}

\usepackage{makecell}
\usepackage[ampersand]{easylist}
\usepackage{todonotes}
\usepackage{caption}

\newcommand{\methodlong}{generative causal testing}
\newcommand{\method}{GCT}
\newcommand{\methodunderlined}{\underline{g}enerative \underline{c}ausal \underline{t}esting}

\newcommand{\add}[1]{{#1}}

\usepackage{adjustbox}

\usepackage{titlesec}
\titleformat{\section}
  {\normalfont\large\bfseries}
  {\thesection}{1em}{}
\begin{document}

\title[Article Title]{Generative causal testing to bridge data-driven models and scientific theories in language neuroscience}


\author[1]{\fnm{Richard J.} \sur{Antonello}}\email{rjantonello@utexas.edu}
\equalcont{These authors contributed equally to this work.}

\author[2]{\fnm{Chandan} \sur{Singh}}\email{chansingh@microsoft.com}
\equalcont{These authors contributed equally to this work.}

\author[3,a]{\fnm{Shailee} \sur{Jain}}\email{shailee.jain@ucsf.edu}

\author[2,4]{\fnm{Aliyah} \sur{Hsu}}\email{aliyahhsu@berkeley.edu}

\author[7]{\fnm{Sihang} \sur{Guo}}\email{sunaguo@berkeley.edu}

\author[2]{\fnm{Jianfeng} \sur{Gao}}\email{jfgao@microsoft.com}

\author[4,5,6]{\fnm{Bin} \sur{Yu}}\email{binyu@berkeley.edu}
\equalcontsup{These authors jointly supervised this work.}

\author[1,5,7]{\fnm{Alexander G.} \sur{Huth}}\email{alex.huth@berkeley.edu}
\equalcontsup{These authors jointly supervised this work.}

\affil[1]{\orgdiv{Computer Science Department}, \orgname{University of Texas at Austin,} \state{TX,} \country{USA}}
\affil[2]{\orgname{Microsoft Research}, \city{Redmond,} \state{WA,} \country{USA}}
\affil[3]{\orgdiv{Neurosurgery Department}, \orgname{University of California}, \orgaddress{\city{San Francisco,} \state{CA,} \country{USA}}}
\affil[4]{\orgdiv{EECS Department}, \orgname{University of California}, \orgaddress{\city{Berkeley,} \state{CA,} \country{USA}}}
\affil[5]{\orgdiv{Statistics Department}, \orgname{University of California}, \orgaddress{\city{Berkeley,} \state{CA,} \country{USA}}}
\affil[6]{\orgdiv{Center for Computational Biology}, \orgname{University of California}, \orgaddress{\city{Berkeley,} \state{CA,} \country{USA}}}
\affil[7]{\orgdiv{Neuroscience Department}, \orgname{University of California}, \orgaddress{\city{Berkeley,} \state{CA,} \country{USA}}}
\affil[a]{Work done while at UT Austin}


\abstract{Representations from large language models (LLMs) are highly effective at predicting BOLD fMRI responses to language stimuli.
However, these representations are largely opaque: it is unclear what features of the language stimulus drive the response in each brain area.
We present \methodunderlined~(\method{}), a methodological framework for generating concise natural language explanations of language selectivity in the brain from predictive models and then testing those explanations in follow-up experiments using LLM-generated stimuli.
This approach is successful at explaining selectivity both in individual voxels and cortical regions of interest (ROIs), including newly identified micro ROIs in prefrontal cortex.
We show that explanatory accuracy is closely related to the predictive power and stability of the underlying predictive models.
Finally, we show that \method{} can dissect fine-grained differences between brain areas with similar functional selectivity.
These results demonstrate that LLMs can be used to bridge the widening gap between data-driven models and formal scientific hypotheses.
}
\keywords{Language models, Encoding models, fMRI, Synthetic data}


\maketitle

\section{Introduction}
\label{sec:intro}

Science faces an explainability crisis: data-driven deep learning methods are proving capable of modeling many natural phenomena, like protein folding, meteorological events, and computations in the brain~\cite{abramson2024accurate, kochkov2024neural, yamins2016using}. However, these models are not scientific theories that describe the world in natural language. Instead, they are implemented in the form of vast neural networks with millions or billions of largely inscrutable parameters. One emblematic field is language neuroscience, where large language models (LLMs) are highly effective at predicting human brain responses to natural language, but are virtually impossible to interpret or analyze by hand~\cite{abnar2019, goldstein2022shared, vaidya2022self, jain2023computational, antonello2023scaling, NEURIPS2023_3a0e2de2}. 

To overcome this challenge, we introduce the \methodunderlined~(\method) framework.
\method{} translates predictive deep learning models of language selectivity in the brain into concise verbal explanations, and then designs follow-up experiments using LLM-generated stimuli to evaluate the explanations. Our primary contribution is an end-to-end framework for hypothesis construction and experimental evaluation in language neuroscience. Starting from an uninterpretable encoding model of the brain, \method{} produces concise, targeted verbal hypotheses about semantic selectivity and then generates controlled stimulus interventions to evaluate them.
This process synthesizes an emerging line of work that provides data-driven descriptions of brain regions but does not validate those descriptions in follow-up experiments~\cite{JAINNEURIPS2020, chen2023cortical, Vo2023.08.03.551886,luo2024brainscuba,singh2023explaining} with works that build stimuli to drive brain activity without assigning interpretations~\cite{tuckute2023driving,bashivan2019neural,abbasi2018deeptune, murty2021,ponce2019evolving,walker2019inception}.
Combining the strengths of these methods, our approach enables identifying data-driven explanations that are \textit{causally} related to brain activity, i.e., the explanation prescribes a stimulus intervention that elicits specific changes in brain activity. Throughout, we use the term ``explanation'' to mean a compact, human-interpretable hypothesis that supports a targeted stimulus intervention and yields a measurable change in response.
Such explanations can be \emph{partial} characterizations of selectivity where the success of one explanation provides evidence that the corresponding feature is causally relevant, without necessarily implying that the explanation is monosemantic, i.e. that no other feature could also drive the same voxel or ROI (see Appendix~\ref{sec:appendix_polysemantic} for discussion and analysis of polysemanticity).

\begin{figure}[ht]
\centering
\makebox[\textwidth]{
\includegraphics[width=1.2\textwidth]{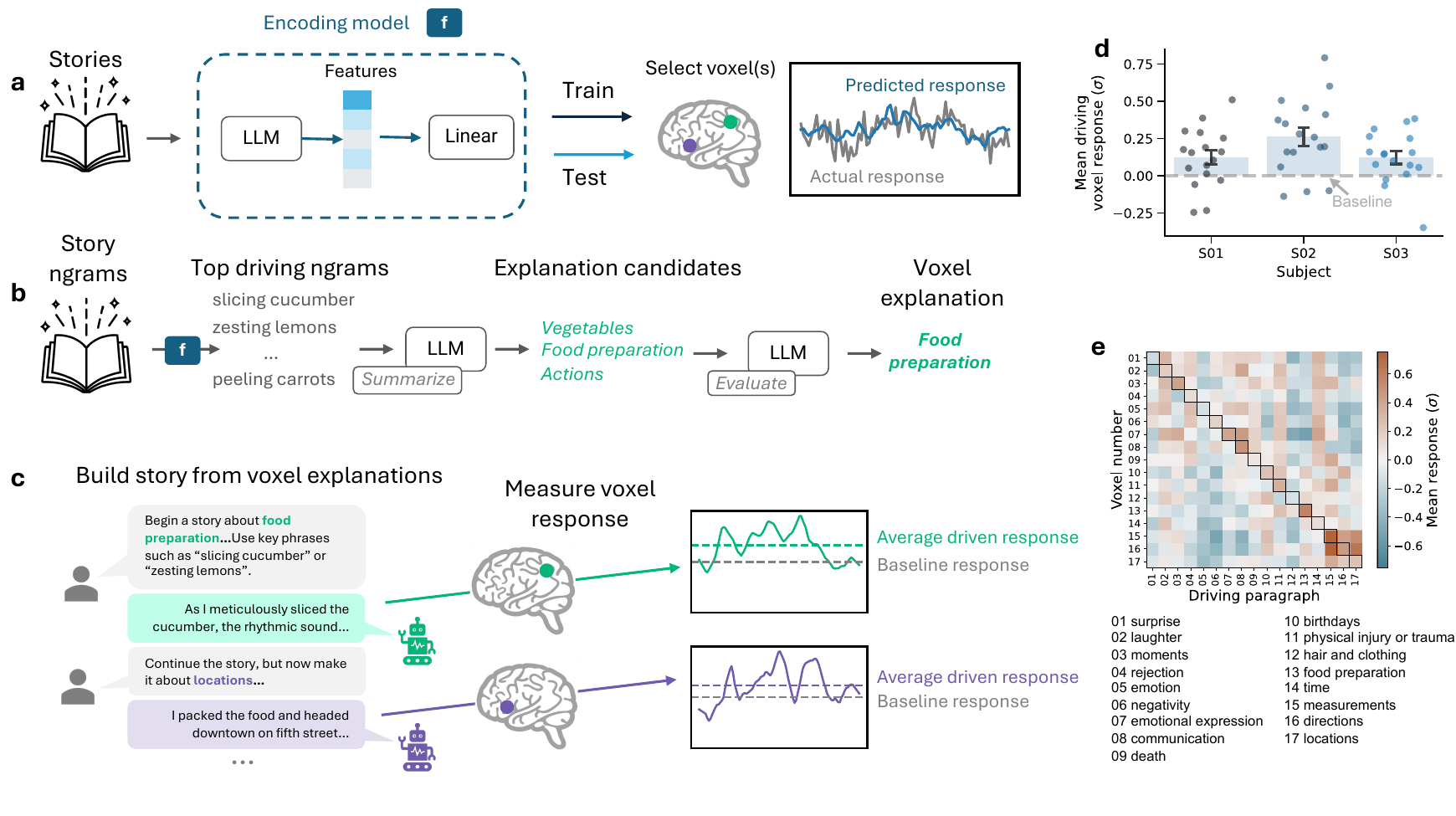}
}
    \caption{\textbf{Driving single-voxel response with \methodlong.}
        (a) Voxelwise BOLD responses were recorded using fMRI as human subjects listened to 20 hours of narrative stories. An encoding model $f$ was fit to predict these responses from the story text. $f$ consists of a linear model fit on representations extracted from an LLM, which are not readily interpretable. Encoding models were tested by predicting responses on held-out fMRI data, and only well-performing models were selected for further analysis.
        (b) We used an automated procedure to find a verbal description of the function that $f$ computes for each voxel. First, we tested $f$ on a large catalog of $n$-grams ($n=1,2,3)$ and found those that maximally drove predicted responses. These $n$-grams were then summarized into stable explanation candidates using a powerful instruction-tuned LLM. Finally, we evaluated each explanation candidate by generating corresponding synthetic sentences and testing that these sentences yielded large predictions from $f$. 
        (c) To test whether the generated explanations were causally related to activation in the brain, we used an LLM to produce synthetic narrative stories where each paragraph is designed to drive responses based on the generated explanation for one voxel.
        For each subject we constructed stories to drive 17 well-modeled voxels with diverse selectivity. These stories were then presented to the subjects in a second fMRI experiment.
        (d) Difference between each voxel's average BOLD response during its driving paragraph versus during its baseline, i.e. average response to all non-driving paragraphs.
        Each point shows a single voxel.
        On average, driven responses were significantly higher than baseline for each subject ($p=0.020$ (S01), $p<10^{-5}$ (S02), $p=0.009$ (S03); permutation test, FDR-corrected).
        For well-driven voxels, this means that the generated explanation is causally related to activation of that voxel, and thus that we have successfully translated the LLM-based encoding model into a verbal explanation.
        (e) Average BOLD response for each selected voxel to each of the driving paragraphs in one subject (S02). Responses to the driving paragraph generated using the explanation for that voxel appear along the main diagonal. Explanations that were used to construct the driving paragraphs are shown below. BOLD responses were generally high for the driving paragraphs for each voxel as well as semantically related paragraphs (e.g. \textit{directions} and \textit{locations}, \textit{emotional expression} and \textit{laughter}).
    }
\label{fig:fig1}
\end{figure}

\section{Language encoding models}\label{sec:language_encoding_models}
We designed \method{} to interpret \textit{language encoding models}, which are data-driven, LLM-based models of cortical language selectivity. We thus began by fitting encoding models (\cref{fig:fig1}a) for each voxel in each of 3 human subjects using 20 hours of passive language listening fMRI data~\cite{lebel2022natural}. The 20 hours of speech stimuli were transcribed and then passed through open-source large language models~\cite{zhang2022opt, touvron2023llama} to obtain activation vectors for each word. Multiple models were built using different LLMs to ensure that the generated predictions are stable~\cite{yu2020veridical,abbasi2018deeptune}, i.e. not idiosyncratic to our choice of LLM. Regularized linear regression was then used to predict the response timecourse of each voxel as a weighted combination of LLM activations.
Encoding models were tested by predicting responses on held out fMRI data and then computing the correlation between predicted and actual responses.
Encoding models using advanced LLMs achieve very high prediction performance~\cite{antonello2023scaling}; here, only models with sufficient prediction performance ($r > 0.15$) were used in further analyses, with many voxels predicted at $r > 0.7$.
Full preprocessing, modeling, and data collection details are provided in the \hyperref[sec:methods]{Methods}.

\section{Generating and testing natural-language explanations for single voxels}\label{sec:results}

Each encoding model represents the language selectivity of one part of the brain, but does so as a linear combination of LLM activations that are not human-interpretable.
The first step in \method{} is to convert these models into concise natural language explanations: a word or short phrase summarizing the properties of language that drive responses most strongly.
Because voxel responses can reflect multiple semantic dimensions, a single phrase should be interpreted as an operational hypothesis: a direction in stimulus space that reliably modulates the response.
Accordingly, \method{} is designed to identify driving semantic factors that can be tested experimentally, rather than to produce a complete decomposition of selectivity.

We accomplish this by using an instruction-finetuned LLM~\cite{openai2023gpt4system} to generate candidate explanations by summarizing the \textit{n}-grams that yield the largest predictions from the model (\cref{fig:fig1}b); see details in the \hyperref[sec:methods]{Methods}. 
The resulting explanation is given in easily understandable natural language. For example, we found that some voxels in the visual ventral stream appeared to be selective for language describing \textit{Food preparation}, supporting prior work \cite{khosla2022highly, jain2023selectivity} suggesting the presence of a food-selective region (\cref{fig:fig1}b).
To ensure that the generated explanation correctly captures the function computed by the encoding model, we used the same LLM to generate synthetic text from the explanation and checked that this text could successfully drive the voxel encoding model.
However, this does not guarantee that this function is correctly aligned to the activity of the brain in the real world.


To fully close the loop and confirm that a generated explanation accurately reflects selectivity in the brain, the second step of \method{} is to automatically design a new neuroimaging experiment to test the generated explanation \textit{in vivo}.
This was done by prompting an instruction-finetuned LLM~\cite{openai2023gpt4system} to generate narratives that should selectively drive cortical activation based on that explanation~(\cref{fig:fig1}c).
If the explanation generated for a voxel is accurate, then text generated according to that explanation should elicit large responses in that voxel; 
this would demonstrate that the explanation has causal influence over that voxel's activity. 
Given a selected set of voxels and their explanations, we constructed narrative stories by iteratively prompting an LLM~\cite{openai2023gpt4system} to prioritize a different voxel's explanation as the main focus of each paragraph. This allowed us to test whether voxels selectively respond to paragraphs that match their explanation.
The LLM ensures these stories remain coherent and engaging, helping to keep subjects attentive during the fMRI experiment~\cite{hamilton2020revolution, jain2023computational}.


We used \method{} to generate and test explanations for 17 voxels with strong encoding model predictive performance in each of 3 subjects. The number of voxels was selected so that resulting narratives would be similar in length to the stimuli used to initially fit the encoding models~\cite{lebel2022natural}.
We then measured the average response of each target voxel during the paragraph that was designed to drive it and compared this to the average response to other paragraphs (\cref{fig:fig1}d).
Across voxels, responses to driving paragraphs were significantly greater than baseline responses for each subject ($p=0.020$ (S01), $p<10^{-5}$ (S02), $p=0.009$ (S03); permutation test with Benjamini-Hochberg false discovery rate correction~\cite{benjamini1995controlling} applied for each subject).
Of the 51 tested voxels across the 3 subjects, 41 show increased response, with an average increase of 0.198 standard deviations per TR over baseline when responses are averaged temporally across driving paragraphs. This contrasts with a computed ``driving ceiling'' for most voxels that typically ranges from 0.4 sigma to 0.7 sigma~(\cref{tab:noise_ceilings}). 
Differences in effect size are related to head motion of subjects during test time and variation in sample size~(\cref{tab:head_motion}).
This demonstrates that the generated explanations are mostly effective drivers of brain activity for their chosen voxels.
Further, because this experiment was not based directly on the LLM encoding model~\cite{tuckute2023driving} but was designed using the generated explanation, this suggests that we have successfully found explanations that are causally linked to brain activity.

The results in \cref{fig:fig1}d compare responses during the driving paragraphs generated for each voxel to average responses over all other paragraphs. But if two voxels have very similar explanations and both are correct, then we would expect each to also be driven by the other's generated stimuli. To explore this question we disaggregated the results by showing the response of each voxel to every driving paragraph in one subject (\cref{fig:fig1}e). Most voxels are driven by their own explanations, as can be seen by the positive values on the main diagonal. However, many voxels are also strongly driven by related explanations (e.g. \textit{directions}, \textit{measurements}, and \textit{locations}; \textit{communication} and \textit{emotional expression}). This demonstrates that, in most cases, semantically related explanations will drive similar sets of voxels. Off-diagonal structure is expected when explanations share semantic content or when voxels are tuned to overlapping feature sets; in this sense, the matrix visualizes both selectivity and shared representational structure.
The off-diagonal structure of \cref{fig:fig1}e shows that our results should therefore not be interpreted as a strict one-to-one ``labeling'' of voxels, but as experimentally grounded hypotheses that predict when responses should increase under controlled semantic manipulations. 
We additionally find that the same setting succeeds in driving pairs of voxels rather than individual voxels~(\cref{subsec:multivoxel}), and that our use of a written stimulus for driving as opposed to audio did not meaningfully impact driving performance~(\cref{fig:audiodriving}). Furthermore, we examined several other orthogonal methods for driving, including a representation steering approach and a synthetic decoding approach. Full details of these attempts can be found in \cref{sec:driving_vars}. In all cases, we found these alternative methods less effective than \method{}.

\begin{figure}[t]
    \centering
    \makebox[\textwidth]{
    \includegraphics[width=1.2\textwidth]{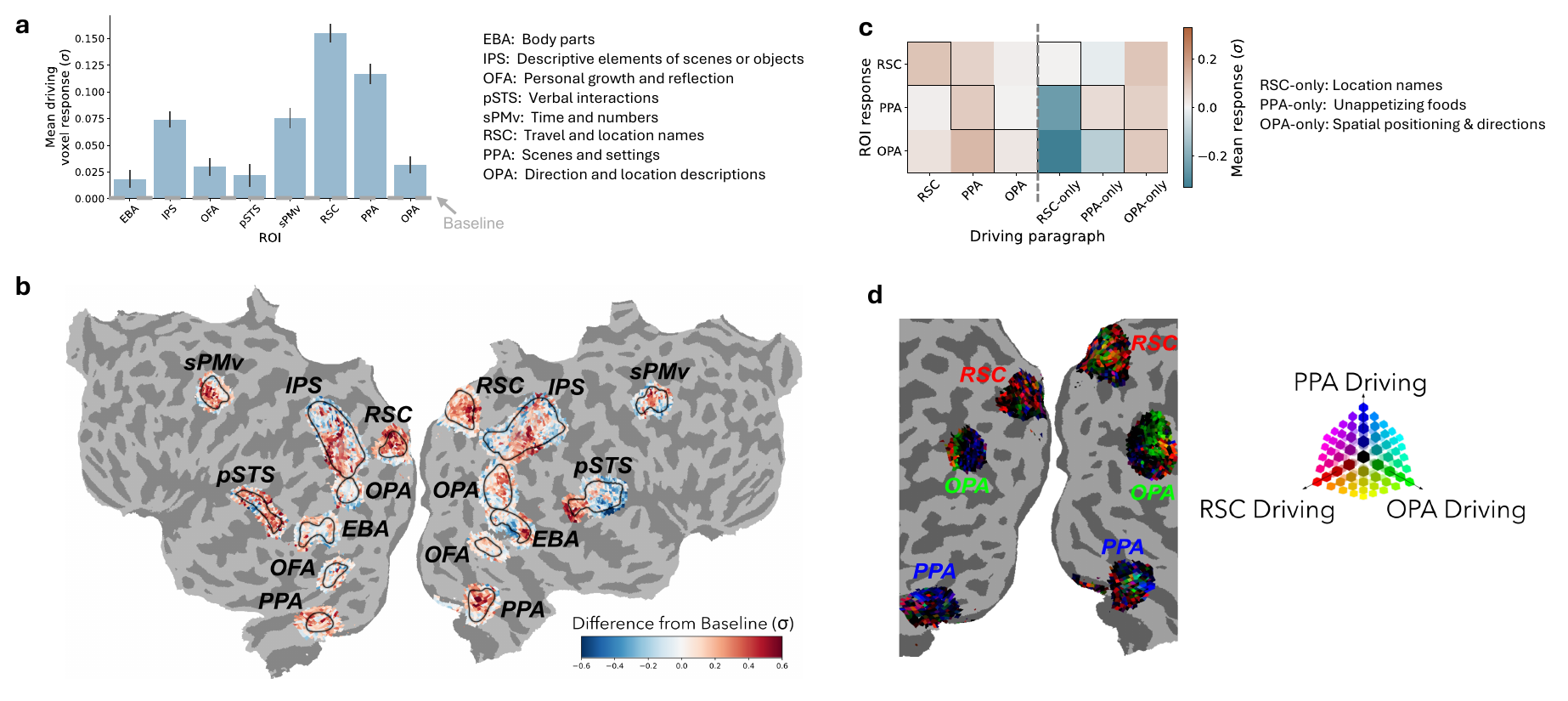}
    }
    \caption{
    \textbf{Driving ROI response with generative causal testing for subject S02.
    }
    (a) Explanations were generated and used to drive 8 well-defined regions of interest.
    Responses in all ROIs were significantly driven above baseline ($p<0.05$; permutation test, FDR-corrected). 
    (b) To understand driving performance with more granularity, we color each voxel in each ROI by how well it was driven by its corresponding driving paragraph.
    The resulting composite flatmap occasionally shows subregions within ROIs that are more selectively driven for a particular explanation.
    (c) \method{} can also be used to build more nuanced theories of cortical semantic selectivity.
    We focused on three ROIs that are known to have similar selectivity for place concepts: retrosplenial cortex (RSC), the parahippocampal place area (PPA), and the occipital place area (OPA).
    When explanations were generated for each ROI independently we found that each ROI was driven by all three driving paragraphs (left side).
    To distinguish these ROIs, we used \method{} to find new explanations and construct stories that would selectively drive each area while suppressing the other two.
    Testing these stories in an fMRI experiment showed that we succeeded in finding selective explanations for two ROIs: RSC is selectively driven by \textit{location names} ($p < 10^{-3}$; permutation test) and PPA by \textit{unappetizing foods} ($p < 10^{-3}$; permutation test). However, the explanation for OPA, \textit{spatial positioning \& directions}, drove responses in all three ROIs (right side).
    (d) Visualization of the place area driving experiment with voxel-level granularity.
    We show a 3-channel flatmap showing the outcome of each location-selective driving experiment;
    a voxel is more red/green/blue if that voxel was driven by the corresponding ROI explanation.
    }
    \label{fig:fig2}
\end{figure}

\section{Explaining selectivity in regions of interest (ROIs)}
\label{sec:roi_driving}


Theories of language selectivity often revolve around regions of interest (ROIs) that are believed to be selective for specific semantic categories \cite{huth2016natural}, such as locations (OPA)~\cite{julian2016occipital, kamps2016occipital} or body parts (EBA)~\cite{taylor2007functional, calvo2010extrastriate}. We explored whether \method{} could be used to independently uncover the semantic selectivity of regions of interest, both for generating new hypotheses as well as re-establishing known functional selectivity.
We used \method{} to find explanations and design synthetic driving stimuli for a diverse selection of individual ROIs that were localized using separate stimuli: the extrastriate body area (EBA), intraparietal sulcus (IPS), occipital face area (OFA), posterior superior temporal sulcus (pSTS), superior premotor ventral (sPMv), retrosplenial cortex (RSC), the parahippocampal place area (PPA), and the occipital place area (OPA). These ROIs were identified in each subject using separate localizer scans.
\cref{fig:fig2}a shows the generated explanations and average response above baseline for each ROI during each driving experiment. The explanations found by \method{} and validated in the follow-up experiment are well-matched to known selectivity, e.g. \textit{Body parts} in EBA, and \textit{Scenes and settings} in PPA.
Driving succeeded for every ROI (all $p < 0.05$; permutation test with FDR correction), although for some ROIs, e.g. RSC, the effect is stronger.
\cref{fig:fig2}b visualizes driving responses for each ROI on a composite flatmap, where each voxel is colored on the basis of how much it was driven by its ROI's explanation. This map suggests that the variability in driving between ROIs reflects functional heterogeneity within each region.

We next tested whether \method{} could be used to describe differences between regions that appear to have similar selectivity \cite{huth2016natural}.
Examining three ROIs known to be selective for places or locations (RSC, PPA, OPA)~\cite{epstein2019scene}, we found that the driving paragraphs designed independently for each ROI also drove responses in the others (\cref{fig:fig2}c left).
To differentiate these ROIs, we used \method{} to generate new explanations that should selectively drive each one of these regions while suppressing the other two~\cite{shinkle2023control}.
These selective explanations deviated somewhat from the original ROI explanations, 
e.g. for RSC the explanation changed from \textit{Travel and location names} to just \textit{Location names}, suggesting that the \textit{travel} part of the explanation is common across ROIs while \textit{location names} are more unique to RSC. New stories were generated to include only instances of the corresponding explanation, for instance, when driving only RSC, location names were included, while directions were specifically excluded (see prompts in \cref{sec:appendix_prompts}). Since these two categories commonly cooccur in natural language, these synthetic stimuli enabled us to examine differences that might not be present in typical naturalistic studies.
Testing these stories in another fMRI experiment showed that \method{} was able to find explanations that can selectively drive two of the ROIs \textit{Location names} for RSC and \textit{Unappetizing foods} for PPA ($p < 10^{-3}$; permutation test of difference between ROI and other location ROIs).
In contrast, the explanation of \textit{Spatial positioning \& directions} for OPA still drove responses in all three ROIs (\cref{fig:fig2}c right; \cref{fig:fig2}d).
These results demonstrate that \method{} can be used to build more nuanced theories of cortical semantic selectivity beyond general trends that are well-understood.

\section{\add{Discovering new functional ROIs using \method}}
\label{sec:micro_rois}

The previous results showed that \method{} can evaluate explanations for known ROIs, so we next asked if \method{} can also discover new functional ROIs. To take advantage of \method's strengths, we focused our search on very small `micro ROIs' in prefrontal cortex. Micro ROIs are particularly easy to miss in group fMRI analyses~\cite{NietoCastanon2012,amunts2000brodmann}, and earlier work has suggested that prefrontal cortex may contain many small specialized functional areas ~\cite{huth2016natural,braga2017parallel}. Discovering such regions would help develop a more granular picture of cortical functional organization, but this discovery process can be quite challenging owing to the expansive hypothesis space of potential explanations.

We generated \textit{candidate ROIs} across PFC by selecting small circular regions of cortex for two subjects with a mean size of 1,019 voxels.
This resulted in 582 candidate ROIs for S02 and 535 candidate ROIs for S03.
We then ran the \method{} explanation pipeline (\cref{fig:fig1}a-b) to generate explanations for each candidate ROI.
We filtered out explanations with a stability score below 0.6 (see \cref{sec:methods}); this resulted in 73 stable candidate ROIs for S02 and 128 stable candidate ROIs for S03.
Finally, we grouped the explanations into 4 common categories using keyword search and built \method{} stories to evaluate the driving scores of those explanations. These stories were used in a new fMRI experiment.

We found that many of the candidate ROIs were driven as expected (\cref{fig:fig3}a).
Among the selected candidate ROIs, 47 were significantly driven for S02 and 21 were significantly driven for S03 ($p<0.05$, 1-tailed t-test with FDR correction).

\cref{fig:fig3}b shows the driving scores for a selection of these candidate ROIs across the two subjects.
Many of the candidate ROIs are in consistent locations across subjects. The first set of candidate ROIs are selective for the concept of ``recognition'' and are found in a small region near Brodmann area 9 in medial prefrontal cortex. The selectivity is specific to instances where one person is described to have recognized or noticed another (e.g. \textit{my friends saw me}, \textit{the guard spotted us}). This new finding may relate to earlier findings linking this broad region to theory of mind~\cite{goel1995modeling}. However, our results show an unexpectedly high degree of specificity both anatomically and semantically. The second set of candidate ROIs are found near the right hemisphere analogue of Broca's area and are selective for words that signal dialogue between two parties (such as \textit{said} or \textit{told}). The proximity of this ROI, which is selective for language describing speech acts, to Broca's area, which carries out speech acts, enhances grounding theories of cortical organization~\cite{hauk2004somatotopic}. The third set of candidate ROIs are near Brodmann area 46 and are selective for mentions of times (e.g. \textit{one o'clock}). This, along with similar selectivity found in the final set of candidate ROIs for numeric measurement words (e.g. \textit{fifty feet}) refines research that has found this area may be involved in numeric processing or calculation~\cite{burbaud1995lateralization}. The similarities between these two selectivities extend past this single localized region and suggest the presence of a network of regions in right PFC that is specialized for numeric processing. Taken together, these new micro-ROIs provide a refined understanding of semantic processing across PFC.


\section{Adversarially testing functional similarity with \method}

\begin{figure}[t]
    \centering
    \makebox[\textwidth]{
    \includegraphics[width=1.2\textwidth]{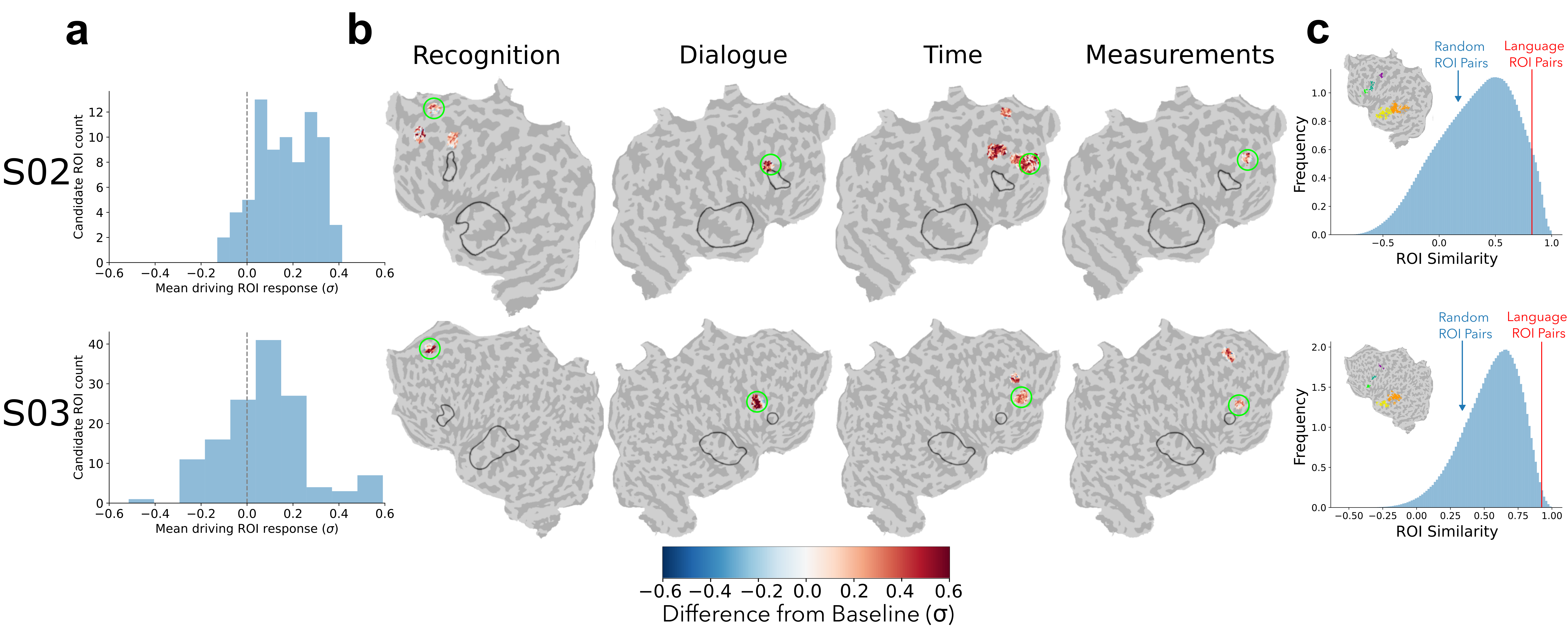}
    }
    \caption{
    \textbf{\add{Evaluating hypothesized micro-ROIs in prefrontal cortex using \method}}.
    (a) To measure \method's ability to aid in the discovery of new brain regions, spatially-contiguous candidate ROIs were defined in a grid pattern throughout prefrontal cortex. Candidate ROIs with high stability scores (see~\cref{sec:factors}) were filtered out to define stable ROIs. \method{} was used to automatically generate explanations and driving stimuli for these ROIs. The ROI responses to the driving stimuli were then measured in an fMRI experiment, and their average driving scores are shown in a histogram for subject S02 and S03. In both subjects, high-stability candidate ROIs are driven using their corresponding explanation at statistically significant rates; 47 out of 73 are significantly driven for S02 and 21 out of 128 are significantly driven for S03 ($p<0.05$, 1-tailed t-test, FDR-corrected).
    (b) Significantly driven candidate ROIs across the two subjects are visualized, colored by their driving scores. Significant candidate ROIs for various explanations (\textit{Recognition, Time, Dialogue, and Measurements}) are in consistent locations across the subjects, suggesting population level trends.
    (c) We next examined whether \method{} can validate functional similarity claims between 5 regions in the language network, localized using an established localizer~\citep{fedorenko2012language}.
    We concatenated the driving scores of each of these ROIs into a vector and then computed the cosine similarity between the vectors for different ROIs.
    We find that the driving vectors of language network ROIs are on average more similar to each other than over 94\% of randomly selected pairs of similarly-sized ROIs in UTS02 and over 99\% of randomly selected pairs in UTS03 ($p < 10^{-5}$; 1-tailed $t$-test), supporting claims of functional similarity across the language network.
    }
    \label{fig:fig3}
\end{figure}

Proving that two brain areas have different functional selectivity requires only that the areas differ along one functional dimension. But proving that two brain areas have the same selectivity is much more difficult, as one must show that there is no dimension that differentiates the areas. \method{} offers an approach to tackle this issue, because it can adversarially search for stimuli that differentially activate brain areas as we showed in \cref{fig:fig2}. Provided that we have high confidence that we would be able to generate such a stimulus if it existed, then our failure to do so would provide affirmative evidence for the similarity of the two regions.

As a demonstration of this use case, we focused on the putative `language network' \cite{fedorenko2024language}, a set of regions that respond readily to language and are claimed to be functionally indistinguishable. First, we localized five language regions in each of two subjects using established methods~\cite{fedorenko2012language}. For each region, we then used \method{} to generate stimuli that would drive that region while minimizing the response in the other regions, as we did successfully in \cref{fig:fig2}c. Using this approach, we found \textit{no stimuli} that could drive one region while suppressing the average of the others, strongly supporting claims of functional homogeneity in the language network. \cref{fig:fig3}c summarizes our results. We found that the driving behaviors of language network ROIs are more similar to each other, as measured by cosine similarity across explanations, than over 94\% of randomly selected pairs of similarly-sized ROIs in UTS02 and 99\% of all pairs in UTS03 ($p < 10^{-5}$; 1-tailed t-test). This result both serves to validate claims of functional similarity across the language network~\cite{fedorenko2024language} and to  demonstrate that \method{} can be used as a powerful tool for detecting functional similarity as well as functional differences across ROIs. A full dissection of these language network driving experiments is given in Appendix~\ref{sec:app_language_network}.

\section{Factors affecting \method’s ability to drive a voxel}
\label{sec:factors}
\begin{figure}[t]
    \centering
    \makebox[\textwidth]{
    \includegraphics[width=1.16\textwidth]{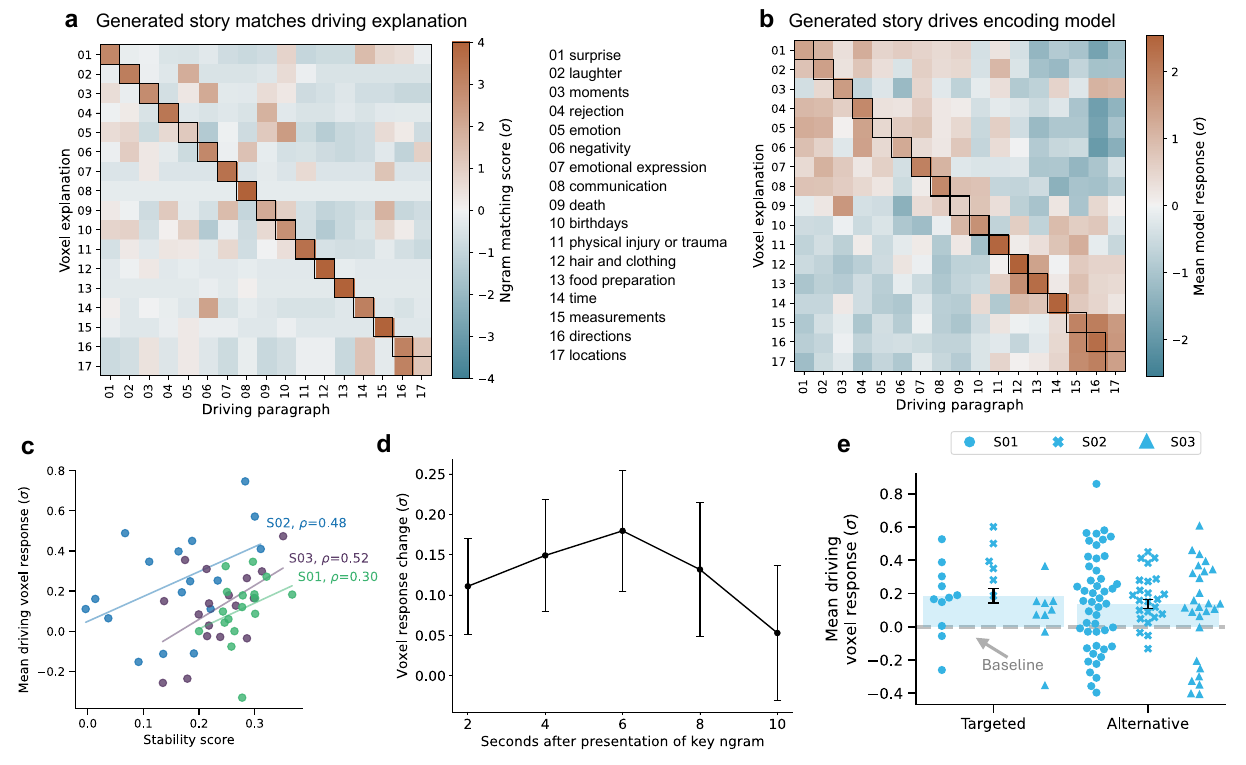}
    }
    \caption{\textbf{Analyzing factors that impact generative causal testing.}  
    To evaluate whether \method{} succeeds in generating effective stimulus stories, we assessed the driving paragraphs of the stimulus.
    (a) To confirm that generated paragraphs match the explanation used to construct them, a matching score was computed for each explanation and paragraph by using an LLM to evaluate the fraction of trigrams in the paragraph that are relevant to the paragraph’s generating explanation and then z-scoring the result. Each driving paragraph showed a strong match with its generating explanation. Plot shows one subject (S02), similar plots for other subjects are shown in \cref{subsec:stratifying_driving_patterns}.
    (b) To confirm that each driving paragraph effectively drives its corresponding encoding model predictive performance, we computed the predicted response in each selected voxel to each generated paragraph. This revealed strong matches for most voxels, along with some matches between driving paragraphs and voxels with semantically similar explanations, e.g. \textit{directions} and \textit{locations}. Plot shows one subject (S02), similar plots for other subjects are shown in \cref{subsec:stratifying_driving_patterns}. Paragraph order is presented here according to semantic similarity, with more similar explanations presented closer to each other.
    (c) After running the fMRI driving experiment, we found that a key factor determining whether a voxel is driven well by the \method{} framework was the stability score for the voxel, i.e. the correlation between the \textit{n}-gram rankings provided by the LLaMA-based encoding model and the OPT-based encoding model.
    (d) Another important factor for eliciting increased driving responses is the presence of key \textit{n}-grams in the driving paragraphs.
    These \textit{n}-grams induce a standard hemodynamic response curve that peaks at around 6 seconds, yielding a significant increase ($p = 0.009$; one-sided t-test).
    (e) Finally, to test whether the driving results were sensitive to the particular voxels that were selected, we evaluated whether the \method{} stories drove alternative voxels in each subject that were assigned the same explanation as the target voxels being driven.
    Both the targeted voxels and alternative voxels also showed significantly increased driving responses ($p < 0.05$; permutation test, FDR-corrected).
    }
    \label{fig:fig4}
\end{figure}

The experiments thus far showed that \method{} can effectively explain and drive a variety of locations throughout cortex, but fails to successfully drive some voxels.
To understand these limitations, we evaluated factors that influence driving performance, beginning with factors surrounding the generation of story stimuli.

First, we considered whether driving failures may arise from a semantic mismatch between the proposed explanation and the generated story stimulus.
We measured the match between each driving paragraph and its target explanation for subject S02 by using an LLM to evaluate similarity between the explanation and trigrams in the driving paragraph (see \hyperref[sec:methods]{Methods}).
There is a strong correspondence between each voxel explanation and its driving paragraph (\cref{fig:fig4}a, orange diagonal), suggesting that LLM stimulus generation does not account for the failures.
Moreover, there is no clear correlation between these scores and driving success (average correlation of -0.05; see \cref{fig:encoding_score}a).
Second, we tested whether the failed cases result from a misalignment between the original encoding model for each voxel and the generated story stimulus.
Again, each voxel's encoding model showed an increased response for its driving paragraph (\cref{fig:fig4}b, orange diagonal), suggesting the driving paragraphs are successfully aligned with the encoding model.
In this case, higher driving scores for the encoding model did yield higher driving scores in the follow-up experiment for 2 of the 3 subjects (average correlation of 0.19; see \cref{fig:encoding_score}b). 



Given that driving failures are neither the result of stimulus-explanation mismatches nor stimulus-model mismatches, we concluded that they must largely stem from limitations in the encoding models.
We only applied \method{} to voxels with high encoding model test performance, but this does not guarantee that the encoding model is accurate. Model performance is measured on a dataset that is limited in scope and that comprises real, noisy data. Both of these factors could cause a voxel to appear well-predicted despite having an inaccurate encoding model, or to appear poorly-predicted despite having an accurate encoding model.
Since these limitations cannot be addressed directly, we investigated factors that improve confidence in the generated explanations
by assessing their stability, a key principle underlying effective statistical interpretation according to the predictability, computability, and stability (PCS) framework~\cite{yu2020veridical,yu2024veridical}.

To identify which explanations are stable, we defined a stability score that measures agreement between the predictions of two encoding models built from different LLMs (see \hyperref[sec:methods]{Methods}) for the same voxel using the same fMRI data.
We found a strong positive correlation between the stability score of a voxel and its mean driving response across all three subjects (\cref{fig:fig4}c), suggesting that the stability score reliably informs the causal reliability of an explanation. This result underscores the idea that models must be predictive and stable in order to effectively generate hypotheses for follow-up experiments~\cite{abbasi2018deeptune,behr2024learning}.

We also assessed the stability of our framework to many modeling choices including
the prompts used to generate synthetic stories (\cref{fig:prompt_variability}) and the granularity of units we drive (e.g. single voxels versus semantic clusters of voxels in \cref{fig:cluster_vs_single}).
In both cases, we found that the magnitude of the evoked driving responses may change, but the overall trends in driving voxels are consistent.

To further understand how driving paragraphs succeed, we measured whether the inclusion of key driving \textit{n}-grams extracted from the encoding model drove responses in a temporally specific fashion (\cref{fig:fig4}d).
During the fMRI experiment, the presentation of these \textit{n}-grams evoked a significant increase in responses, peaking 6 seconds after presentation (\cref{fig:fig4}d; $p = 0.009$; one-sided t-test).
This strongly aligns with hemodynamic response curves which tend to peak about six seconds after stimulus presentation~\cite{liao2002estimating}, and are further validation that the generated stimuli are the cause of the driving effects we observe.

Finally, we tested whether the \method{} framework is sensitive to the particular voxels we selected for followup experiments.
To evaluate this,
we evaluated whether the driving paragraphs we generated also drove alternative voxels that were assigned the same explanation by \method.
\cref{fig:fig4}e shows the mean driving voxel responses for targeted voxels versus alternative voxels and finds that both are driven significantly ($p < 0.05$; permutation test with FDR correction).
The alternative voxels are driven slightly less reliably (mean 0.14$\sigma$ vs 0.19$\sigma$), likely because the prompts for generating the stories contain \textit{n}-grams that specify the explanation in slightly more detail than the explanation alone.

Besides the factors discussed above,
all follow-up experiments use stimuli that are presented visually rather than through audio (see \hyperref[sec:methods]{Methods}).
This shift helps to test for semantic generalization.
Despite this shift and the inherent noise in the voxels and imaging process, \method{} still generally succeeds in targeted driving.

\begin{figure}[t]
    \centering
    \makebox[\textwidth]{
    \includegraphics[width=1.2\textwidth]{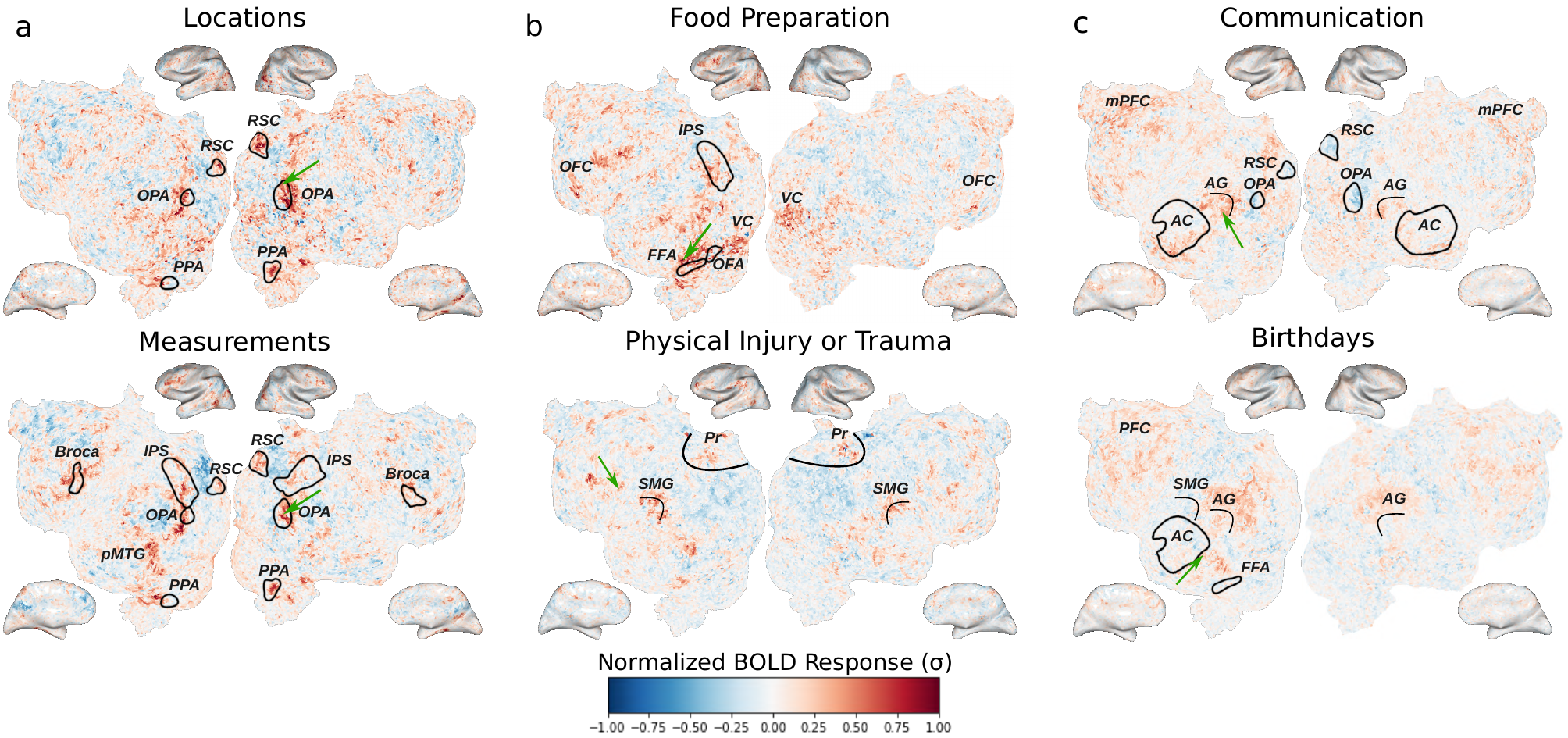}
    }
    \caption{
    \textbf{Cortex-wide responses to driving paragraphs reproduce known semantic contrasts and open new hypotheses}.
    To understand how each explanation activates the entire cortex, we visualize the driven response (average normalized BOLD response) for six explanations.
    (a) Some maps recovered well-established results, e.g. the explanation \textit{Locations} gave strong responses in the place areas RSC, OPA, and PPA.
    (b) Other maps independently confirmed newer hypotheses, e.g. the explanation \textit{Food Preparation} activates a region in ventral occipital cortex near the fusiform face area (FFA)~\cite{jain2023selectivity, khosla2022highly}.
    (c) Finally, some explanations did not clearly map to any known hypotheses, but may suggest new directions for future research.
    All observed activation patterns along with ROI localization explanations are presented in Appendix~\ref{sec:appendix_topic_activation}.
    }
    \label{fig:fig3_old}
\end{figure}

\section{Responses to synthetic stories reveal cortex-wide contrast effects}
The \method{} framework is modular: each stage---encoding model fitting, explanation generation, and stimulus construction---can be used independently or combined with external inputs. One practical consequence is that the stimulus generation stage (\cref{fig:fig1}c) does not require the earlier explanation discovery stages (\cref{fig:fig1}a-b). A researcher with a pre-existing hypothesis about what drives a brain area can directly use \method{} to generate controlled narrative stimuli for that hypothesis and test it in an fMRI experiment.

Furthermore, although \method{} stories are designed to activate specific voxels or ROIs, whole-brain fMRI measurements capture responses everywhere in cortex simultaneously. Each driving paragraph thus provides a cortex-wide activation map for its target explanation, analogous to a classical block-design contrast but embedded within a coherent narrative that sustains subject engagement~\cite{hamilton2020revolution, jain2023computational}.


To demonstrate the utility of these cortex-wide maps, we examined the whole-cortex average activation map for each explanation, computed by averaging responses over timepoints within each paragraph.
\cref{fig:fig3_old} shows the results for six selected explanations for subject S02; maps for other explanations are provided in Appendix~\ref{sec:appendix_topic_activation}).
Some explanations were similar to previously well-established contrasts (\cref{fig:fig3_old}a) and gave the expected results. For example, the explanation \textit{Locations} gave strong responses in the place areas RSC, OPA, and PPA~\cite{epstein2019scene}.
Others confirmed newer hypotheses. For example, the \textit{Food Preparation} explanation (\cref{fig:fig3_old}b \textit{Top}) activates a region in ventral occipital cortex near the fusiform face area (FFA). This area was recently found to respond selectively to ``food'' images~\cite{jain2023selectivity, khosla2022highly}, but our results demonstrate for the first time that this selectivity also extends to linguistic descriptions of food.
Finally, some explanations did not clearly map to any known hypotheses (\cref{fig:fig3_old}c), but still demonstrate suggestive patterns of activation that could be explored in future experiments.

The GCT average responses can be compared quantitatively to selectivity maps that a researcher may be interested in exploring. As an example, \cref{tab:eng1000_comparison} shows quantitative comparisons to selectivity maps originating from an interpretable encoding model in a prior study~\cite{huth2016natural}, often showing strong agreement between the selectivity maps and the GCT average responses.

\FloatBarrier
\section{Discussion}\label{sec:discussion}

The main advance in this work is a closed-loop approach for translating opaque, high-performing encoding models into testable hypotheses and then evaluating those hypotheses with model-guided stimulus interventions.
By applying this approach at multiple spatial scales, we show that it can recover established semantic contrasts and also discover finer-grained functional structure that is difficult to probe with standard naturalistic or block-design paradigms.
More broadly, \method{} provides a clear pathway to make experimentally testable several central unresolved questions in language neuroscience, in particular the granularity of semantic organization, the dissociability of ostensibly similar regions, and the role of stimulus dependencies in semantic selectivity.
Prior studies have demonstrated the predictive power of black-box encoding models in language neuroscience across modalities such as fMRI, ECoG, EEG, and MEG~\cite{JAINNEURIPS2018, TONEVANEURIPS2019, abnar2019, goldstein2022shared, antonello2023scaling}, but these models have been criticized for their inability to produce accurate and reliable explanations~\cite{chirimuuta2021prediction} and limited efforts have been made to explicitly interpret these models \cite{JAINNEURIPS2020, chen2023cortical, Vo2023.08.03.551886,luo2024brainscuba}. Our work takes the first steps in overcoming this criticism by enabling us to directly test the causality of an encoding model explanation through the use of LLM-generated synthetic stimuli.

This work is related to earlier studies that generated stimuli to drive neurons in the visual cortex~\cite{bashivan2019neural,abbasi2018deeptune, murty2021,ponce2019evolving,walker2019inception}, as well as studies that used decoding to provide insight into the selectivity of the visual cortex \cite{luo2024brainscuba} or generated controversial visual stimuli to differentiate models~\cite{golan2020controversial}.
In language fMRI, the closest work is \citeauthor{tuckute2023driving}~\citeyear{tuckute2023driving}, which demonstrated the ability to drive the language network as a whole (rather than individual voxels or ROIs) using sampling techniques as demonstrated in \ref{fig:checkerboard}. In comparison to these prior works, ours is the first to demonstrate driving at an explanatory level, not only identifying stimuli that can drive a particular voxel or ROI, but also hypothesizing the shared semantic features that cause the underlying activity. 
This distinction provides an important scientific benefit, allowing us to automate the discovery of new and unexpected regions of interest for further study. As encoding models improve, the reliability of such discoveries will naturally increase as well.

Cognitive neuroscience has always been concerned with finding the features of stimuli that specifically cause activation in the brain.
Traditionally this problem has been approached by manually searching for these critical stimulus properties across multiple experiments~\cite{kanwisher2006fusiform,kamps2016occipital}.
In contrast, data-driven methods~\cite{JAINNEURIPS2018, TONEVANEURIPS2019, abnar2019, goldstein2022shared, antonello2023scaling} efficiently probe a much larger hypothesis space, but fail to causally validate their findings.
Our method combines the principled approach of causal experiments with the strengths of data-driven models to speed up scientific discovery. 

Our work can therefore be applied to generating a new class of stimuli for online, causal experiments of language processing. We consider this work an important part of a larger closed-loop system that continually refines its understanding of brain selectivity by generating hypotheses, testing these hypotheses with new experiments, and refining the hypotheses based on the experimental results.
As an example, such a framework could be an especially effective tool in lesion case studies where general observations must be made efficiently from relatively little data.

While effective,
\method{} has several limitations.
Most notably, it focuses on a single explanation, missing voxel activity that is polysemantic, i.e. that is not driven well by any single explanation.
See \cref{sec:appendix_polysemantic} for a more detailed discussion of this limitation and how future work might overcome it. Second, the effectiveness of \method{} depends on staying within the manifold of stimuli trained on during encoding model training.
This limitation is explored further in \ref{sec:appendix_gpt1_checkerboard} where we show that poorly generated, off-manifold stimuli are more difficult to predict and yield inferior driving outcomes.
\method{} is also sensitive to the properties of the underlying encoding models--using different encoding models could yield different explanations which might be equally valid. Thus we must be cautious not to interpret the explanations given by \method{} as uniquely capturing the functional properties of a voxel or ROI. However, this issue will become less severe over time as better and more diverse encoding models become available.

The work here opens the door to a great deal of future research.
One avenue is to use \method{} to more systematically probe what degree of explanatory granularity is appropriate for cortical language selectivity.
For example, one of the explanations proposed by \method{} is \textit{birthdays}, but it would be somewhat surprising to have a specifically birthday-selective area in human cortex. Rather, it is more plausible that such a voxel is selective for something more generic, such as social gatherings~\cite{bedny2014shindigs}. Such ambiguities are of the type that are liable to be resolved by higher-fidelity modelling or iterative causal experiments, such as the hypothesis-refinement feedback system proposed above. \method{} is also fundamentally limited by the spatiotemporal granularity of the scanning modality we test it on, fMRI. Applying \method{} with faster or more fine-grained scanning modalities, such as 7T laminar fMRI, ECoG, or NeuroPixels, could reveal selectivity patterns that cannot be effectively observed through fMRI. 

\method{} demonstrates a general framework for aligning the data-driven predictions of models to the theory-oriented hypotheses of classical science, extending the predictability, computability, and stability (PCS) framework for veridical data science~\cite{yu2020veridical,yu2024veridical} into the domain of active experimental design.
In this paper, we show that applying techniques from modern LLM interpretability research can improve our understanding of cognitive brain processes. Through simple prompting-based methods, we successfully recover semantic selectivity hypotheses that have previously been the providence of painstaking controlled experiments created over decades.
Moreover, we open the door to many more similar results including the identification of micro ROIs in poorly understood brain areas.
Our framework is extensible to other modalities such as vision and - requiring only naturalistic data from a single subject - can be easily adapted for use in case study analyses. More generally, the high-level idea of \method{}, using generative models as a tool to operationalize imprecise hypotheses about the natural world, has general applications well outside the bounds of neuroscience entirely.
\method{} will help move toward a future with increasingly nuanced, operationalized, and falsifiable explanations in our scientific discourse. 

\FloatBarrier



{
    \bibliography{refs}
}

\section{Methods}\label{sec:methods}

\textbf{Data collection}
Two sets of MRI data are used in this study: the first for fitting encoding models and the second for selectively driving brain regions.
The original set is described and made openly available in previous work~\citep{lebel2022natural,tang2023semantic};
we describe the details for the second newly collected set here.
Functional magnetic resonance imaging (fMRI) data were collected from the same 3 human subjects as the original set as they listened to English-language podcast stories over Sensimetrics S14 headphones.
Subjects were not asked to make any responses, but simply to listen attentively to the stories. 
All subjects were healthy and had normal hearing.
The experimental protocol was approved by the Institutional Review Board at the University of Texas at Austin.
Written informed consent was obtained from all subjects.

All MRI data were collected on a 3T Siemens Skyra scanner at University of Texas at Austin using a 64-channel Siemens volume coil.
    Functional scans were collected using a gradient echo EPI sequence with repetition time (TR) = 2.00 s, echo time (TE) = 30.8 ms, flip angle = 71°, multi-band factor (simultaneous multi-slice) = 2, voxel size = 2.6mm x 2.6mm x 2.6mm (slice thickness = 2.6mm), matrix size = 84x84, and field of view = 220 mm.
    Anatomical data were collected using a T1-weighted multi-echo MP-RAGE sequence with voxel size = 1mm x 1mm x 1mm following the Freesurfer morphometry protocol ~\citep{fischl2012freesurfer}.

\textbf{Data preprocessing}
    All functional data were motion corrected using the FMRIB Linear Image Registration Tool (FLIRT) from FSL 5.0. FLIRT was used to align all data to a template that was made from the average across the first functional run in the first story session for each subject. These automatic alignments were manually checked for accuracy. 
    
    Low frequency voxel response drift was identified using a 2nd order Savitzky-Golay filter with a 120 second window and then subtracted from the signal. To avoid onset artifacts and poor detrending performance near each end of the scan, responses were trimmed by removing 20 seconds (10 volumes) at the beginning and end of each scan, which removed the 10-second silent period and the first and last 10 seconds of each story. The mean response for each voxel was subtracted and the remaining response was scaled to have unit variance.

\textbf{Voxelwise encoding models}
    Encoding models were fit for each of the three subjects on listening data from roughly 20 hours of unique stories across 20 scanning sessions, yielding a total of  \(\sim \)33,000 datapoints for each voxel across the whole brain.
    For model testing, the subjects listened to the two test stories five times each, and one test story 10 times, at a rate of 1 test story per session. These test responses were averaged across repetitions.
    
    For each subject and each voxel, we fit a separate encoding model to predict the BOLD response from the features we extract from the stimulus.
    Features were extracted from the 18th layer of the 30-billion parameter LLaMA model~\citep{touvron2023llama},
    and the 33rd layer of the 30-billion parameter OPT model~\citep{zhang2022opt}. These layers were chosen based on prior work~\citep{antonello2023scaling} that determined that they were the most performant at predicting brain activity. 
    To temporally align word times with TR times, Lanczos interpolation was applied with a window size of 3.
    The hemodynamic response function was approximated with a finite impulse response model using 4 delays at -8,-6,-4 and -2 seconds \citep{huth2016natural}.

    To evaluate the voxelwise encoding models, we used the learned encoding model to generate and evaluate predictions on a held-out test set.
    The OPT features achieved a mean voxelwise correlation of about 0.128 whereas the LLaMA features achieved a mean voxelwise correlation of about 0.132.
    These performances exceed that of previously published models on the same dataset (mean correlation 0.111) that were able to produce meaningful semantic decoding~\citep{tang2023semantic}.

\textbf{Voxel selection for explanation}
    We selected 500 well-modeled, diverse voxels to explain for each subject.
    To ensure that these voxels were well-modeled, we selected only from voxels with a test correlation above 0.15, (this corresponds to the top \~40\% most well-predicted voxels).
    Then, we applied principal components analysis to the learned ridge weights across all well-modeled voxels in cortex. We project all voxels to the first four principal components, which are known to encode differences in semantic selectivity~\citep{huth2016natural} and then uniformly sampled voxels from the within the convex hull in $\mathcal{R}^4$ induced by this projection. This uniform sampling ensures diversity in the selected voxels.  
    The mean voxel correlation for the 1,500 voxels we study is 0.35.

\textbf{Framework for generating explanations}
    \method{} yields a short, natural-language explanation describing what elicits the strongest response from each of the 1,500 selected voxel encoding models. 
    To obtain this explanation, we follow the summarize and score (SASC) framework introduced in previous work~\cite{singh2023explaining} (see \cref{fig:fig1}b).
    SASC first generates candidate explanations (using GPT-4 \cite{openai2023gpt4system})
    based on the \textit{n}-grams ($n=1,2,3$) that elicit the most positive response from the LLaMA encoding model.
    Each candidate explanation is then evaluated by generating synthetic data based on each explanation.
    The response of the encoding model to the synthetic data is tested, and the candidate explanation corresponding to the synthetic data that elicits the most positive response from the encoding model is selected.

\textbf{Stability score screening}
    After extracting an explanation for 500 voxels per subject, we selected 17 voxels per subject for followup fMRI experiments.
    17 voxels were selected so that the resulting 17-paragraph stories would be similar in length to the stimuli used to initially fit the encoding models~\cite{lebel2022natural}.
    To select voxels which are not only well-predicted but also likely to be well-explained, we introduced the \textit{stability score},
    which measures the correlation of predictions by the encoding models based on OPT and LLaMA for each unique \textit{n}-gram present in the dataset of stories.
    A higher correlation implies greater agreement between the different encoding models and thus greater stability.
    For each subject, we first filter the 40 voxels with the highest stability score and then manually select 17 voxels from this set that have diverse explanations.
    The 17 voxels for each of the three subjects were sampled mostly from the ventral visual stream, the temporal lobe, and frontal regions (see precise locations in \cref{sec:appendix_topic_activation}).

\textbf{Story generation (single-voxel setting)}
    For follow-up experiments, we prompted a large LLM (GPT-4~\citep{openai2023gpt4system}) to generate stories based on the explanations for our selected voxels.
    Stories are generated by repeatedly prompting the LLM in a chat to continue the story, one paragraph at a time.
    For each paragraph, the LLM is asked to focus on one explanation and to include related key \textit{n}-grams (\cref{fig:fig4}a, see full prompts in \cref{sec:appendix_prompts}).
    
    Before running follow-up experiments, we checked that each paragraph’s text matched its generated explanation (\cref{fig:fig4}a).
    This match was measured by prompting an LLM~\citep{zhong2022describing} to evaluate the fraction of trigrams in the paragraph that are relevant to the paragraph’s generating explanation.
    We also validated that each story paragraph drives the encoding model for its corresponding voxel (\cref{fig:fig4}b).
    We generated 8 different stories by changing the random seed for each subject and kept the best 2 for S01, the best 6 for S02 (the pilot subject), and the best 2 for S03. During presentation, these newly presented stories were visually presented at approximately conversational cadence, instead of being listened to, as in the original encoding setting.

    
\textbf{ROI setting}
The ROI setting uses the same framework as the single-voxel setting, but instead of maximizing the response in a single voxel, we sought to maximize the average response of the encoding model outputs over all voxels in an ROI.
When selecting explanations, we used the average outputs of the OPT and LLaMA encoding models to help make results more stable.
In the `micro ROI' setting (\cref{sec:micro_rois}), subject S01 was omitted due to a logistical constraint with bringing subject S01 back to the lab for follow-up experiments.

\textbf{Permutation testing details}
The main \method{} driving results in the paper evaluate whether a particular voxel (or region of interest) yielded an increased average response to the driving paragraphs, i.e. the synthetic paragraphs generated specifically to drive that voxel.
The permutation tests we conducted compared this average response to the voxel's average response to other random paragraphs in the experiment.
Beyond permutation testing, GCT can be viewed more generally as a causal test where the treatment variable is the explanation underlying the driving paragraph and the outcome is the average observed fMRI response, though it may not perfectly match the assumptions underlying common frameworks such as the Neyman-Rubin causal model~\cite{holland1986statistics}.

\section*{Data availability}

All newly collected fMRI data will be made publicly available upon acceptance.
Data for fitting encoding models and generating explanations is available on OpenNeuro: \url{https://openneuro.org/datasets/ds003020}.

\section*{Code availability}

Code for running all experiments (as well as applying \method{} in new settings) is available on Github at \href{https://github.com/microsoft/automated-brain-explanations}{github.com/microsoft/automated-brain-explanations}.
Code uses python 3.10,
huggingface transformers 4.29.088–100, sklearn 1.2.040, and OpenAI API gpt-4 (gpt-4-0613 and gpt-4-0125-preview).

\section*{Author contributions}

R.J.A., C.S., S.J., B.Y., and A.G.H. contributed to experimental conception and design. R.J.A., S.J., and S.G. collected fMRI data. R.J.A. and C.S. performed data analysis. Voxelwise explanation code was written by C.S. and R.J.A. trained the encoding models used in the study. A.H. performed the stability analysis. Writing was primarily handled by R.J.A., C.S., and A.G.H., with editing from all authors. J.G., B.Y., and A.G.H. provided funding and supervision for the project.

\section*{Funding information}

We gratefully acknowledge support from NSF grants DMS-2413265, NSF grant 2023505 on Collaborative Research: Foundations of Data Science Institute (FODSI), the NSF and the Simons Foundation for the Collaboration on the Theoretical Foundations of Deep Learning through awards DMS-2031883 and 814639, NSF grant MC2378 to the Institute for Artificial CyberThreat Intelligence and Opera- tioN (ACTION) and NSF grant 1R01DC020088-001. In addition to these sources, this work received further support from the Burroughs-Wellcome Foundation, the Dingwall Foundation, and a gift from Microsoft Research.

\newpage
\begin{appendices}

{\LARGE \section{Extended Data}}

\subsection{Prompting details}
\label{sec:appendix_prompts}
\paragraph{Prompts used for explanation}
Following the original SASC study~\citep{singh2023explaining}, we use the following prompts for explanation.

The summarization step summarizes 30 randomly chosen ngrams from the top 50 and generates 5 candidate explanations using the prompt \textit{Here is a list of phrases:\textbackslash n\textcolor{darkgray}{\{phrases\}}\textbackslash nWhat is a common theme among these phrases?\textbackslash nThe common theme among these phrases is \blank}.

In the synthetic scoring step, we generate similar synthetic strings with the prompt \textit{Generate 10 phrases that are similar to the concept of \textcolor{darkgray}{\{explanation\}}:}.
For dissimilar synthetic strings we use the prompt \textit{Generate 10 phrases that are not similar to the concept of \textcolor{darkgray}{\{explanation\}}:}.
Minor automatic processing is applied to LLM outputs, e.g. parsing a bulleted list, converting to lowercase, and removing extra whitespaces.

\paragraph{Prompts used for story generation}

We use two variations of prompts.
In the main setting, we begin with \textit{Write the beginning paragraph of a long, coherent story. The story should be about "{expl}". Make sure it contains several words related to "{expl}", such as {examples}}.
Between paragraphs, it changes to \textit{Write the next paragraph of the story, staying consistent with the story so far, but now make it about...}

In the second setting, we instead begin with \textit{Write the beginning paragraph of an interesting story told in first person. The story should have a plot and characters. The story should be about… and continue with Write the next paragraph of the story, but now make it about...}

\paragraph{Prompts used for ROI-driving story generation}

For driving ROIs (\cref{sec:roi_driving}), many explanations are related to locations, and so we append a suffix to the prompt for reach paragraph (see \cref{tab:prompt_suffixes}) to help make paragraphs more distinct.

\begin{table}[ht]
  \centering
  \footnotesize
  \caption{Suffixes added to the prompt for generating the story paragraph for each ROI in UTS02.}
  \begin{tabular}{llp{7cm}}
  \toprule
  \textbf{ROI} & \textbf{Driving explanation} & \textbf{Suffix} \\
  \midrule

  RSC      & Travel and location names                  &  None \\
  RSC-only & Location names                             &  None \\

  OPA      & Direction and location descriptions        & Avoid mentioning any specific location names\newline (like ``New York'' or
  ``Europe''). \\
  OPA-only & Spatial positioning and directions          & Avoid mentioning any specific location names\newline (like ``New York'' or
  ``Europe''). \\
  PPA      & Scenes and settings                        & Avoid mentioning any specific location names\newline (like ``New York'' or
  ``Europe''). \\
  PPA-only & Unappetizing foods                         & Avoid mentioning any specific location names\newline (like ``New York'' or
  ``Europe''). \\
  IPS      & Descriptive elements of scenes or objects & Avoid mentioning any locations. \\
  pSTS     & Verbal interactions                        & Avoid mentioning any locations. \\
  OFA      & Personal growth and reflection              & Avoid mentioning any locations. \\
  sPMv     & Time and numbers                           & Avoid mentioning any locations. \\
  EBA      & Body parts                                 & Avoid mentioning any locations. \\

  \bottomrule
  \end{tabular}
  \label{tab:prompt_suffixes}
  \end{table}

\subsection{Polysemanticity in voxels}
\label{sec:appendix_polysemantic}

\subsubsection{Measuring voxel polysemanticity with question-answering encoding models}

To measure polysemanticity, we follow the setup proposed in \cite{singh2025evaluating}, which evaluates explanations by rephrasing them as yes-or no questions, annotating their answers using an LLM, then building an encoding model from the answers.
For each of the three subjects, 
we fit monosemantic encoding models to each voxel by selecting the single-question encoding model (from the 606 single-question models considered in \cite{singh2025evaluating}) that achieved the best cross-validation performance on the training set.
We then compare the performance of the single-question encoding model to the full encoding model.

\begin{figure}[H]
    \centering
    \includegraphics[width=\linewidth]{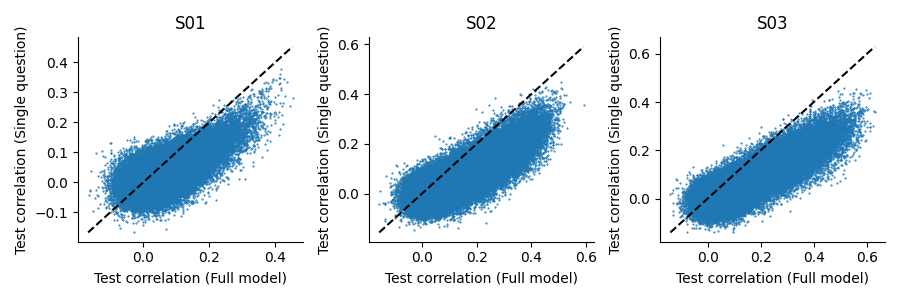}
    \caption{Full encoding model tends to outperform single-question monosemantic encoding model of voxels.
    Across the subjects, 79\% of voxels show an voxels show an improvement when predicting the full model (70\% for S01, 82\% for S02, and 85\% for S03).
    }
    \label{fig:voxel_polysemanticity}
\end{figure}

\cref{tab:voxel_polysemanticity} shows that single-question models achieve on average about half the predictive performance of the full model (0.056 test correlation versus 0.104).
The majority of voxels (79\%) show an improvement using the full model (\cref{fig:voxel_polysemanticity}).
These results indicate that a single-question encoding model is often insufficient to capture the selectivity of a voxel, supporting the presence of polysemanticity in many voxels.

\begin{table}[h!]
    \centering
    \caption{Average voxelwise test correlation achieved by two encoding models across subjects.
    The full encoding model tends to outperform single-question monosemantic encoding model of voxels, achieving roughly twice the test correlation.}
    \begin{tabular}{lcccc}
        \toprule
        & S01 & S02 & S03 & AVG\\
        \midrule
        Single-question & 0.029 & 0.060 & 0.062 & 0.056\\
        Full model & 0.057 & 0.123 & 0.133 & 0.104 \\
        \bottomrule
    \end{tabular}
    \label{tab:voxel_polysemanticity}
\end{table}

\subsubsection{Driving polysemantic voxels with \method}
We additionally run experiments seeking to drive explanations descriptions corresponding to the same voxel.
These experiments generate and test \method{} stories in the same manner as testing individual explanations~(\cref{fig:fig1}c) but differ in the way they source voxel descriptions.
In our first investigation (with subject S02), we select two explanations that both come from SASC, in the case that its summaries of the top ngrams yield two distinct explanations.
In our second investigation (now with subject S03), we instead select two explanations that each come from a different encoding model.
In both cases, each story tests 8 voxels with two explanations each.
We average the results over two stories per subject.

\begin{figure}[H]
    \centering
    \includegraphics[width=0.6\textwidth]{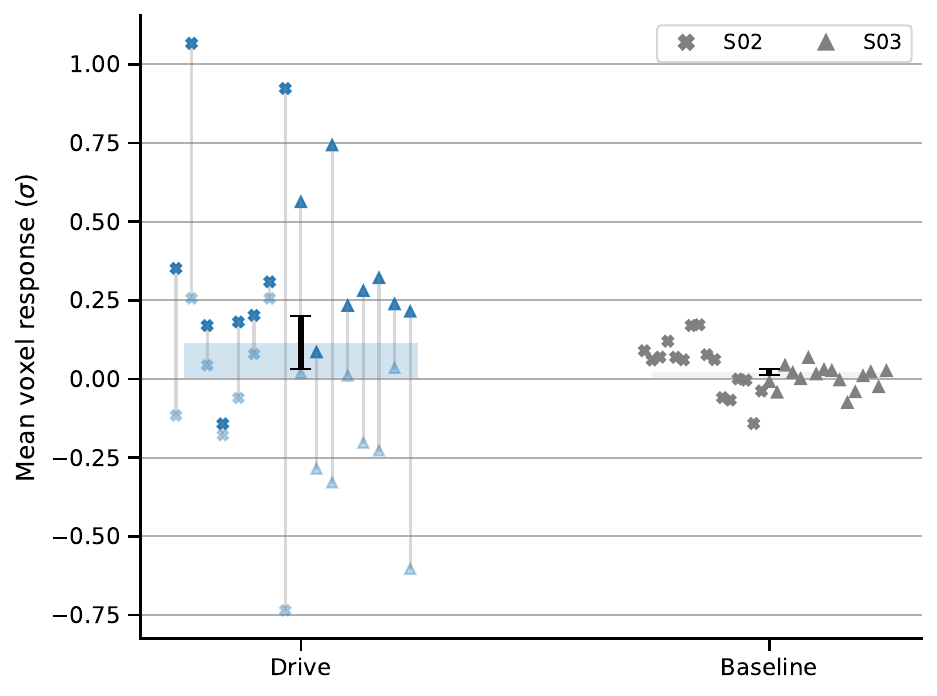}
    \caption{\textbf{Results for driving polysemantic voxels.}
    (a) Voxel response for driving paragraphs (blue) show a small increase relative to the baseline responses of the remaining paragraphs (gray).
    Each voxel appears as two points connected by a vertical line corresponding; each point shows the result when driving the voxel using a different explanation. Most voxels are only driven successfully for a single explanation.}
    \label{fig:polysemantic}
\end{figure}

We generally find that we are able to only successfully drive one of the two explanations we seek to drive for an individual voxel~(\cref{fig:polysemantic}).
We believe this is due to SASC's overreliance on summarizing top-driving ngrams, and could potentially be mitigated by summarizing a broader range of encoding model inputs.

\subsection{Story-level breakdowns}

\begin{figure}[H]
    \centering
    \includegraphics[width=0.8\textwidth]{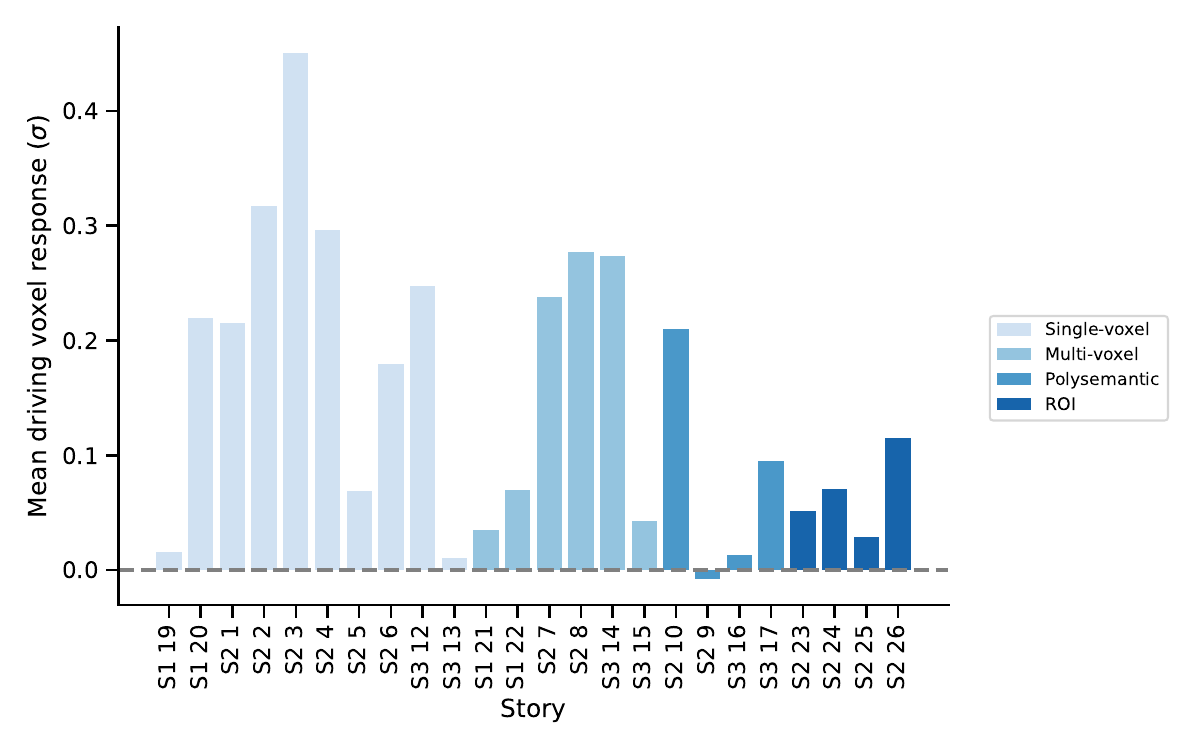}
    \caption{Different stories vary in their ability to drive voxels.
    There is large variance between the driving response produced for different stories, even for the same subject.
    Driving response shows the difference from baseline paragraphs.
    For ROI stories, this shows the average response of voxels in an ROI; for other stories this shows the average response of the single target voxel.
    }
    \label{fig:story_breakdowns}
\end{figure}

\begin{figure}[H]
    \centering
    \includegraphics[width=0.55\textwidth]{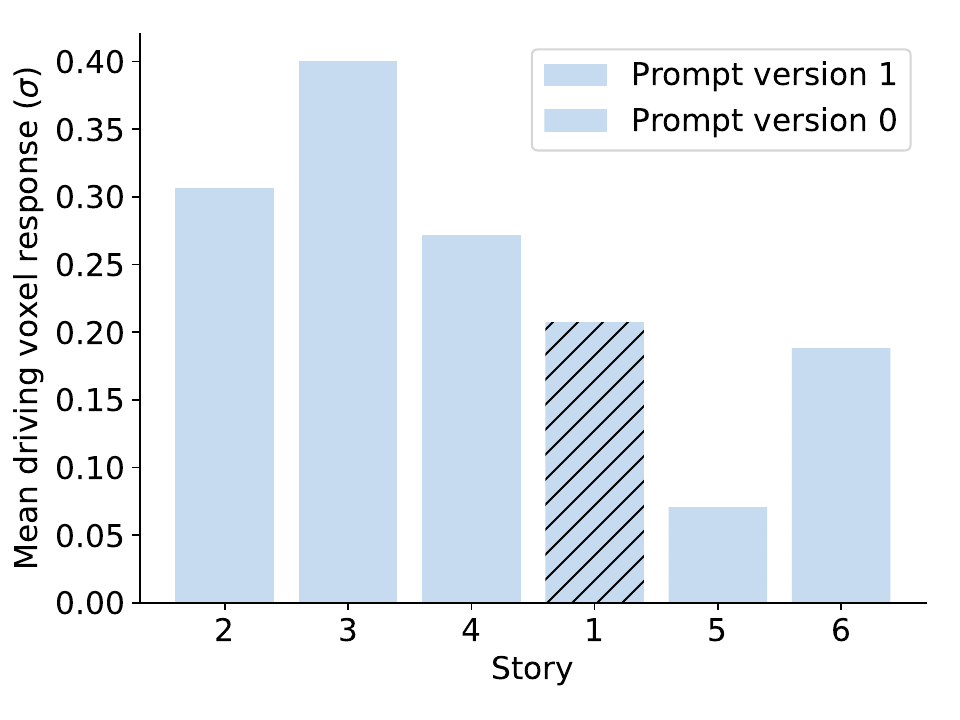}
    \caption{Different prompts yield different driving performance. The driving ability of the generated story differs based on the precise wording of the prompt. We compare two versions of the prompt, and find that results for version 1 are noticeably better.
    Version 1 is our default prompt, described in \cref{sec:appendix_prompts}.
    Version 0 instead begins with \textit{Write the beginning paragraph of an interesting story told in first person. The story should have a plot and characters. The story should be about..} and transitions between paragraphs with the prompt \textit{Write the next paragraph of the story, but now make it about...}.
    Results show the 6 stories run for S02 in the default setting.
    }
    \label{fig:prompt_variability}
\end{figure}

\begin{figure}[H]
    \centering
    \includegraphics[width=0.45\textwidth]{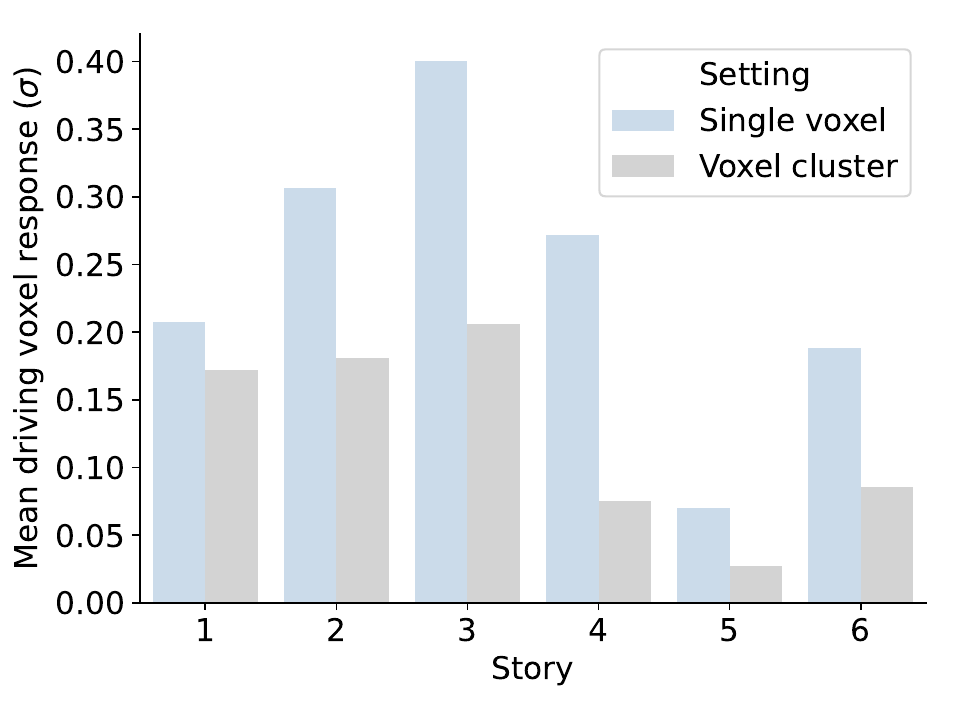}
    \caption{Clusters of voxels semantically related to the voxel cluster are driven, but less reliably than the target voxels themselves. Voxel clusters group semantically similar voxels by using the learned ridge regression weights of the encoding model.
    Results show the 6 stories run for S02 in the default setting.}
    \label{fig:cluster_vs_single}
\end{figure}

In this section, we show results for \method{} at the level of individual stories.
\cref{fig:story_breakdowns} shows the mean driving voxel response across all subjects and settings.
\cref{fig:prompt_variability} stratifies these synthetic stories based on the underlying prompt used to generate the stories and \cref{fig:cluster_vs_single} shows results for these stories at driving voxel clusters rather than individual voxels.

\begin{table}
    \centering
    \caption{Head motion for different subjects during \method{} stories.
    S01 shows substantially larger motion than the other two subjects.}
    \begin{tabular}{lcc}
      \toprule
      Subject & \makecell{Average framewise\\ displacement (mm)} & \makecell{Fraction of  TRs with framewise\\ displacement above 0.5 mm} \\
      \midrule
        S01 & 0.273 & 0.094\\ 
        S02 & 0.108 & 0.000\\
        S03 & 0.102 & 0.000\\
        \bottomrule
    \end{tabular}
    \label{tab:head_motion}
\end{table}

\subsection{Correlates and details of driving performance at the voxel level}
\label{subsec:stratifying_driving_patterns}

In this section we provide additional analyses of what voxel-level factors inform whether driving performance will succeed.
The strongest correlate is the stability score, given in the main text (\cref{fig:fig4}c).

\begin{figure}[ht]
    \centering
    \begin{tabular}{ll}
         \textbf{a} & \textbf{b} \\
        \includegraphics[width=0.48\textwidth]{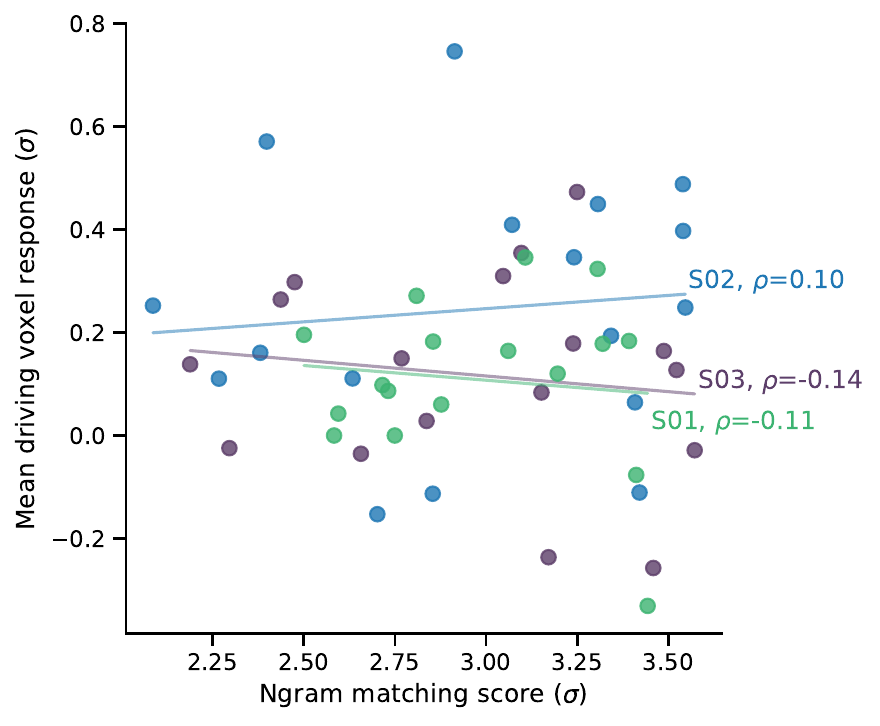} & \includegraphics[width=0.48\textwidth]{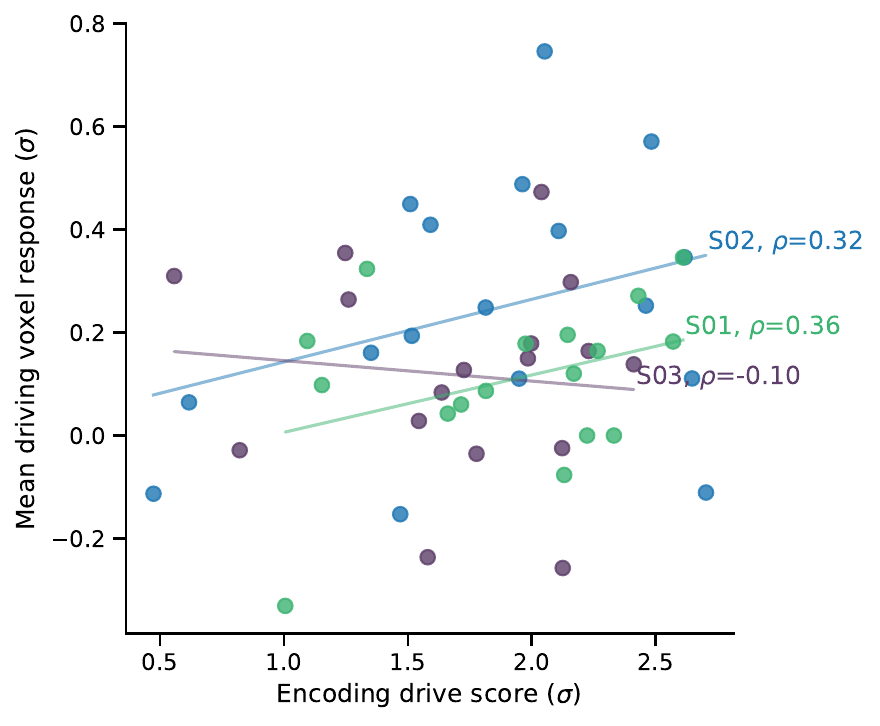}
    \end{tabular}
    
    \caption{
    Correlates of driving performance at the voxel level.
    (a) Ngram matching score, i.e. how well the driving paragraphs match their generating explanations, does not correlate well with driving score.
    (b) Encoding driving score, i.e. how well the driving paragraphs succeed in driving the encoding model, correlates well with driving score for 2 of the 3 subjects.}
    \label{fig:encoding_score}
\end{figure}

\begin{figure}
    \centering
    \begin{tabular}{cc}
         S01 & S03 \\
            \includegraphics[width=0.48\textwidth]{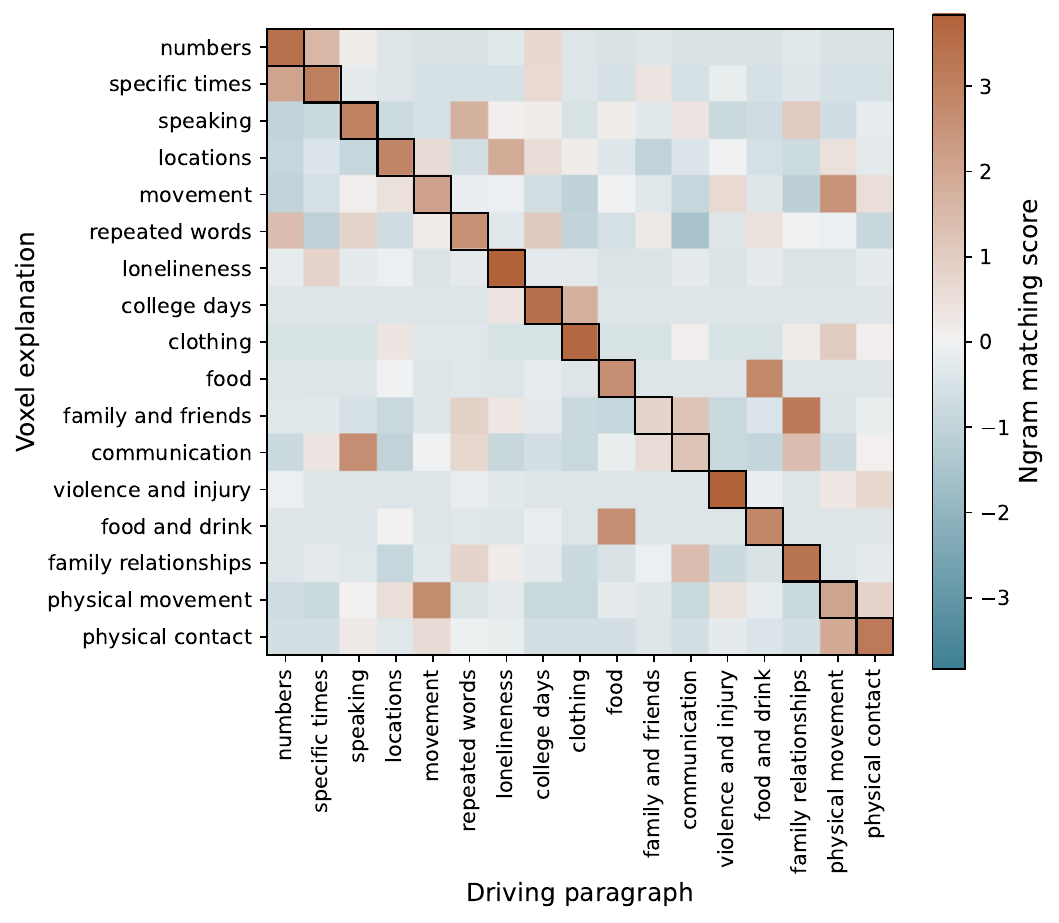}
 & \includegraphics[width=0.48\textwidth]{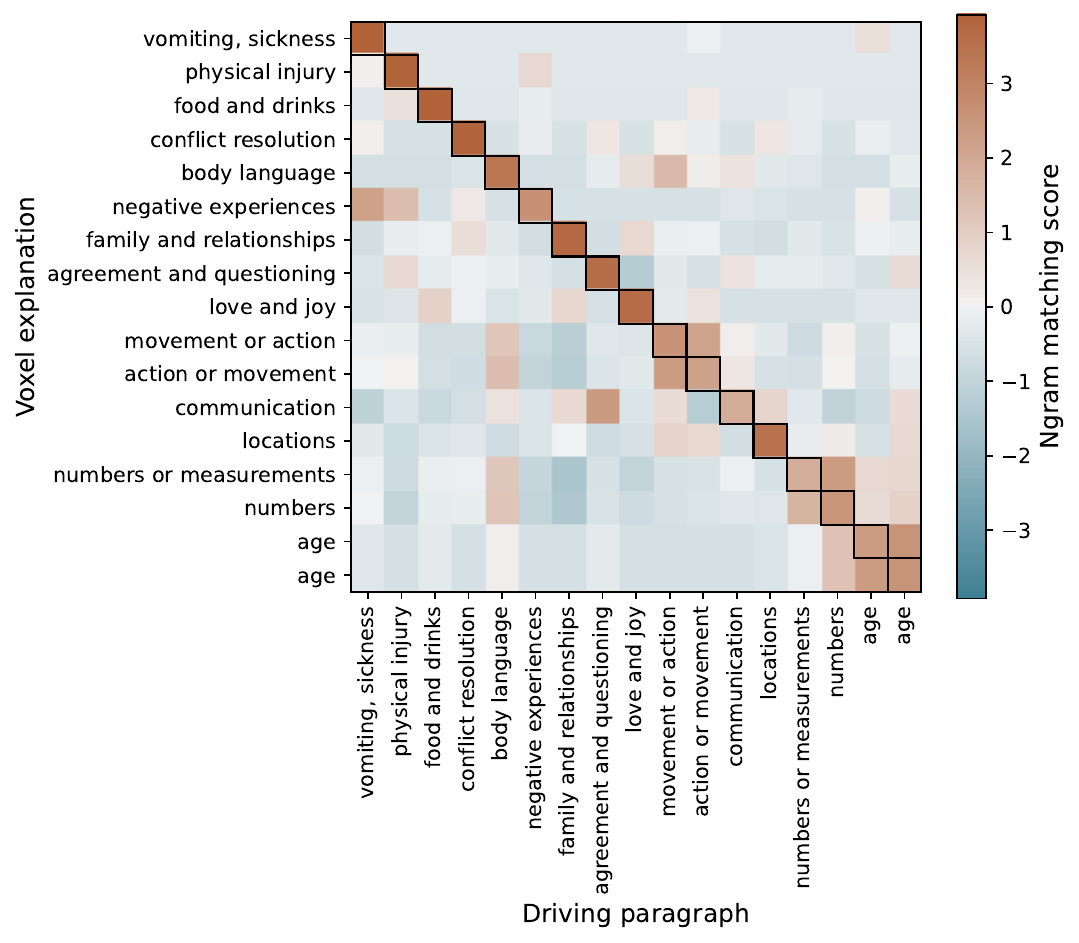}\\
    \end{tabular}
    \caption{To confirm that generated paragraphs match the explanation used to construct them, a matching score was computed for each explanation and paragraph by using an LLM to evaluate the fraction of trigrams in the paragraph that are relevant to the paragraph’s generating explanation and then z-scoring the result. Each driving paragraph showed a strong match with its generating explanation. Plot shows subject S01 and S03; plot for subject S02 shown in \cref{fig:fig4}a.}
    \label{fig:data_heatmap_avgs}
\end{figure}

\begin{figure}
    \centering
    \begin{tabular}{cc}
         \;\;\;\;\;\;\; S01 & \;\;\;\;\;\;\;\;\;\;\;\; S03 \\
            \includegraphics[width=0.45\textwidth]{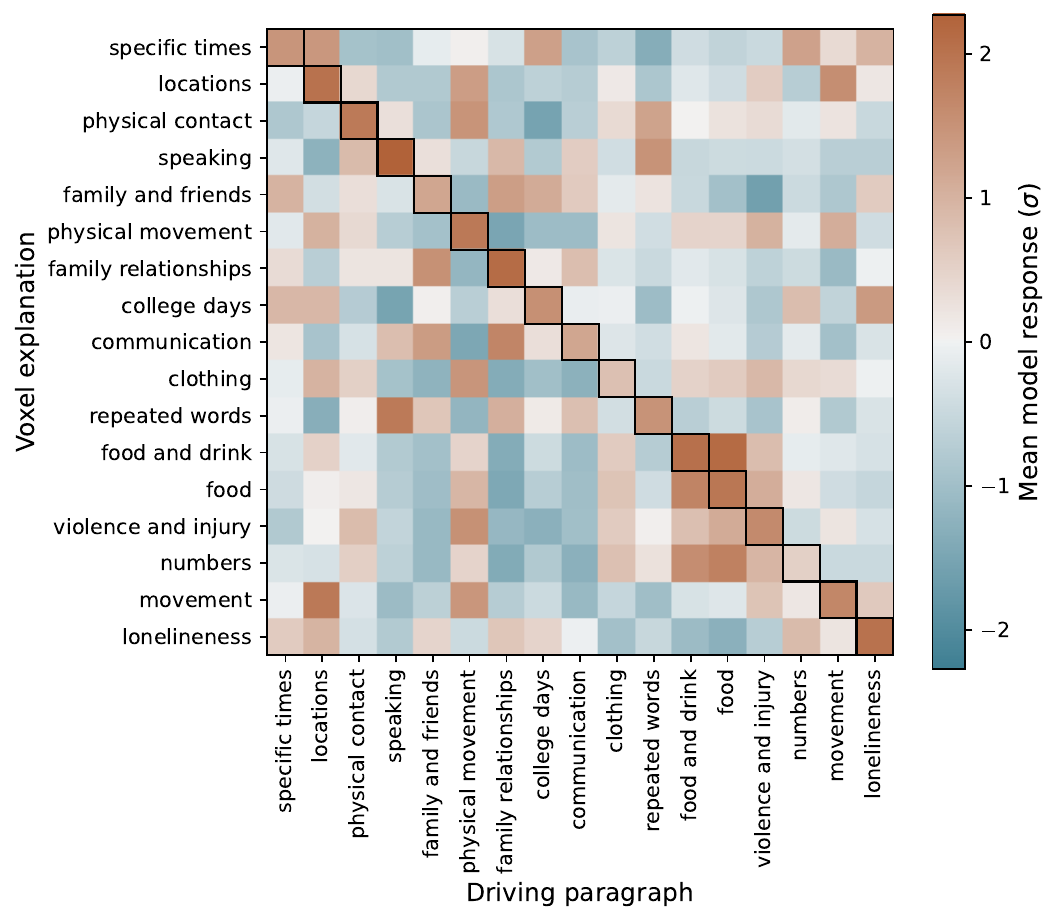}
 & \includegraphics[width=0.45\textwidth]{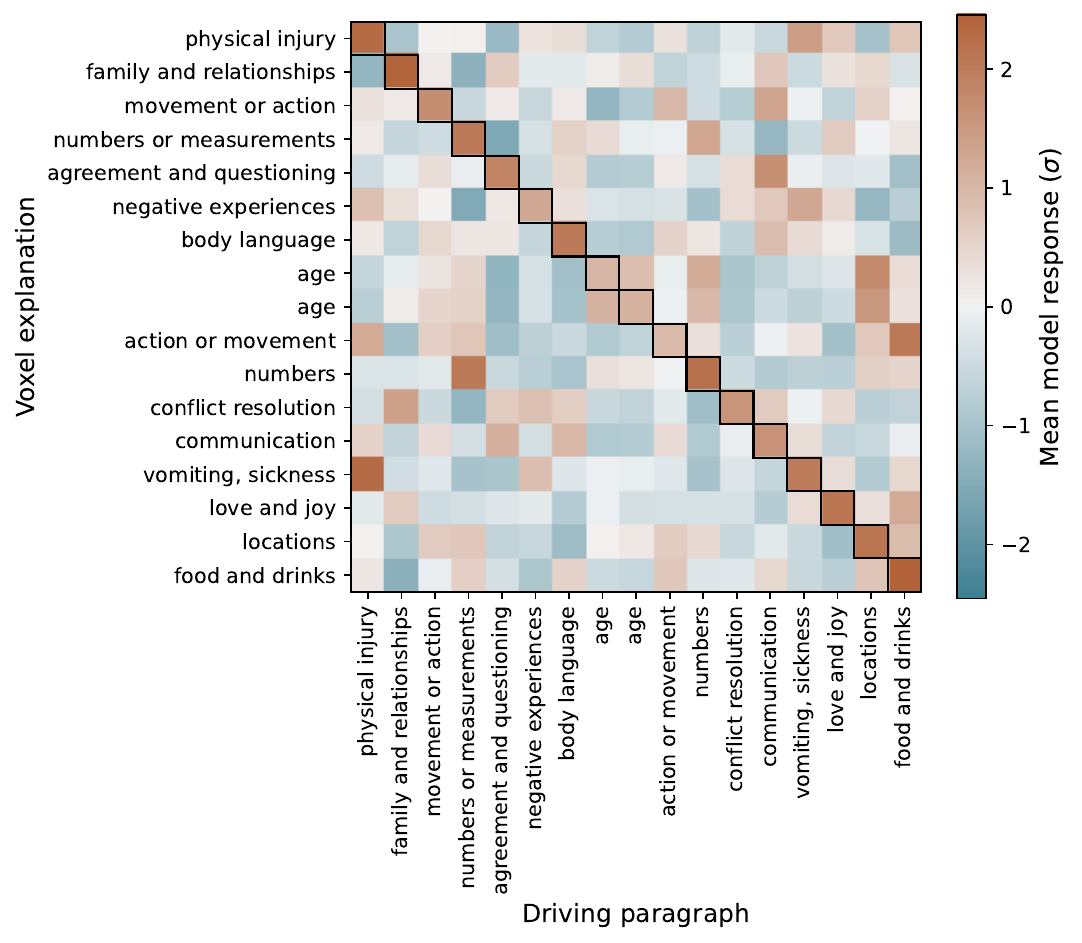}\\
    \end{tabular}
    \caption{To confirm that each driving paragraph effectively drives its corresponding encoding model, we computed the predicted response in each selected voxel to each generated paragraph. This revealed strong matches for most voxels, along with some matches between driving paragraphs and voxels with semantically similar explanations, e.g. \textit{directions} and \textit{locations}. Plot shows subject S01 and S03, plot for subject S02 shown in \cref{fig:fig4}b.}
    \label{fig:encoding_heatmap_avgs}
\end{figure}

\FloatBarrier

\subsubsection{Noise ceilings for single-voxel driving}
\label{tab:noise_ceilings}

We attempted to estimate a ceiling on how well driving could be performed in principle using our approach. Our noise ceiling approach uses the encoding model predictions as a baseline. These predictions are characteristically more extreme than driving performance actual data due to the heavy feature loading our explanation-based approach entails, which drives the encoding model partly out-of-distribution. We then scale these predictions according to a split-half reliability estimate and clip extreme predictions to within biologically plausible ranges. The long prediction tail we see in Section \ref{app:distribution} justifies this prediction-clipping approach. 

Recall that the driving value of our voxels is computed as the difference between the average response in the given voxel in the driving paragraph against all other paragraphs.

  $$d_v = \text{mat}[v,v] - \frac{1}{n-1}\sum_{j \neq v}
  \text{mat}[v,j] \label{eq:driving}$$ \\[4pt]

To attempt to determine how much of the signal that the encoding model predicts is actually attainable, we compute the noise-corrected encoding model accuracy as 
  
  $$r_{\text{corrected},v} =
  \frac{r_{\text{EM},v}}{\sqrt{r_{\text{single},v}}} $$\label{eq:rcorr}
  \\[4pt]

where $r_{\text{single},v}$ is the single-session TR-level split-half reliability for voxel $v$ computed over ten repeats of a standard naturalistic story from our original test set. As our final ceiling on driving performance, we then compute:
  
  $$d_{\text{ceiling},v}(c) = r_{\text{corrected},v} \cdot
  d_{\text{pred},v}(c)$$ \label{eq:ceiling}

  where $d_{\text{pred},v}(c)$ is computed from EM predictions clipped
   to $[-c,\, c]$ at the TR level
  before paragraph averaging. The clipping threshold $c$ is the sole
  free parameter;
  we report results for $c \in \{2\sigma, 3\sigma, 4\sigma\}$.

  \begin{table}[h]
    \caption{Per-voxel noise ceiling analysis. All driving scores in
    units of $\sigma$ (z-scored fMRI). Results below are presented for UTS02 as it is the only subject with 6 experiments per explanation.
    $r_{\text{EM}}$: TR-level correlation between EM prediction and fMRI
     on driving data.
    $r_{\text{s}}$: single-session reliability (from 10 repeats of a
    natural story).
    $r_{\text{c}} = r_{\text{EM}}/\sqrt{r_{\text{s}}}$: noise-corrected
    EM accuracy.
    $d_{\text{pred}}$: EM-predicted driving score (unclipped).
    $d_{\text{p}}(c)$: EM driving score after clipping predictions to
    $[-c,c]$.
    $d_{\text{ceil}}(c) = r_{\text{c}} \cdot d_{\text{p}}(c)$: combined
    ceiling.
    $d_{\text{act}}$: observed driving score.
    The clipping threshold $c$ is the sole free parameter. $d_{act}$ is correlated with $d_{pred}$ ($r=0.67)$ showing that voxels where the EM predicts a larger driving score tend to actually show larger driving scores. Moreover, $d_{act}$ is also correlated with $r_{EM}$ ($r=0.5$) showing that voxels where the encoding model is more accurate on driving data also tend to drive better.}
    \label{tab:noise_ceiling}
    \centering
    \footnotesize
    \setlength{\tabcolsep}{3.5pt}
    \begin{tabular}{l r r r r r r r r r r r}
    \hline
     & & & & & \multicolumn{2}{c}{$c = 2\sigma$} &
    \multicolumn{2}{c}{$c = 3\sigma$} & \multicolumn{2}{c}{$c =
    4\sigma$} & \\
    \cmidrule(lr){6-7} \cmidrule(lr){8-9} \cmidrule(lr){10-11}
    Explanation & $r_{\text{EM}}$
     & $r_{\text{s}}$ & $r_{\text{c}}$ & $d_{\text{pred}}$
      & $d_{\text{p}}$ & $d_{\text{ceil}}$
      & $d_{\text{p}}$ & $d_{\text{ceil}}$
      & $d_{\text{p}}$ & $d_{\text{ceil}}$
      & $d_{\text{act}}$ \\
    \hline
    emotion              & $0.27$ & $0.19$ & $0.62$ & $0.41$ &
     $0.27$ & $0.16$ & $0.34$ & $0.21$ & $0.39$ & $0.24$ & $-0.11$ \\
    surprise             & $0.19$ & $0.17$ & $0.47$ & $1.38$ &
     $1.00$ & $0.47$ & $1.27$ & $0.60$ & $1.38$ & $0.65$ & $-0.14$ \\
    rejection            & $0.11$ & $0.08$ & $0.38$ & $0.99$ &
     $0.38$ & $0.15$ & $0.68$ & $0.26$ & $0.89$ & $0.34$ & $ 0.16$ \\
    laughter             & $0.24$ & $0.19$ & $0.55$ & $1.25$ &
     $0.85$ & $0.47$ & $1.12$ & $0.62$ & $1.25$ & $0.69$ & $ 0.19$ \\
    food preparation     & $0.32$ & $0.08$ & $1.17$ & $1.61$ &
     $1.06$ & $1.24$ & $1.40$ & $1.64$ & $1.60$ & $1.87$ & $ 0.60$ \\
    time                 & $0.08$ & $0.17$ & $0.18$ & $2.25$ &
     $1.73$ & $0.32$ & $2.07$ & $0.38$ & $2.21$ & $0.41$ & $ 0.16$ \\
    birthdays            & $0.08$ & $0.07$ & $0.27$ & $0.77$ &
     $0.76$ & $0.21$ & $0.77$ & $0.21$ & $0.77$ & $0.21$ & $ 0.26$ \\
    negativity           & $0.30$ & $0.22$ & $0.65$ & $1.02$ &
     $0.05$ & $0.03$ & $0.19$ & $0.12$ & $0.48$ & $0.31$ & $ 0.20$ \\
    emotional expression & $0.37$ & $0.23$ & $0.78$ & $1.94$ &
     $0.89$ & $0.69$ & $1.31$ & $1.02$ & $1.61$ & $1.25$ & $ 0.41$ \\
    death                & $0.30$ & $0.13$ & $0.83$ & $0.33$ &
     $0.05$ & $0.04$ & $0.04$ & $0.04$ & $0.12$ & $0.10$ & $ 0.06$ \\
    moments              & $0.34$ & $0.21$ & $0.74$ & $1.84$ &
     $1.15$ & $0.85$ & $1.45$ & $1.07$ & $1.62$ & $1.19$ & $ 0.37$ \\
    injury/trauma        & $0.18$ & $0.12$ & $0.54$ & $1.62$ &
     $1.21$ & $0.66$ & $1.47$ & $0.80$ & $1.59$ & $0.86$ & $ 0.35$ \\
    measurements         & $0.31$ & $0.19$ & $0.71$ & $3.43$ &
     $1.71$ & $1.20$ & $2.24$ & $1.58$ & $2.67$ & $1.88$ & $ 0.79$ \\
    communication        & $0.33$ & $0.07$ & $1.22$ & $1.33$ &
     $0.52$ & $0.64$ & $0.88$ & $1.08$ & $1.17$ & $1.43$ & $ 0.51$ \\
    directions       & $0.14$ & $0.12$ & $0.41$ & $1.97$ &
     $1.36$ & $0.56$ & $1.73$ & $0.72$ & $1.88$ & $0.78$ & $ 0.28$ \\
    hair \& clothing     & $0.08$ & $0.06$ & $0.34$ & $1.30$ &
     $1.28$ & $0.44$ & $1.30$ & $0.45$ & $1.30$ & $0.45$ & $-0.10$ \\
    locations        & $0.38$ & $0.21$ & $0.83$ & $1.90$ &
     $1.08$ & $0.90$ & $1.47$ & $1.23$ & $1.70$ & $1.42$ & $ 0.46$ \\
    \hline
    \textbf{Mean}        &
    $\mathbf{0.24}$ & $\mathbf{0.15}$ & $\mathbf{0.63}$ & $\mathbf{1.49}$
      & $\mathbf{0.90}$ & $\mathbf{0.53}$
      & $\mathbf{1.16}$ & $\mathbf{0.71}$
      & $\mathbf{1.33}$ & $\mathbf{0.83}$
      & $\mathbf{0.26}$ \\
    \hline
    \end{tabular}
    \end{table}

  \begin{table}[h]
  \caption{Sensitivity of the noise ceiling to the clipping threshold
  (mean across 17 voxels). As the EM predictions are frequently far out of distribution (in excess of 10sigma) we clip them to a chosen value $c$. The choice of $c$ varies our noise ceiling.}
  \label{tab:nc_sensitivity}
  \centering
  \begin{tabular}{r l r r r}
  \hline
  Clip $c$ & Justification & $d_{\text{pred}}(c)$ &
  $d_{\text{ceiling}}(c)$ & $d_{\text{act}} / d_{\text{ceil}}$ \\
  \hline
  $2.0\sigma$ & fMRI 99th percentile & $0.902\sigma$ & $0.530\sigma$ &
  $49\%$ \\
  $3.0\sigma$ & fMRI 99.8th percentile & $1.160\sigma$ & $0.706\sigma$ &
  $37\%$ \\
  $4.0\sigma$ & fMRI max (single session) & $1.330\sigma$ &
  $0.827\sigma$ & $32\%$ \\
  $\infty$    & No clipping & $1.491\sigma$ & $0.945\sigma$ & $28\%$
  \\
  \hline
  \end{tabular}
  \end{table}

\FloatBarrier

\newpage

\subsection{Driving method variations}

\label{sec:driving_vars}

In this paper, we propose \method{} as a way to test the driving power of explanations, using stories generated by prompting an LLM with the explanations.
This approach is useful in that it allows testing an explanation, but removing this constraint allows for alternative methods to generate a driving stimulus exist.
In this section, we explore alternative methods for driving a brain region given a fitted encoding model for that brain region.

\subsubsection{Representation-engineering based driving}
\label{sec:repr_steering}

To drive our encoding models without explanatory constraints,
we first consider a representation engineering approach~\cite{zou2023representation,dathathri2019plug}.
This approach works by modifying the inference procedure for text generation: before generating a token, this approach pushes the hidden representation along the gradient of a steering model (in our case, along the linear direction in hidden space provided by an encoding model's weights).
Intuitively, this change should help create generations that drive the steering model.

To run representation engineering, we require an encoding model which uses hidden representations from the same model that will be generating the stimulus.
For this purpose, we train an encoding model based on hidden embeddings from LLaMA-3 70B~\cite{grattafiori2024llama}, which has shown stronger generation performance than the encoding models used in the main text (LLaMA-1 and OPT) and is open source, allowing us to extract hidden embeddings from it.
For each subject, we swept over 5 layers from LLaMA-3 70B (\textit{meta-llama/Meta-Llama-3-70B}, layers 6, 12, 18, 24, 30).
We train each encoding model on a sliding 10-gram window, which has been shown to achieve strong predictive performance~\cite{benara2024crafting} and the use of a relatively short context can help improve steering.
We select the best layer using cross-validation and then report its test performance across
three test stories: \textit{Where There's Smoke} (average of 10 repeats), \textit{On Approach To Pluto} (average of 5 repeats), and \textit{From Boyhood To Fatherhood} (average of 5 repeats).
For subject S02, the selected layer was 36 and the mean test correlation across voxels was 0.106.
For subject S03, the selected layer was layer 24 and the mean test correlation was 0.145.

\begin{figure}[H]
    \centering
    \begin{tabular}{cc}
    \includegraphics[width=0.4\linewidth]{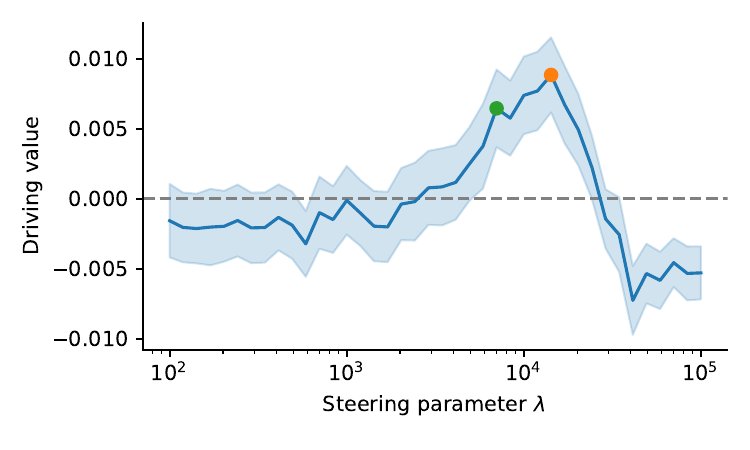} &
    \includegraphics[width=0.4\linewidth]{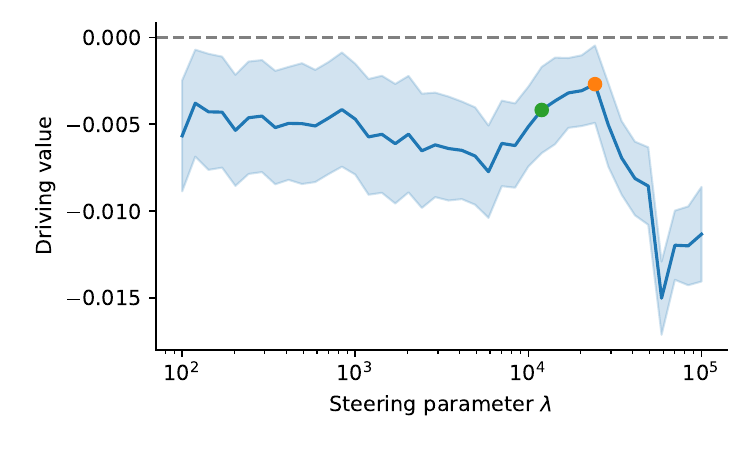}\\
    \includegraphics[width=0.4\linewidth]{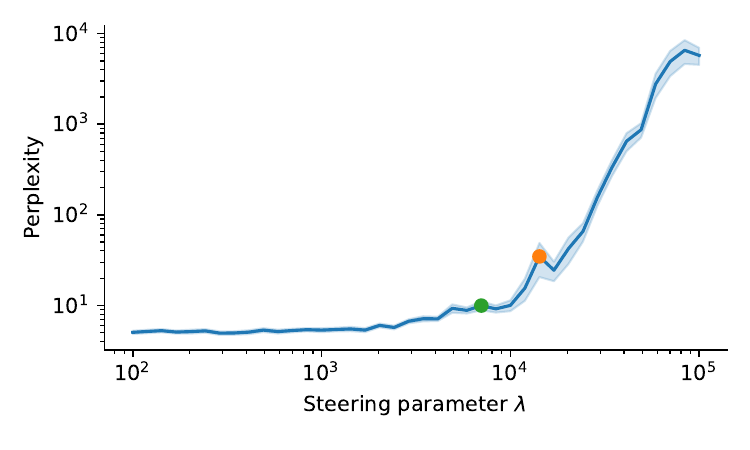} &
    \includegraphics[width=0.4\linewidth]{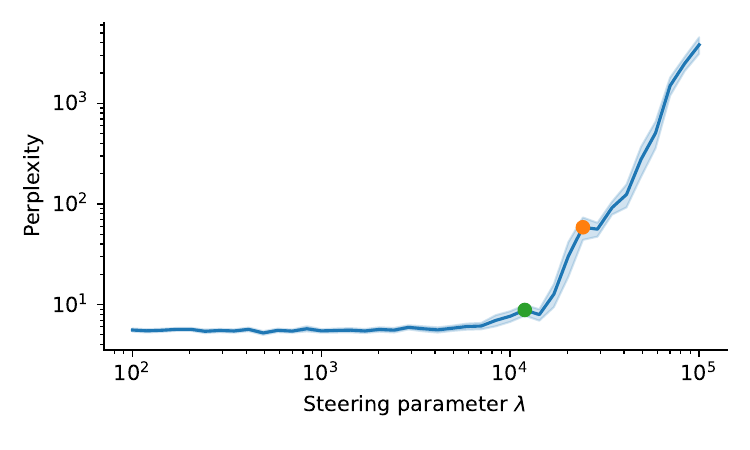}\\
    S02 & S03\\
    \end{tabular}
    \caption{\textbf{Tuning steering hyperparameter for representation steering} for subject S02 (left) and S03 (right).
    Driving value increases as steering parameter increases, but then decreases as text sampling degenerates.
    This can be seen clearly as the perplexity (bottom row) increases along with the steering parameter.
    Orange point shows parameter which maximizes the driving value, and green point shows a point selected to balance driving value and perplexity.
    Averaged over three randomly generated paragraphs for each of the 8 ROIs.
    Error bars show standard error of the mean.
    }
    \label{fig:repr_steering_tuning}
\end{figure}

Given this encoding model, we now leverage it to generate driving paragraphs via representation-engineering.
This requires selecting a hyperparameter that controls the amount of steering applied, with higher amounts generally increasing the driving of the encoding model.
However more steering also moves the LLM's hidden representation more out of distribution, potentially removing the LLM's ability to generate text.
For example, at very high values of the steering parameter, the model often devolves into repeatedly outputting the same token.
We sweep over 40 log-spaced parameters from $10^2$ to $10^5$ and apply representation engineering using encoding models for the 8 ROIs shown in \cref{fig:fig2}a, prompting the model to generate the first paragraph of a story (and limiting it to generating 100 tokens) for subject S02 and S03.
We evaluate the ability of the generated stories to then drive the main encoding model used throughout the experiments in our paper (based on LLaMA-1, see \hyperref[sec:methods]{Methods}).
\cref{fig:repr_steering_tuning} shows that when tuned properly, representation engineering succeeds in significantly driving encoding model response for subject S02 but fails for S03 (in that case overfitting to peculiarities in the LLaMA-3 70B used during representation engineering).

\begin{figure}[H]
    \centering
    \includegraphics[width=0.9\linewidth]{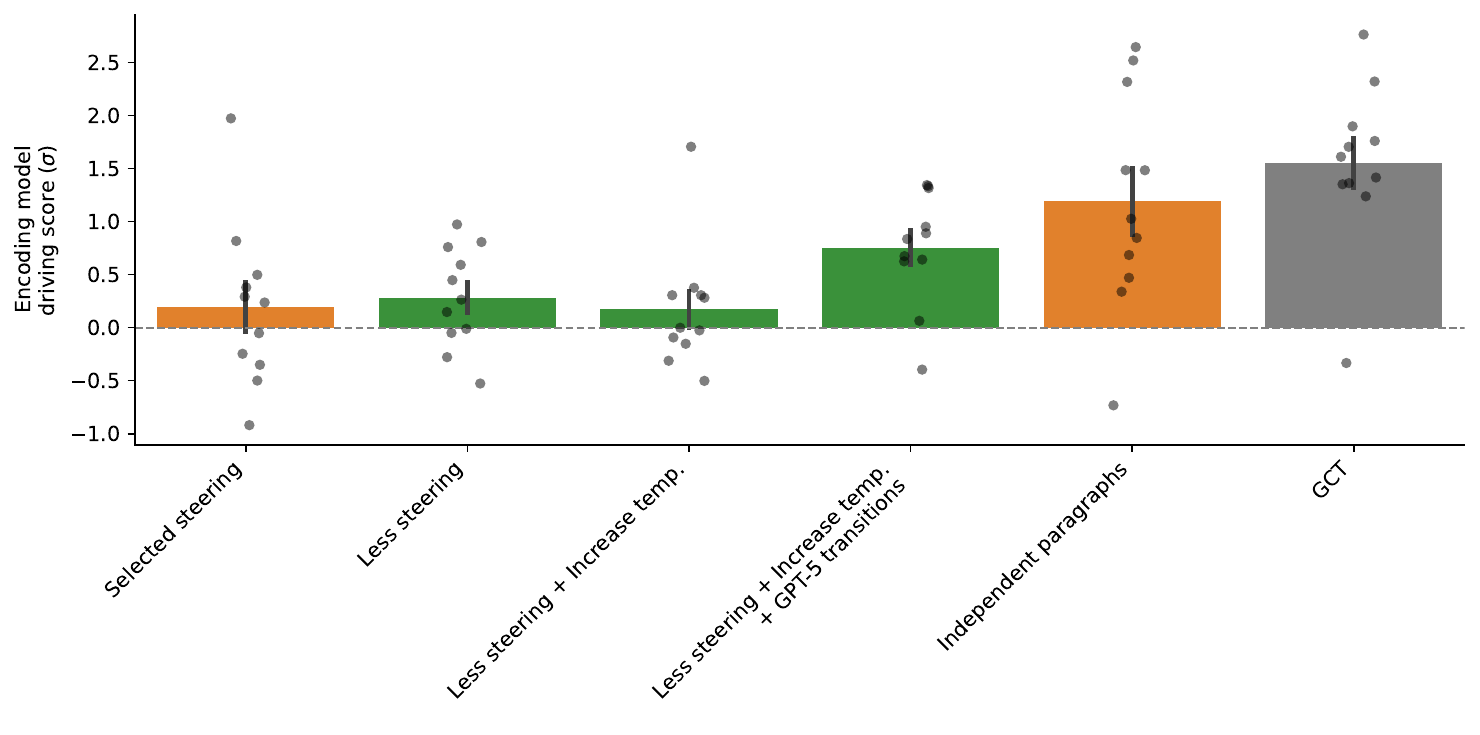}
    \caption{\textbf{Driving score for encoding, when generating stories for different variations} for subject S02.
    \method{} achieves a higher driving score than representation-steering based methods.
    Encoding model driving score measures the driving of the LLaMA-1 encoding model used throughout the paper (see \hyperref[sec:methods]{Methods}).
    Results are averaged over three randomly generated stories for the 11 ROI settings shown in \cref{fig:fig2}a.
    Error bars show standard error of the mean.
    }
    \label{fig:repr_steering_variations}
\end{figure}

Having selected a hyperparameter that successfully generates individual paragraphs that drive the S02 encoding model, we now seek to generate driving stories, i.e. stories where each paragraph  alternates between driving the encoding model for each of the 11 ROI settings shown in \cref{fig:fig2}a.
\cref{fig:repr_steering_variations} shows the encoding model driving score when generating stories with different method variations.

We initially try selecting the hyperparameter $\lambda$ that maximizes the driving value for individual paragraphs (\cref{fig:repr_steering_tuning} orange point), but find that it does not yield a strong driving score.
Different variations, such as decreasing the steering parameter (to the green point in \cref{fig:repr_steering_tuning}) or increasing the sampling temperature (from 0.8 to 1.2) yield little improvement.
One variation that yields reasonable stories prompts GPT-5~\cite{singh2025openai} to complete the end of a paragraph with the prompt ``\textit{Return one and a half concluding sentences for the text. The concluding sentences should start a story, rather than being repetitive.}''
This helps avoid overly repetitive stories and improves encoding model driving.
Nevertheless, this hybrid strategy yields considerably worse driving performance than independent paragraphs, which in turn yields considerably worse driving performance than \method.
This suggests that representation-engineering based driving struggles to achieve the same driving performance as \method{} for this set of targets.

\subsubsection{Stimulus selection-based checkerboard driving}
\label{sec:appendix_gpt1_checkerboard}

\begin{figure}[H]
    \centering
    \makebox[\textwidth]{
    \includegraphics[width=1.3\textwidth]
    {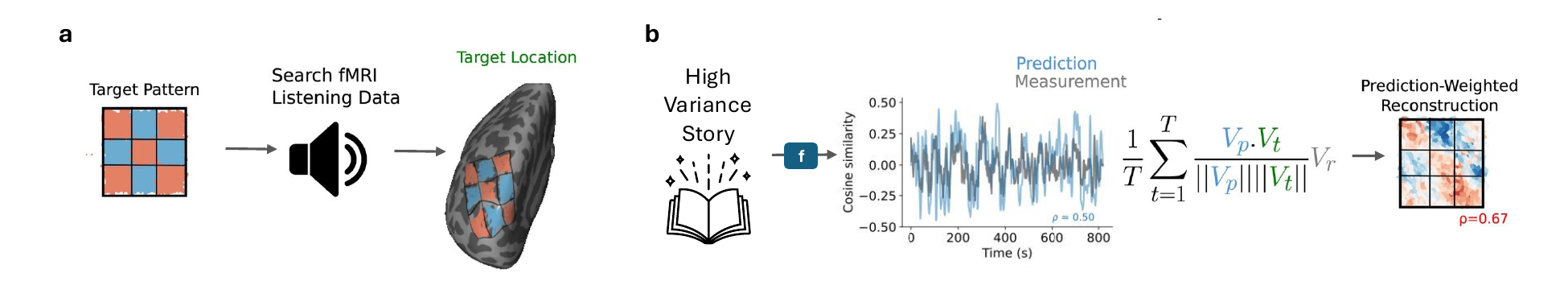}
    }
    \caption{(a) A suitable location for a target checkerboard pattern is chosen from prefrontal cortex by cross-referencing the target pattern with a pre-generated fMRI passive listening dataset.
    (b) A subject listens to a story that elicits high variance for the checkerboard pattern.
    We reconstruct the target pattern from the data by weighting the response at each timepoint by its predicted similarity to the target pattern.
    The reconstruction partially recovers the target pattern, achieving reasonably high correlation with the target pattern.}
    \label{fig:checkerboard}
\end{figure}

To test the limits of our voxel-driving paradigm, we further examine whether they can be used directly to produce arbitrary activation patterns if explanatory considerations were dropped. The experimental pipeline for this test is shown in \cref{fig:checkerboard}a.
We select a location on the cortical surface to drive by searching our training dataset for areas that contain at least one response from our training dataset with a high cosine similarity to a target checkerboard pattern. We choose a checkerboard pattern for this experiment to demonstrate that driving can be achieved even for relatively complex targets, so long as the target is biologically possible.

We demonstrate that we can partially manifest a checkerboard pattern in prefrontal cortex.
As the checkerboard pattern is fairly unnatural, we aim to reconstruct the checkerboard from the response pattern rather than directly drive it (\cref{fig:checkerboard}b).
We compute the sum of the responses for the voxels in the checkerboard location across a single high-variance story,
weighted by the cosine similarity of the responses to the target pattern.
This results in a pattern that bears some resemblance to the desired checkerboard (pearson correlation coefficient $\rho=0.67$).

\subsubsection{Decoding-based checkerboard driving}

\label{sec:decoding_checkerboard}

\begin{figure}[H]
    \centering
    \makebox[\textwidth]{
    \includegraphics[width=0.7\textwidth]
    {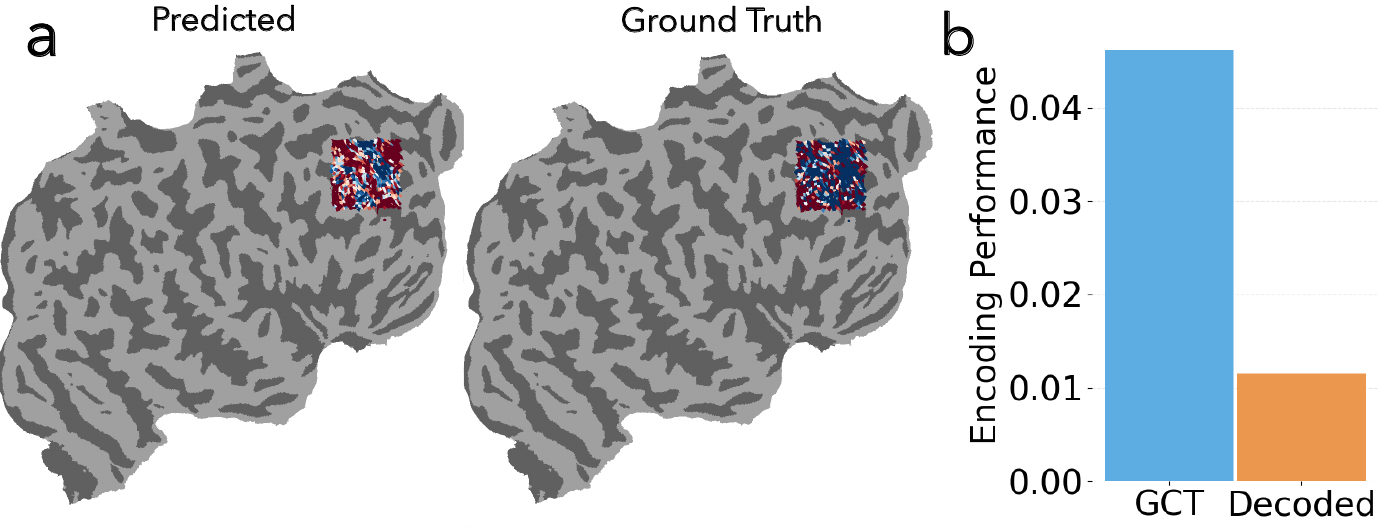}
    }
    \caption{\textbf{Driving with language decoding}  We used a beam-search language decoding algorithm designed to recover semantic content from fMRI responses to attempt to drive neural response. (a) The target response was ``decoded'' as if it were a real response and the ``decoded'' stimuli was then presented to the subject to attempt driving the target response. The result was then compared to the target. \textit{Predicted}: The predicted response of the Bayesian decoder during a story designed to drive a checkerboard pattern in prefrontal cortex. \textit{Ground Truth}: Our \textbf{best case} outcome over 4 trials of driving the checkerboard pattern. (b) Results for this driving paradigm were unreliable due to the worse quality of generated stimuli compared to the podcast training dataset and the LLM-crafted GCT stories. GCT stories show substantially higher single-trial encoding performance than the ``decoded'' stimuli. Results shown for UTS02, measured over 6 stories in GCT setting, 2 for decoded setting. Both results show the performance of an encoding model trained on the OPT-30B model, with representations from layer 33.}
    \label{fig:checkerboard2}
\end{figure}

We additionally attempted to use a decoding paradigm drive a checkerboard pattern in prefrontal cortex. This approach treated the target pattern to be driven as if it were a brain that could be directly decoded using an established language decoding algorithm~\cite{tang2023semantic}. Briefly, this method employs beam-search based combined with an LLM to generate a plausible contiguous stimuli that is maximally likely to have produced the observed response conditional on the noise covariance in the fMRI data. The ``decoded'' output was then fed into ChatGPT to improve its linguistic coherence, and then presented to the subject to see if it would produce the target response. An example excerpt of a ``story'' generated by this method is presented below, with section of the text in \textcolor{blue}{blue} if it was expected to suppress the checkerboard (drive the blue portions of the pattern from \ref{fig:checkerboard} and suppress the red portions) and \textcolor{red}{red} if it was expected to drive the checkerboard (drive the red portions of the pattern from \ref{fig:checkerboard} and suppress the blue portions):

\texttt{ \textcolor{blue}{We need to ascertain who sent a letter, or why someone would send such a letter, requesting an address to be posted but not mailed. The letter also asked for an explanation as to why I had written it and included my personal information for use as evidence.} \textcolor{red}{This occurred when I was in my early twenties, a few years before my first memory of my mother when she was just over two and a half years old.\\\\I was about five or six at the time, and it was the early twenties of a few years before my life took a turn for the worse.} \textcolor{blue}{My parents decided that I could use their credit card and asked me to send them an email explaining why I needed it...}}

This decoding-based driving method successfully produced text that periodically changed its theme in accordance with the underlying predictions of our encoding model, however, it failed to consistently produce the intended response pattern in the subject~(\cref{fig:checkerboard2}a). We hypothesize several reasons for this failure, primarily that the resultant ``stories'' were highly out-of-distribution (in contrast to the GCT stories) and difficult to parse and that the target pattern may not be a biologically plausible configuration of the BOLD response. The encoding performance of these decoded stories is far lower than the performance of the GCT stories~(\cref{fig:checkerboard2}b), giving credence to the theory that the decoded stories did not generalize well. 

\subsubsection{Selectively driving pairs of voxels through prompting}
\label{subsec:multivoxel}


Useful and explanatory scientific theories are robust to unrelated distributional shifts. For example, if a region or voxel is selective for food concepts, then that region should remain selective for food concepts even if those concepts cooccur with unrelated semantic concepts. In order to test this independence of our generated explanations, we test whether \method{} can be used to simultaneously drive pairs of voxels in different areas across the brain. 

\begin{figure}[H]
    \centering
    \makebox[\textwidth]{
    \includegraphics[width=1.25\textwidth]{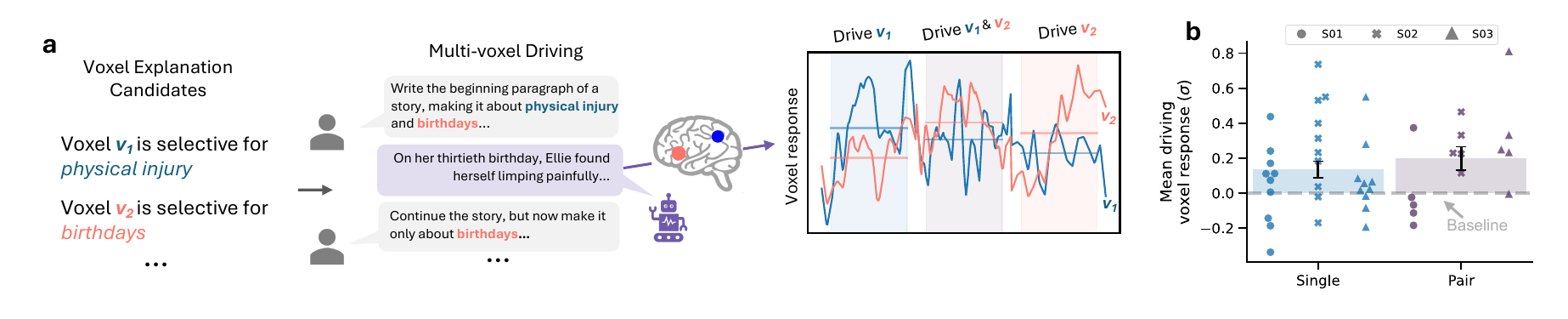}
    }
    \caption{
    \textbf{LLM stories can selectively drive fMRI responses for pairs of voxels.}
    (a) The pipeline for selectively driving a pair voxels generates explanations on a per voxel basis and prompts an LLM to generate paragraphs that either drive an individual voxel or a pair of voxels. (b) Voxel response when alternating between driving a single voxel (blue) or a pair of voxels (purple) generally succeeds in driving relative to the baseline (gray dotted line).
    }
    \label{fig:multivoxel}
\end{figure}

\cref{fig:multivoxel}a shows the multi-voxel driving pipeline. Pairs of voxels with semantically independent generated explanations are selected and stimuli are generated so that the stimuli have the following pattern:
(i) the first voxel is driven without respect to the second voxel by using the explanation for that voxel,
(ii) then both are driven simultaneously with a generated stimulus that contains both generated explanations,
and then (iii) only the second voxel is driven without respect to the first voxel. \cref{fig:multivoxel}b demonstrates that we are able to effectively combine the voxel-level explanations to drive separate voxels at the same time, or independently drive one without driving the other.
This shows that the generated explanations are resilient to unrelated semantic interventions, demonstrating that they do not work merely because they tend to semantically co-occur with stimuli that actually drives those voxels.

\subsubsection{Text-based driving vs. auditory driving}

\begin{figure}[H]
    \centering
    \makebox[\textwidth]{
    \includegraphics[width=1\textwidth]{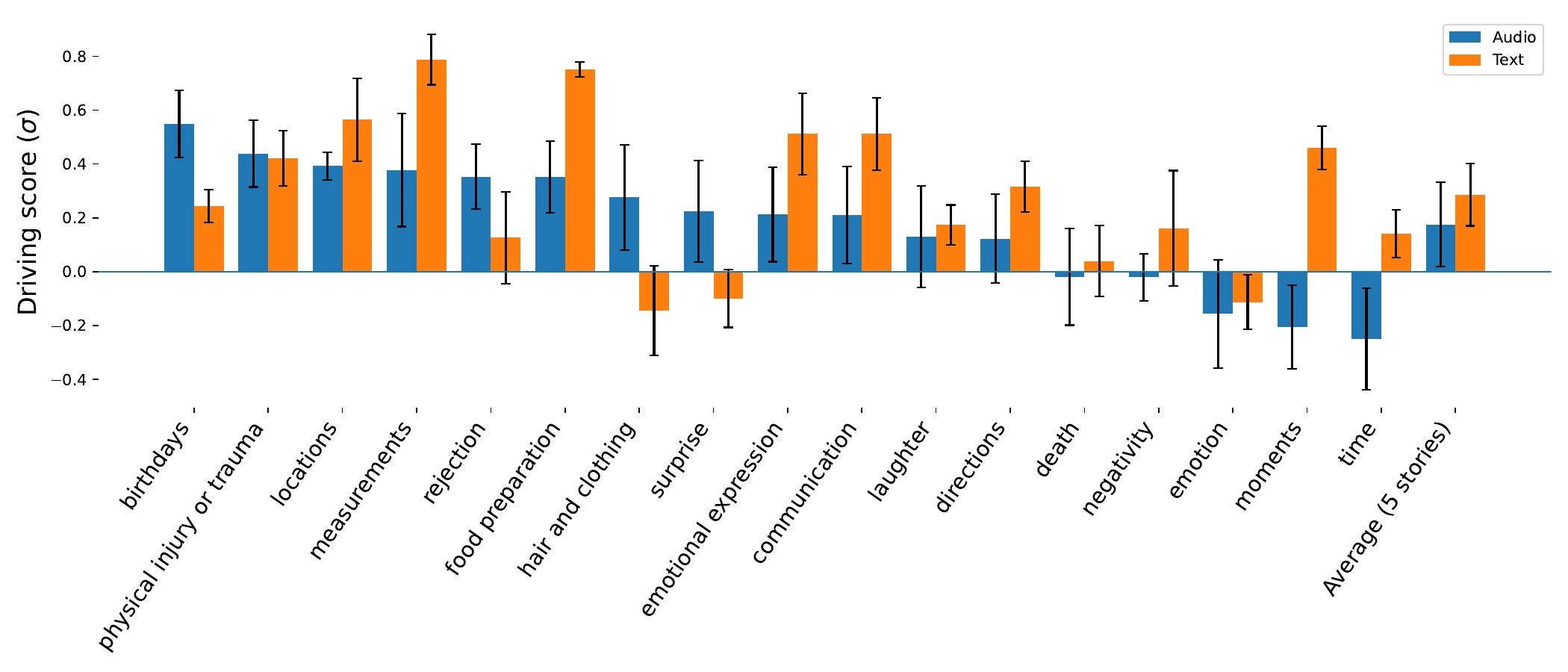}
    }
    \caption{
    \textbf{Auditory stimuli can be similarly used to drive voxelwise activation.} Bar plot shows the average driving performance across explanations, averaged over the same five stories during auditory and visual presentation in one subject (UTS02). We find that auditory stimuli are effective at driving voxelwise activation using GCT ($p < 0.01$), although less-so than visual stimuli. Error bars show standard error.
    }
    \label{fig:audiodriving}
\end{figure}

The GCT pipeline in the main text used visual stimuli (i.e. reading) as opposed to auditory stimuli to drive voxel activation. Given that our original stimuli used to train our encoding models were auditory, it is reasonable to ask what the effect of this choice is on our driving performance. We used a text-to-speech API provider (Eleven Labs) to create auditory equivalents of our stories and evaluated the driving effects on a single subject (UTS02) over 5 of the stories originally presented to that subject.
We find that driving performance is slightly worse in audition (\cref{fig:audiodriving}) although still statistically significant on average across all stimuli ($p < 0.01$, $t$-test).
One explanation for this reduced performance is that the Eleven Labs text-to-speech model is still imperfect, and sometimes generates speech that can come across as artificial, creating a distribution shift from the original speech used to train the encoding models.
\cref{fig:audiodriving} shows these results broken down over the 17 voxels for which driving was attempted (corresponding to the driving results in \cref{fig:fig1}e).

\subsection{Location-ROI extended results}

\begin{figure}[H]
    \centering
    \makebox[\textwidth]{
    \includegraphics[width=1.2\linewidth]{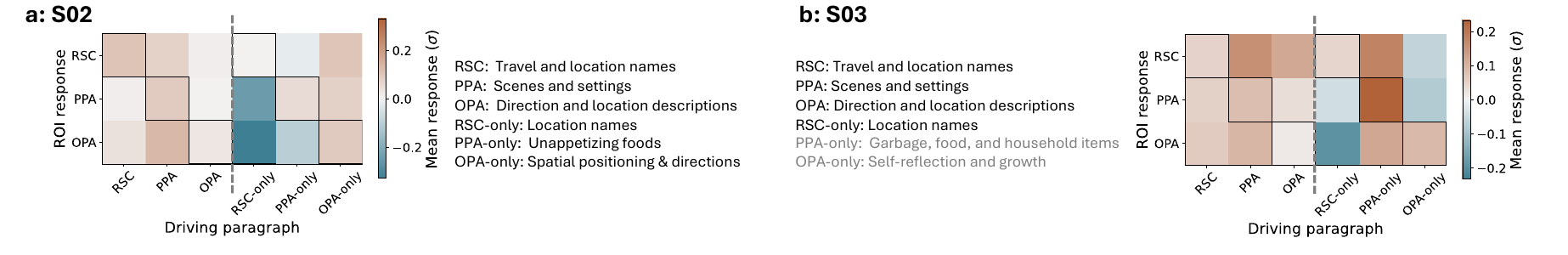}
    }
    \caption{Extended results for \cref{fig:fig2}c for subject S03 in addition to S02.
    We focused on three ROIs that are known to have similar selectivity for place concepts: retrosplenial cortex (RSC), the parahippocampal place area (PPA), and the occipital place area (OPA).
    When explanations were generated for each ROI independently we found that each ROI was driven by all three driving paragraphs (left side).
    To distinguish these ROIs, we used \method{} to find new explanations and construct stories that would selectively drive each area while suppressing the other two.
    Testing these stories in an fMRI experiment showed that in subject S02 we succeeded in finding selective explanations for two ROIs: RSC is selectively driven by \textit{location names} and PPA by \textit{unappetizing foods}. However, the explanation for OPA, \textit{spatial positioning \& directions}, drove responses in all three ROIs (right side).
    In subject S03, we succeed for all three ROIs.
    \method{} explanations are the same for 4 of the 6 ROIs between the two subjects.
    }
    \label{fig:location_rois_extended}
\end{figure}

\subsection{Language network driving extended results}
\label{sec:app_language_network}

\begin{figure}[H]
    \centering
    \caption{Driving explanations and driving scores for selectively driving the five language network regions (i.e. driving the difference between a region and the average of the others) in \cref{fig:fig3}c for S02 (left) and S03 (right).
    Diagonal elements show driving scores corresponding to the driving explanation and are generally not very positive.
    }
    \includegraphics[width=0.49\linewidth]{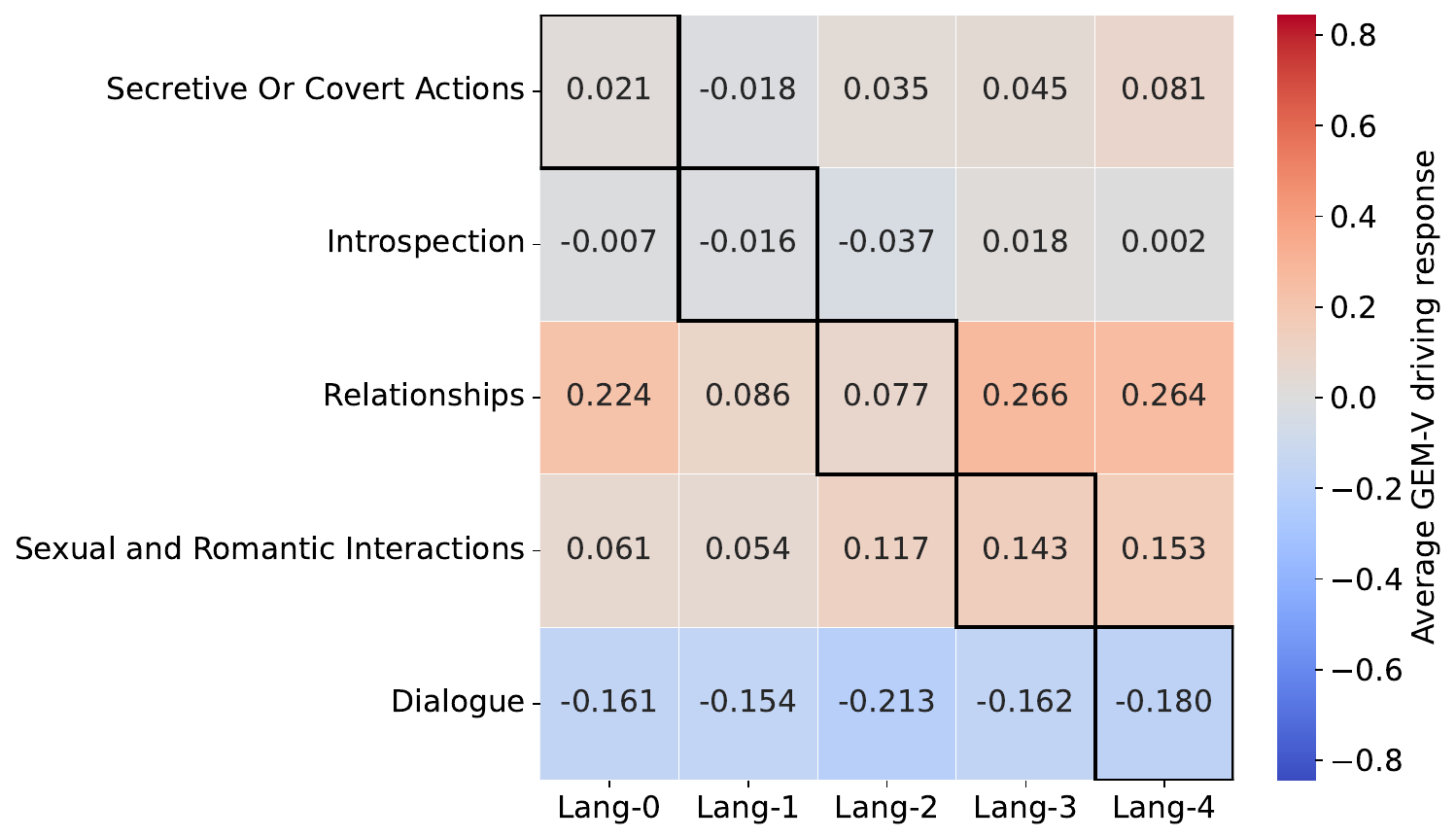}
    \includegraphics[width=0.49\linewidth]{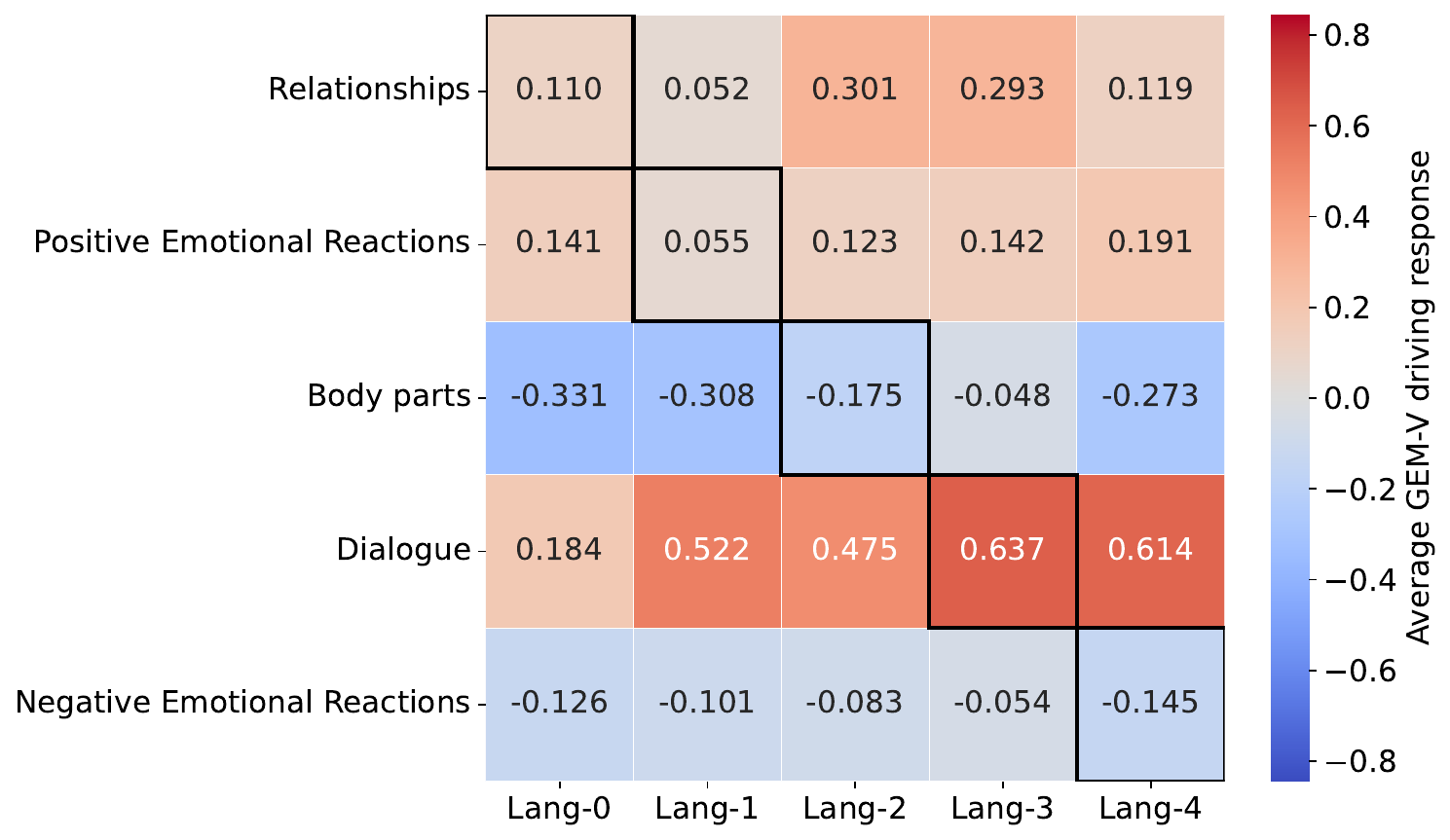}
    \label{fig:fed_drive_full}
\end{figure}

\begin{figure}[H]
    \centering
    \caption{Explanations and driving scores for the five language network regions in \cref{fig:fig3}c for S02 and S03.
    Region responses show considerable similarity within each subject, but major differences across the two subjects.
    }
    \includegraphics[width=0.95\linewidth]{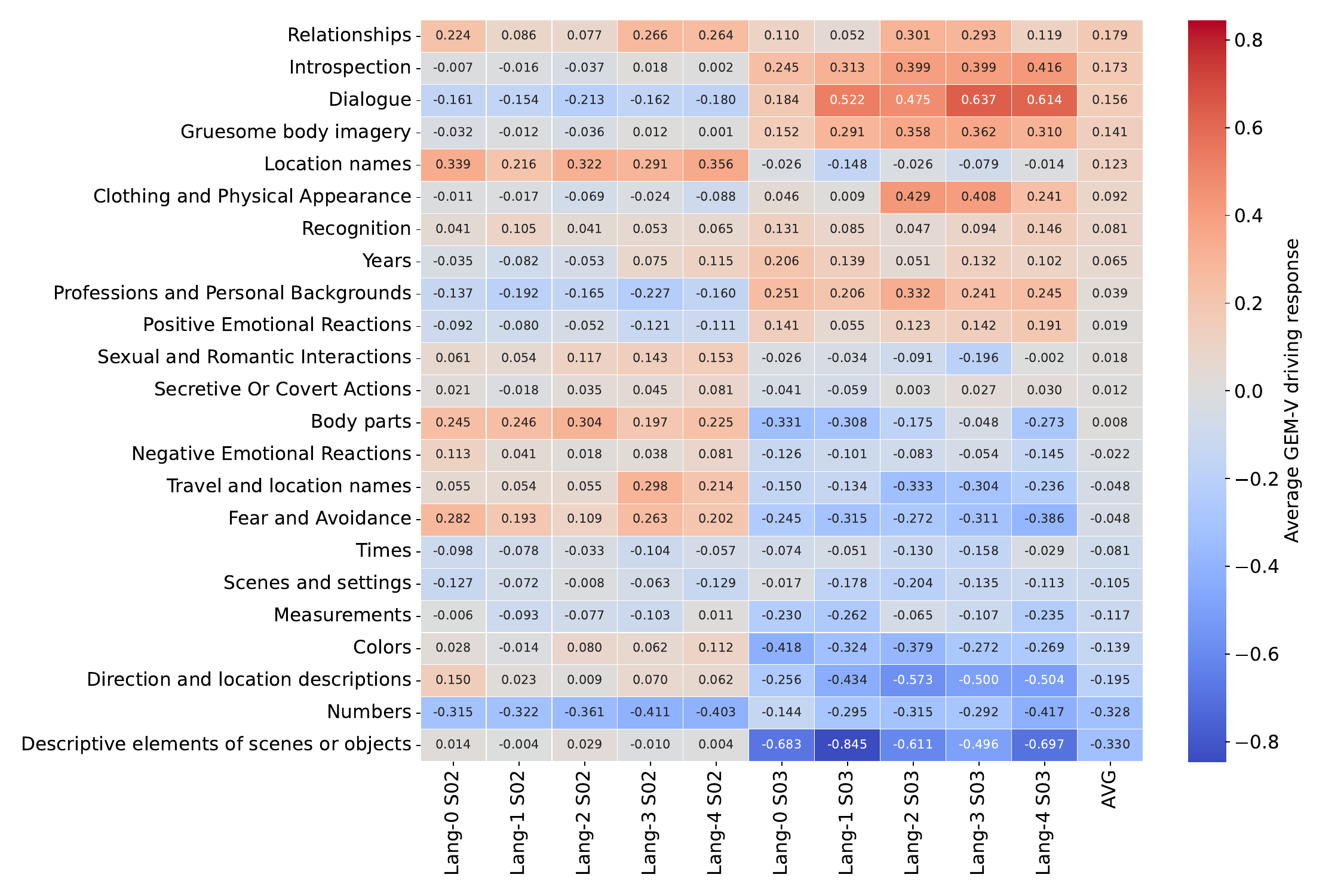}
    \label{fig:fed_drive_full2}
\end{figure}


\label{sec:appendix_topic_activation}

\subsection{Cortex-wide GCT driving patterns}

\subsubsection{Using GCT to distinguish to de-correlate explanations with high semantic similarity}
\label{subsec:cortex_wide_similar_separation}

\begin{figure}[H]
    \centering
    
    \includegraphics[width=1\textwidth]{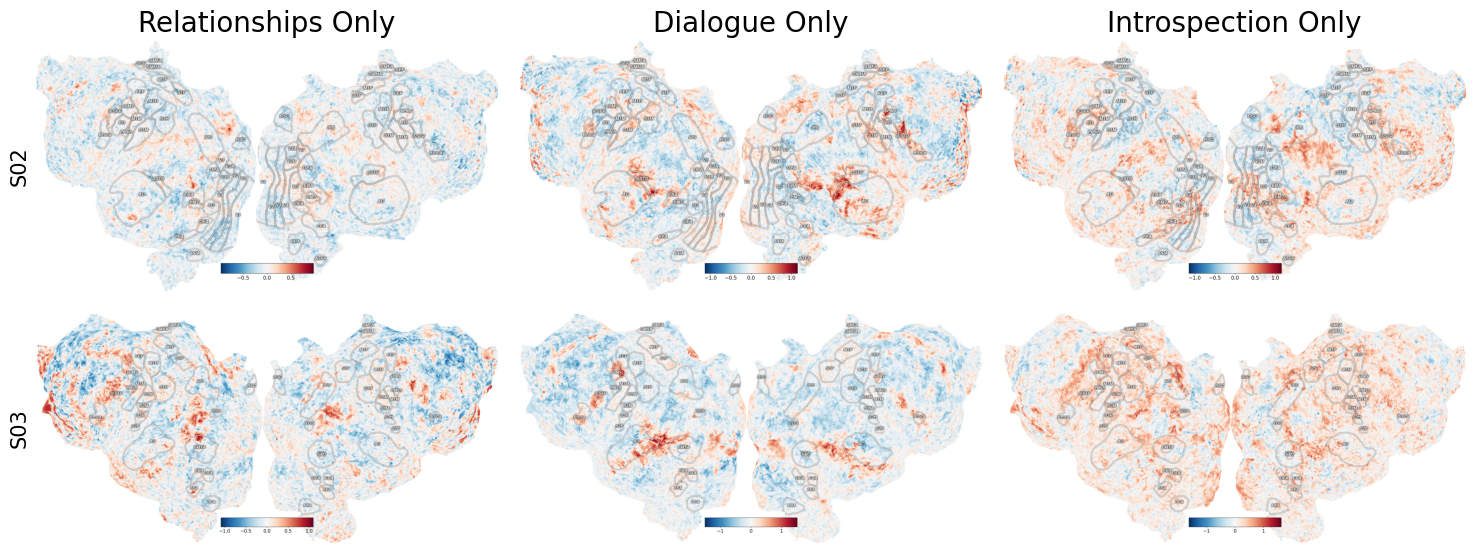}
    \caption{Response contrast maps for semantically correlated categories that have been explicitly de-correlated. The LLM was prompted to generate stories that contain mentions of, for example, text that mentions relationship n-grams but explicitly does not contain dialogue or introspection n-grams. We find that explicitly de-correlating semantics can lead to wider variability in selectivity patterns.}
    \label{fig:semcorrelates}
\end{figure}

We evaluated whether GCT could be used to de-correlate semantically correlated explanations to produce activation patterns. Specifically, in cases where two explanations were semantically correlated (see \cref{fig:fig4} and \ref{fig:data_heatmap_avgs} for further details), we attempted to see if by explicitly ablating the semantic correlation through prompting, that we could produce novel activation patterns in high-level regions such as prefrontal cortex. We took sets of concepts from our explanation set that were likely to show up simultaneously according to the results in \cref{fig:fig4} and \ref{fig:data_heatmap_avgs} and prompted an instruction-finetuned LLM to generate stories that explicitly excluded one of these concepts while retaining the others. \cref{fig:semcorrelates} shows an example comparison of this type performed between three similar explanations with cooccurrent n-gram selectivity. We find that the activation maps to these concepts are fairly consistent between subjects but different once they have been semantically decorrelated. This suggests that GCT could improve our interpretation of brain selectivity from studies that use natural language stimuli, where these kinds of semantic correlations abound.


\subsubsection{Comparing GCT explanation contrasts to Eng1000 encoding model weights.}
\label{subsec:eng1000_comparison}
\cref{tab:eng1000_comparison} shows the correlation between weights from an Eng1000 encoding model (an interpretable encoding model from a prior work~\cite{huth2016natural}) and average cortical responses to GCT paragraphs.

\begin{table*}[t]
\scriptsize
\centering
\caption{
\textbf{Correlation between Eng1000 encoding model weights and average cortical responses to GCT paragraphs.}
Eng1000 builds an encoding model feature space using co-occurence with a set of 985 words, yielding a weight for each word and voxel pair that can together be interpreted as a cortical selectivity map~\cite{huth2016natural}.
We fit the Eng1000 encoding model following the same pipeline used to fit the encoding models in the main text (see \hyperref[sec:methods]{Methods}); it achieves a test correlation of 0.101 for S03, 0.095 for S02.
For each GCT explanation used in the single-voxel driving experiments, we then find the closest match between the explanation and a word in Eng1000, shown in the table.
For S03, no close match was found for `conflict resolution', `communication', `movement or action', `vomiting, sickness'.
For S02, no close match was found for `rejection', `emotion', `emotional expression', `time', `communication'.
We then compute the correlation between the average GCT responses for an explanation and the corresponding matched Eng1000 encoding model weights.
Before computing the correlation, we mask both maps to the 10\% of voxels that are top-predicted by the Eng1000 model (to avoid noise).
The correlations are generally positive, particularly for subject S02 (for which 6 GCT stories were run, whereas only 2 were run for S03).
The average correlation for S02 was 0.2492 (p-value 0.0140; permutation test comparing against correlations between Eng1000 encoding model weights) and for S03 was 0.0945 (p-value: 0.2130).
}
\begin{minipage}{0.55\linewidth}
\centering
\begin{tabular}{llr}
\toprule
S02 GCT explanation & Eng1000 word & Correlation \\
\midrule
measurements & measure & 0.457 \\
food preparation & cook & 0.428 \\
laughter & laugh & 0.427 \\
birthdays & birthday & 0.322 \\
locations & place & 0.268 \\
directions & place & 0.258 \\
physical injury or trauma & hurt & 0.242 \\
negativity & bad & 0.233 \\
hair and clothing & clothes & 0.182 \\
moments & moment & 0.166 \\
surprise & surprise & 0.080 \\
death & die & -0.071 \\
\bottomrule
\end{tabular}
\end{minipage}
\hfill
\begin{minipage}{0.55\linewidth}
\centering
\begin{tabular}{llr}
\toprule
S03 GCT explanation & Eng1000 word & Correlation \\
\midrule
agreement and questioning & agree & 0.707 \\
locations & place & 0.527 \\
physical injury & hurt & 0.215 \\
negative experiences & trouble & 0.167 \\
food and drinks & drink & 0.136 \\
body language & body & 0.132 \\
action or movement & move & 0.097 \\
age & age & 0.045 \\
numbers or measurements & measure & -0.088 \\
love and joy & love & -0.093 \\
age & age & -0.114 \\
family and relationships & family & -0.118 \\
numbers & number & -0.385 \\
\bottomrule
\end{tabular}
\label{tab:eng1000_comparison}
\end{minipage}
\end{table*}

\newpage
\subsubsection{Cortex-wide GCT explanation contrasts}
\begin{figure}[H]
    \centering
    \includegraphics[width=1\textwidth]{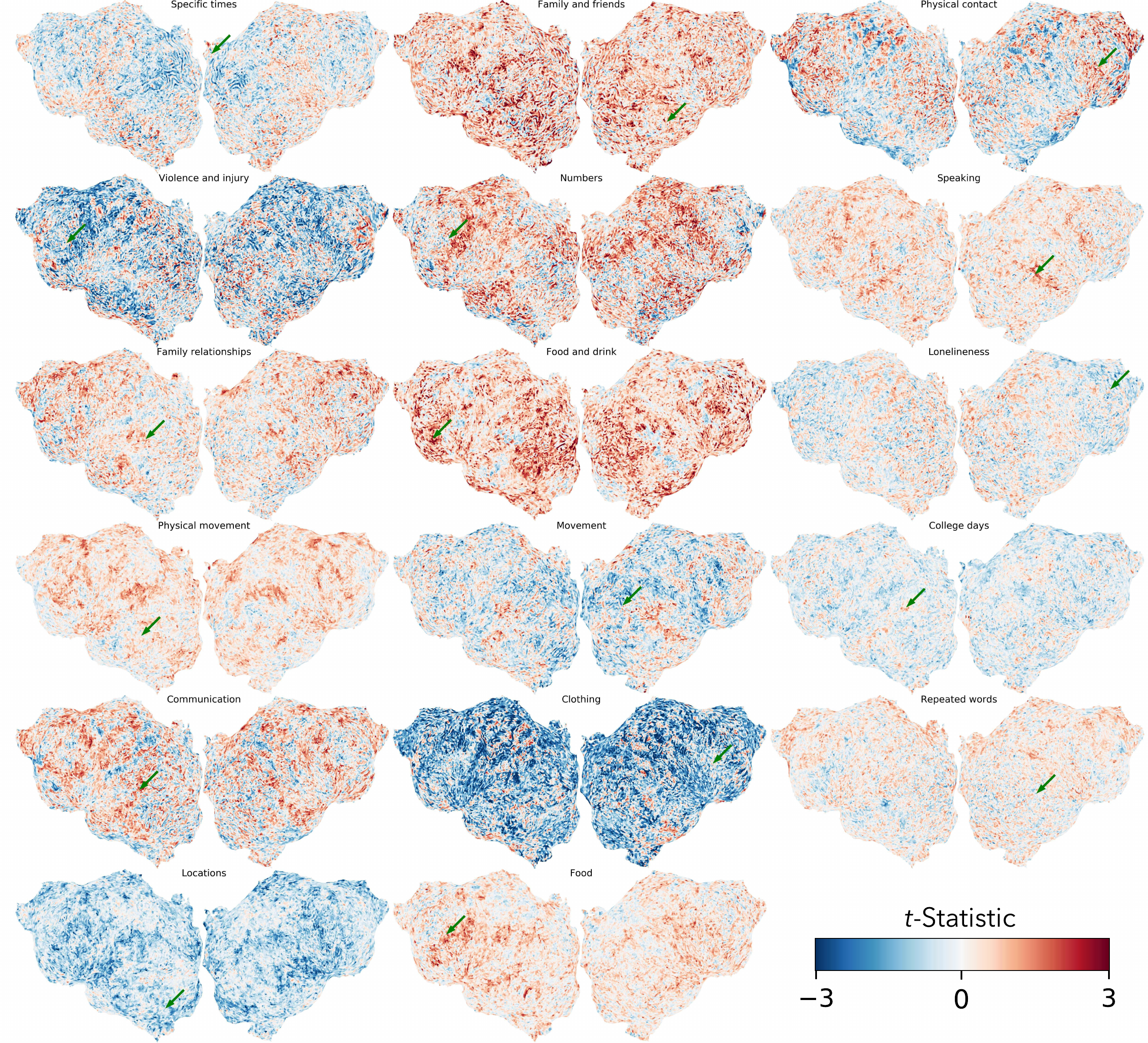}
    \caption{Explanation contrast for all tested explanations in S01. Arrows indicate source voxel where driving was attempted.}
    \label{fig:speaking}
\end{figure}

\begin{figure}[H]
    \centering
    \includegraphics[width=1\textwidth]{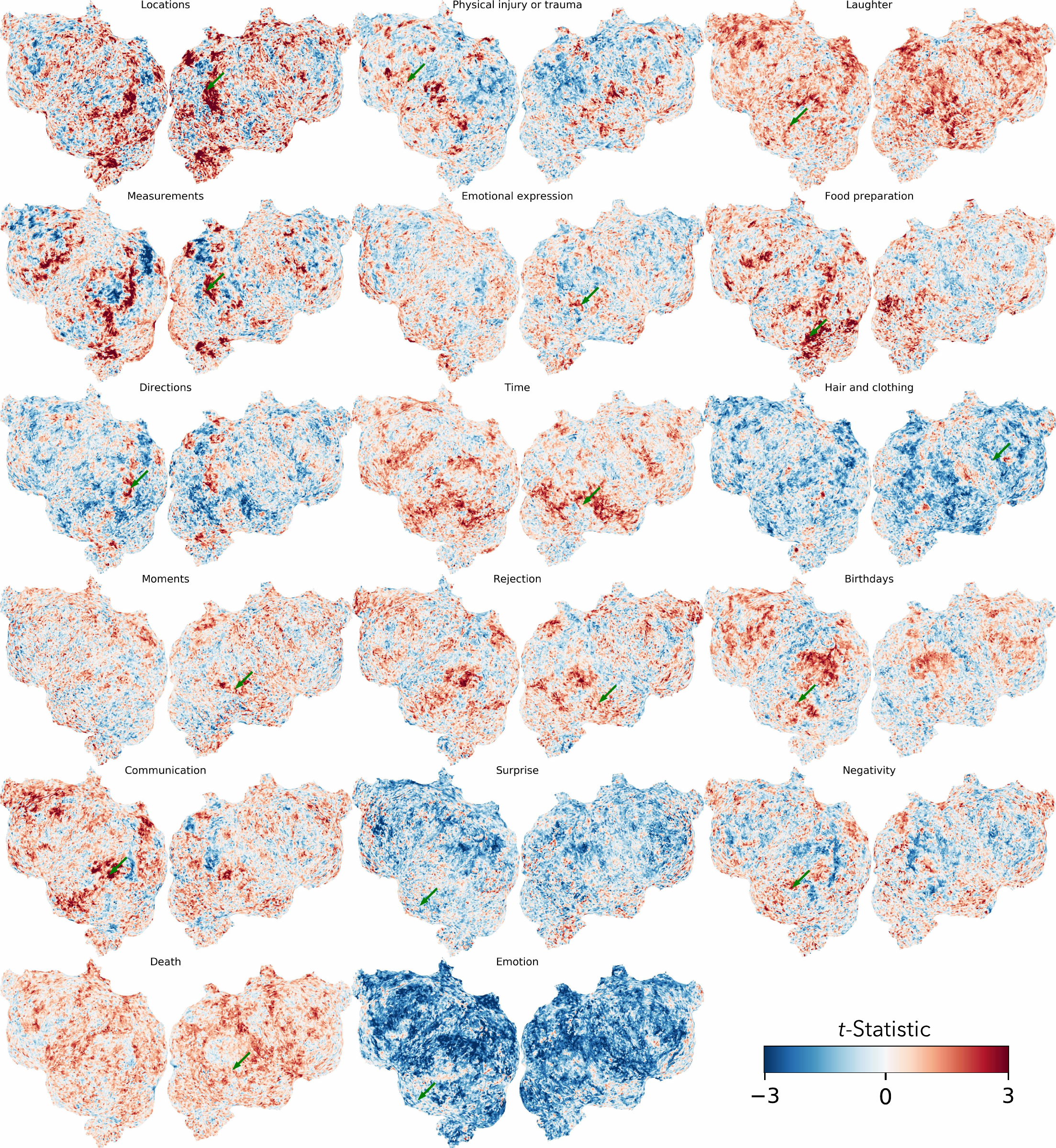}
    \caption{Explanation contrast for all tested explanations in S02. Arrows indicate source voxel where driving was attempted.}
    \label{fig:physical_movement}
\end{figure}

\begin{figure}[H]
    \centering
    \includegraphics[width=1\textwidth]{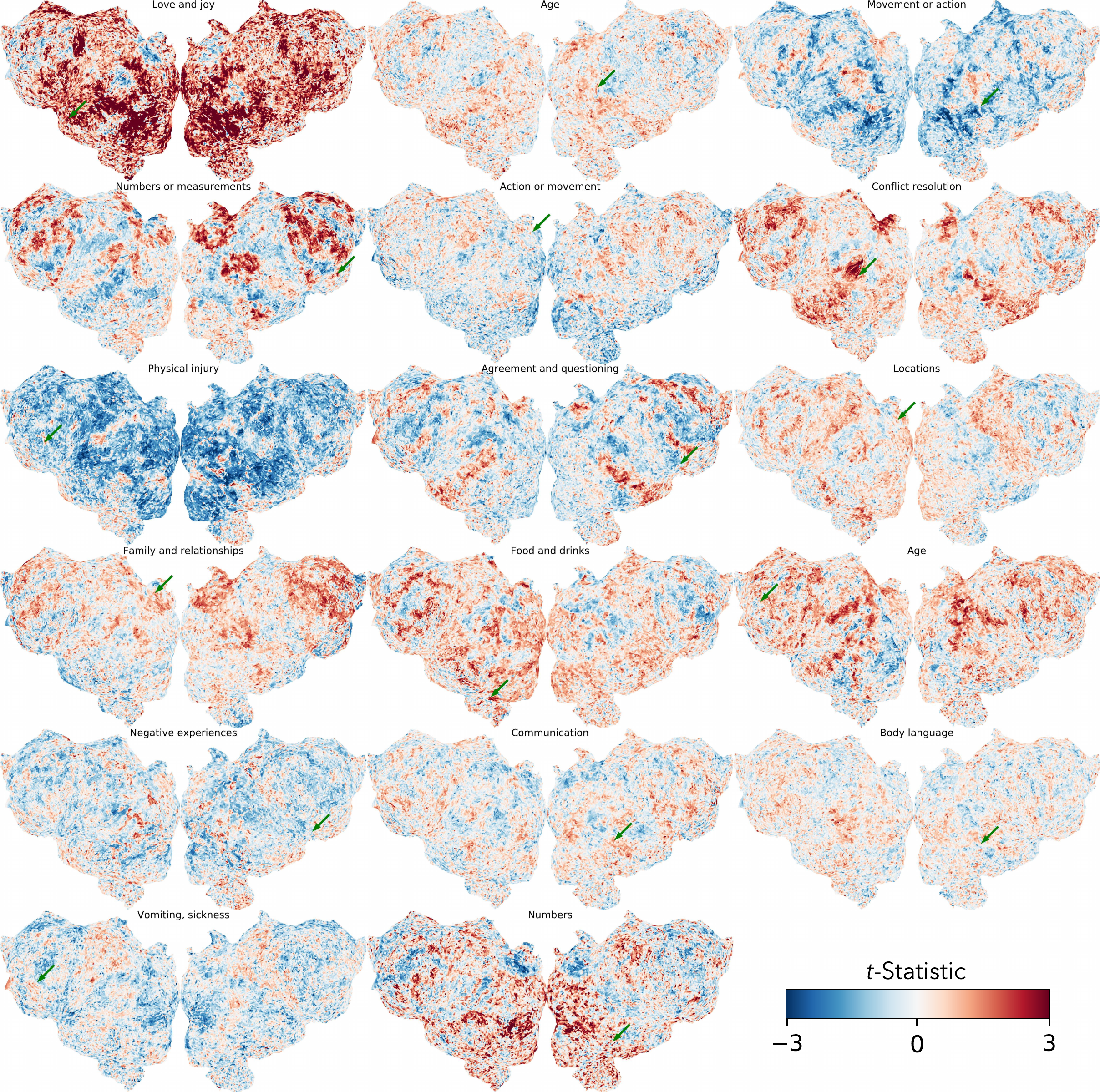}
    \caption{Explanation contrast for all tested explanations in S03. Arrows indicate source voxel where driving was attempted.}
    \label{fig:rejection}
\end{figure}

\FloatBarrier
\subsection{Encoding model performance}

\begin{figure}
    \centering
    \begin{tabular}{lccc}
         & S01 & S02 & S03 \\
         \rotatebox{90}{LLaMA} & \includegraphics[width=0.3\textwidth]{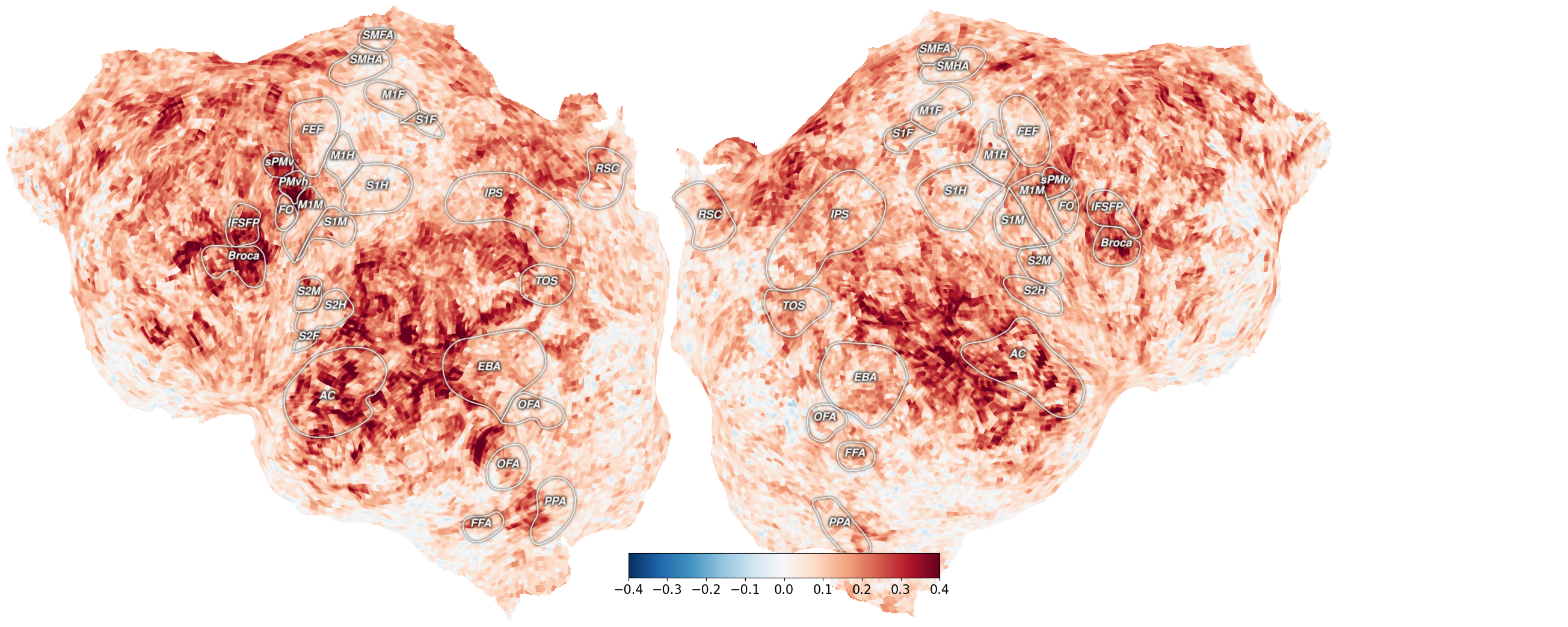} & \includegraphics[width=0.26\textwidth]{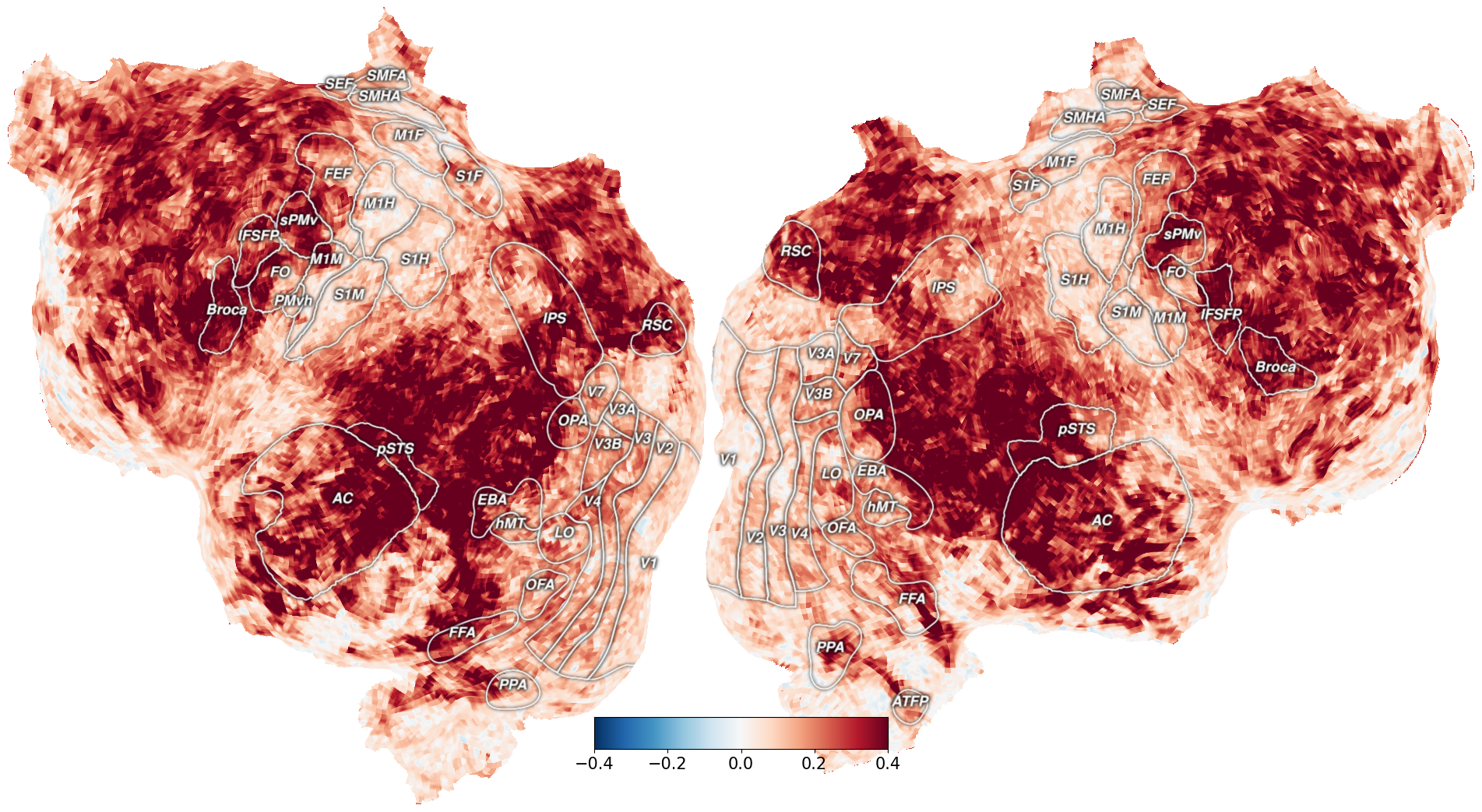} & \includegraphics[width=0.26\textwidth]{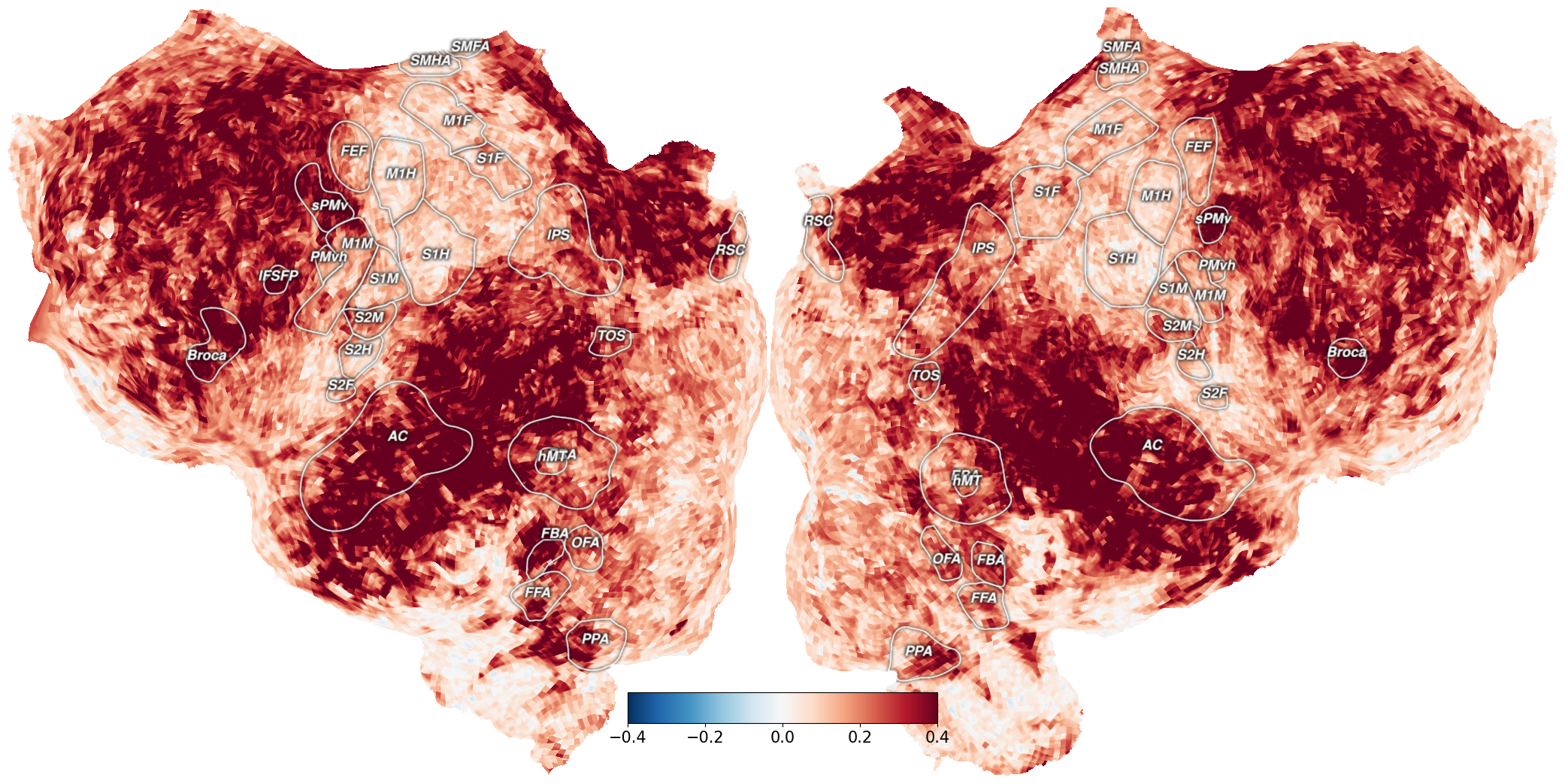} \\
         \rotatebox{90}{OPT} & \includegraphics[width=0.3\textwidth]{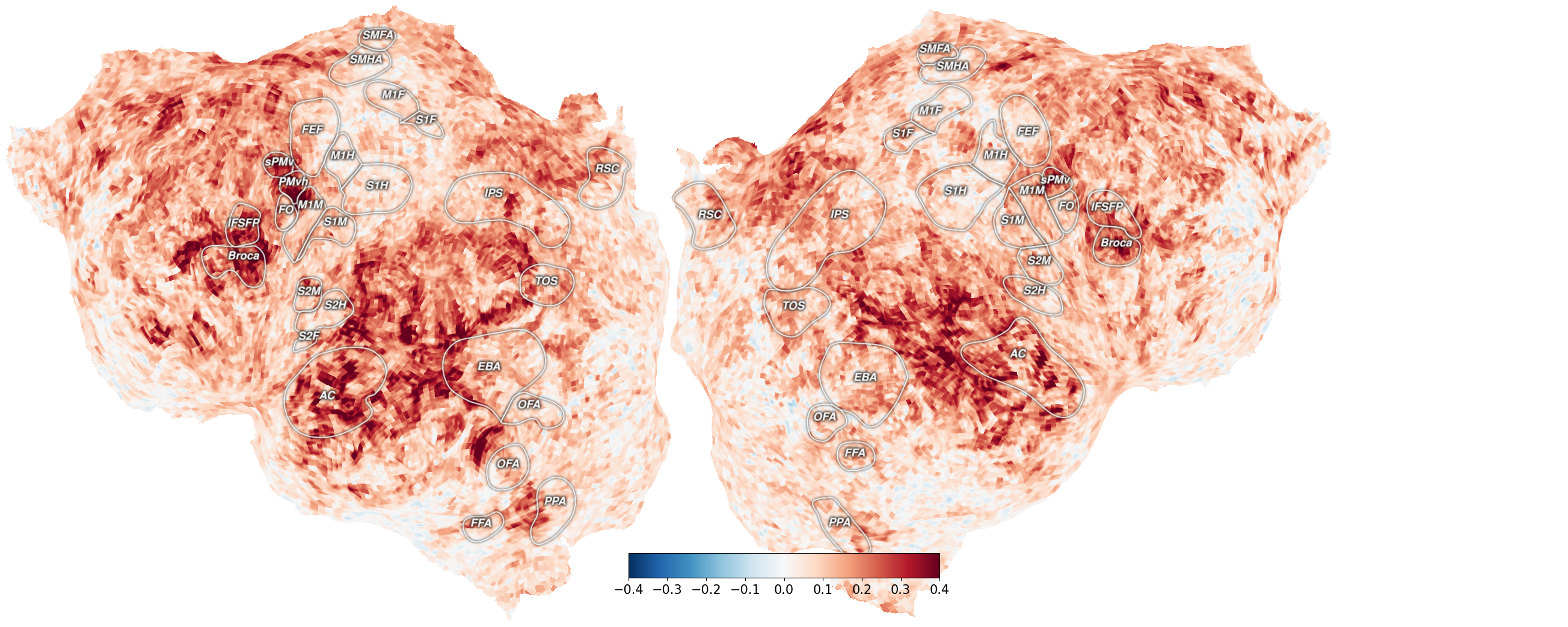} & \includegraphics[width=0.26\textwidth]{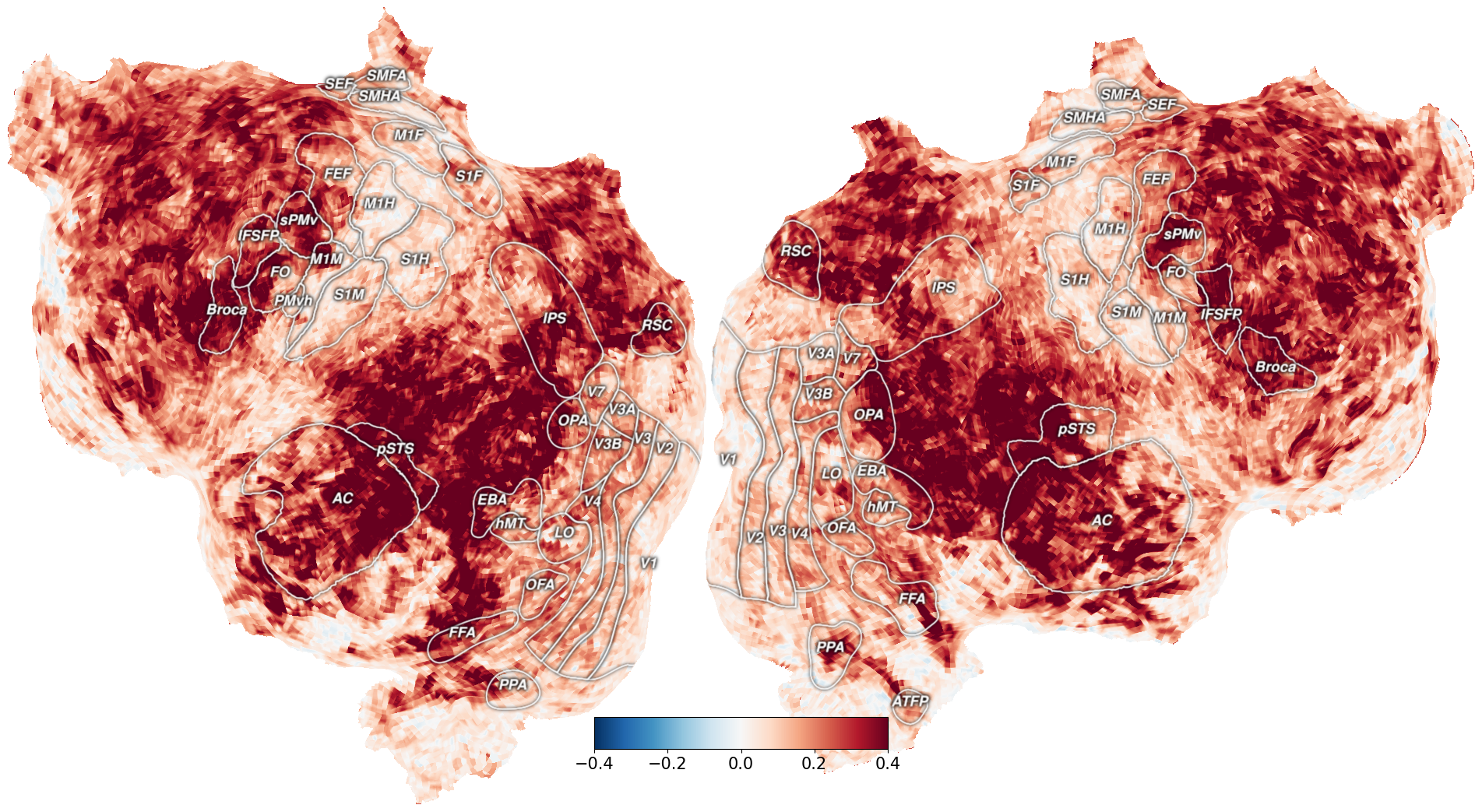} & \includegraphics[width=0.26\textwidth]{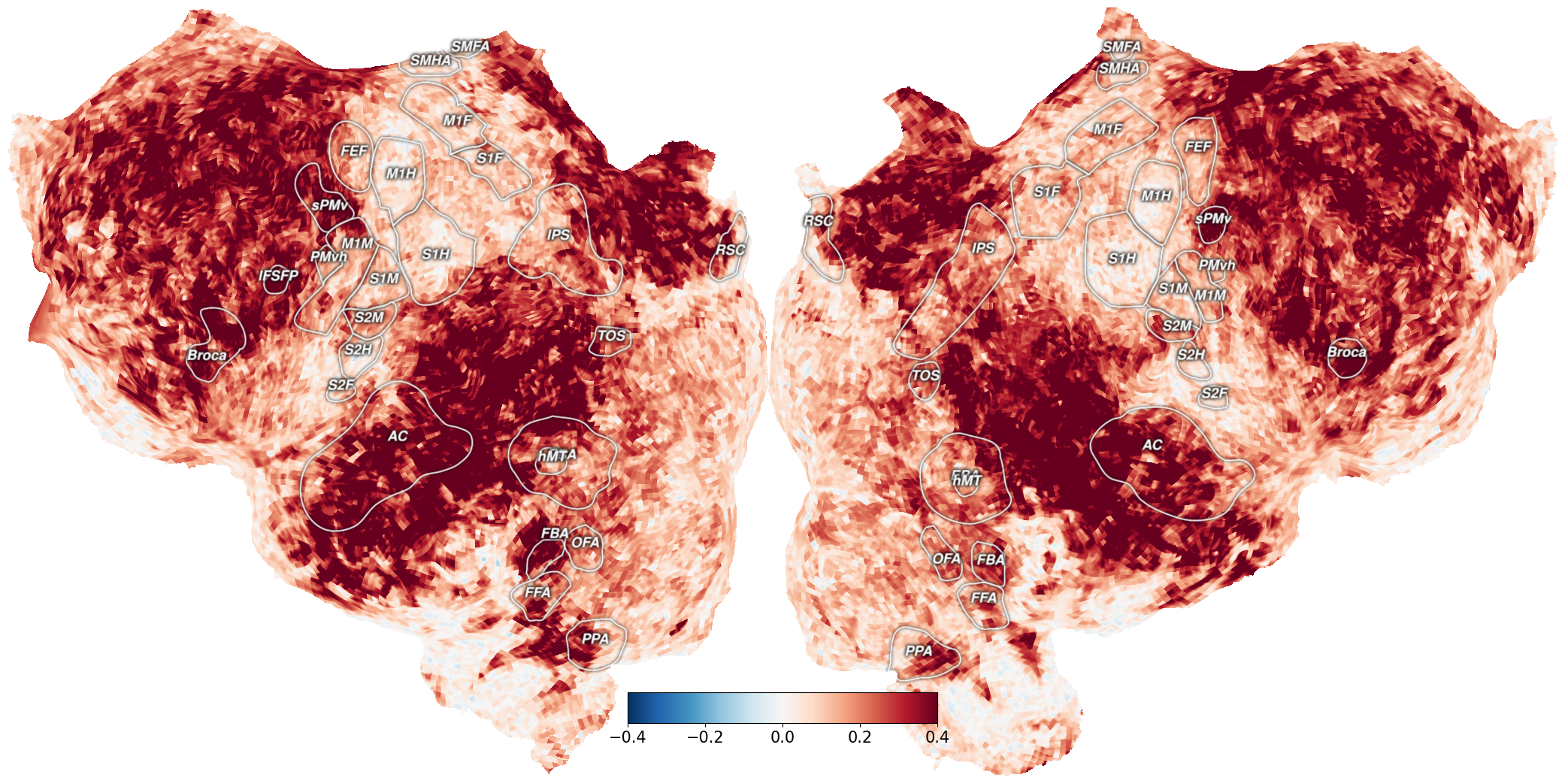} \\
         \rotatebox{90}{Word-rate} & \includegraphics[width=0.3\textwidth]{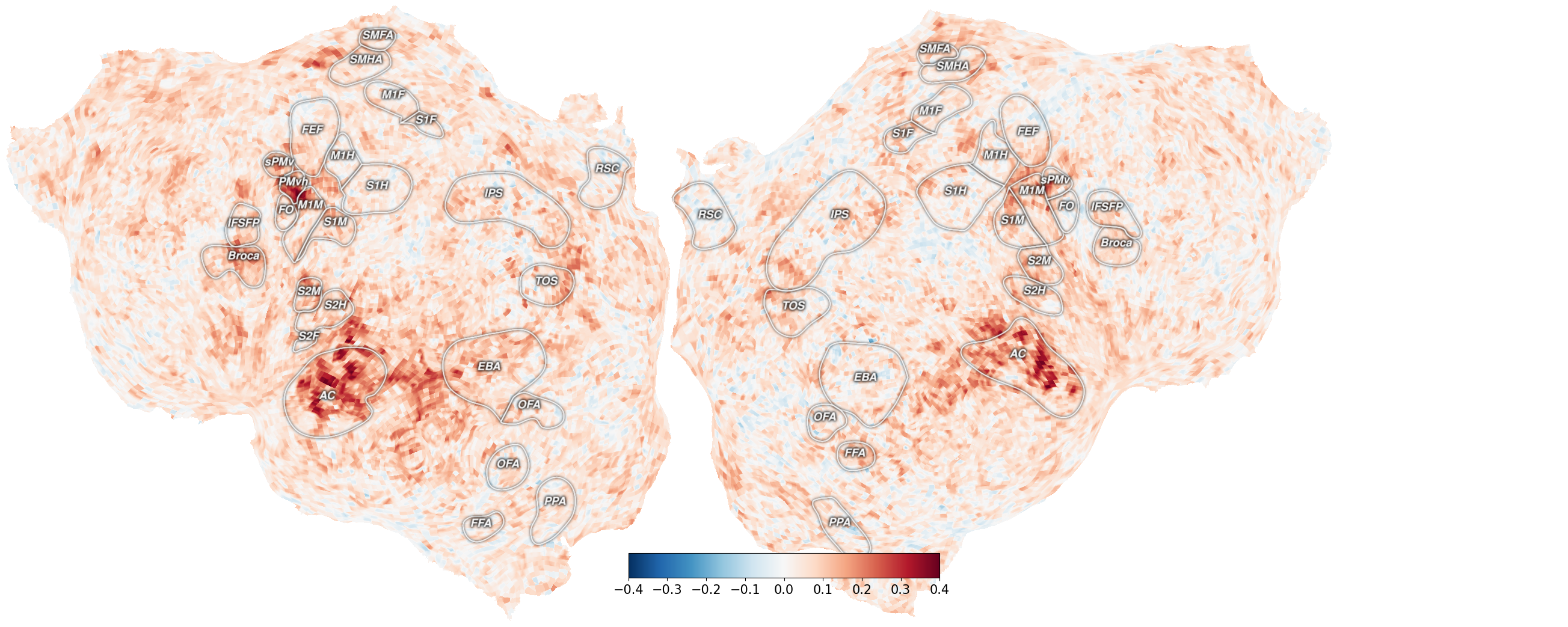} & \includegraphics[width=0.26\textwidth]{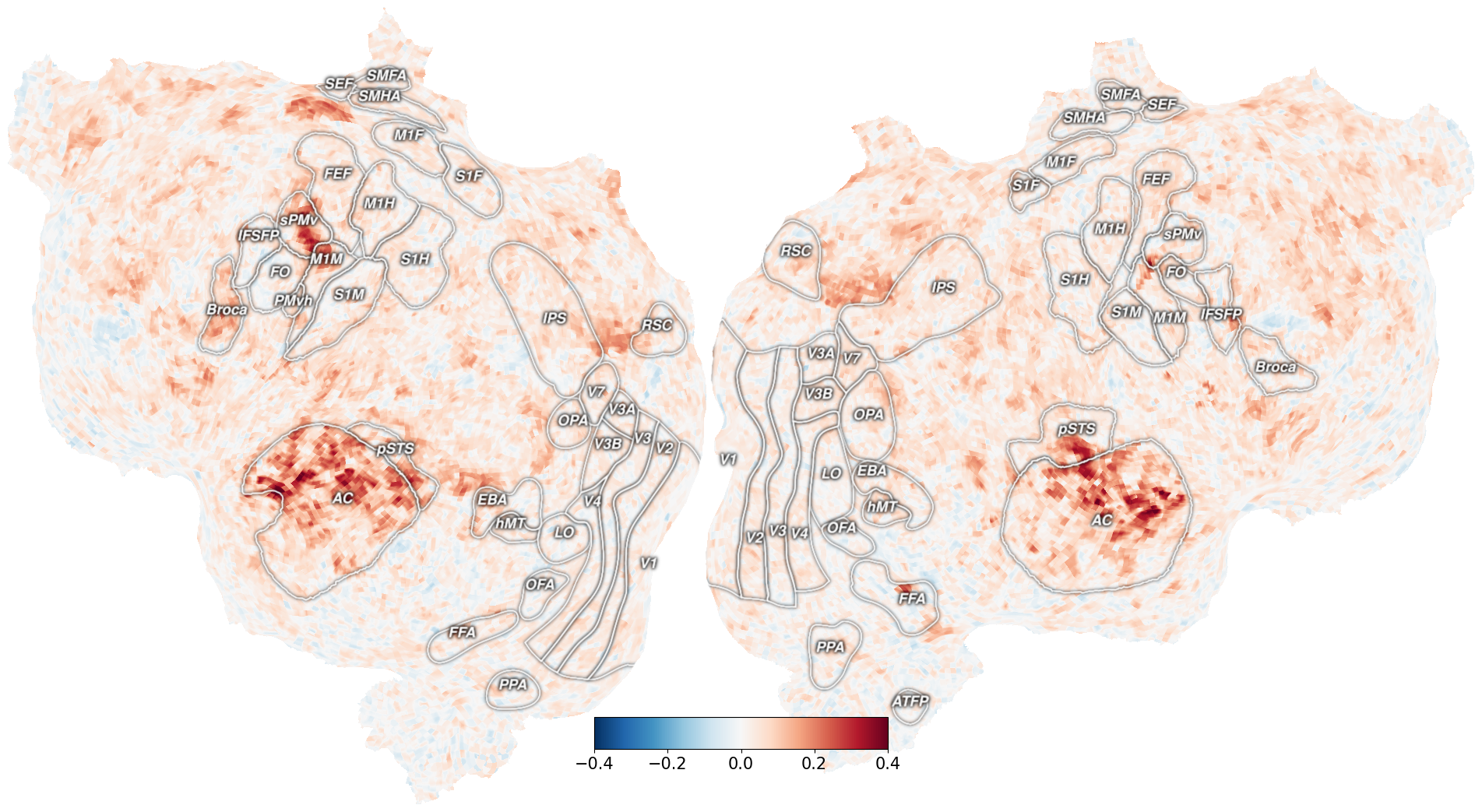} & \includegraphics[width=0.26\textwidth]{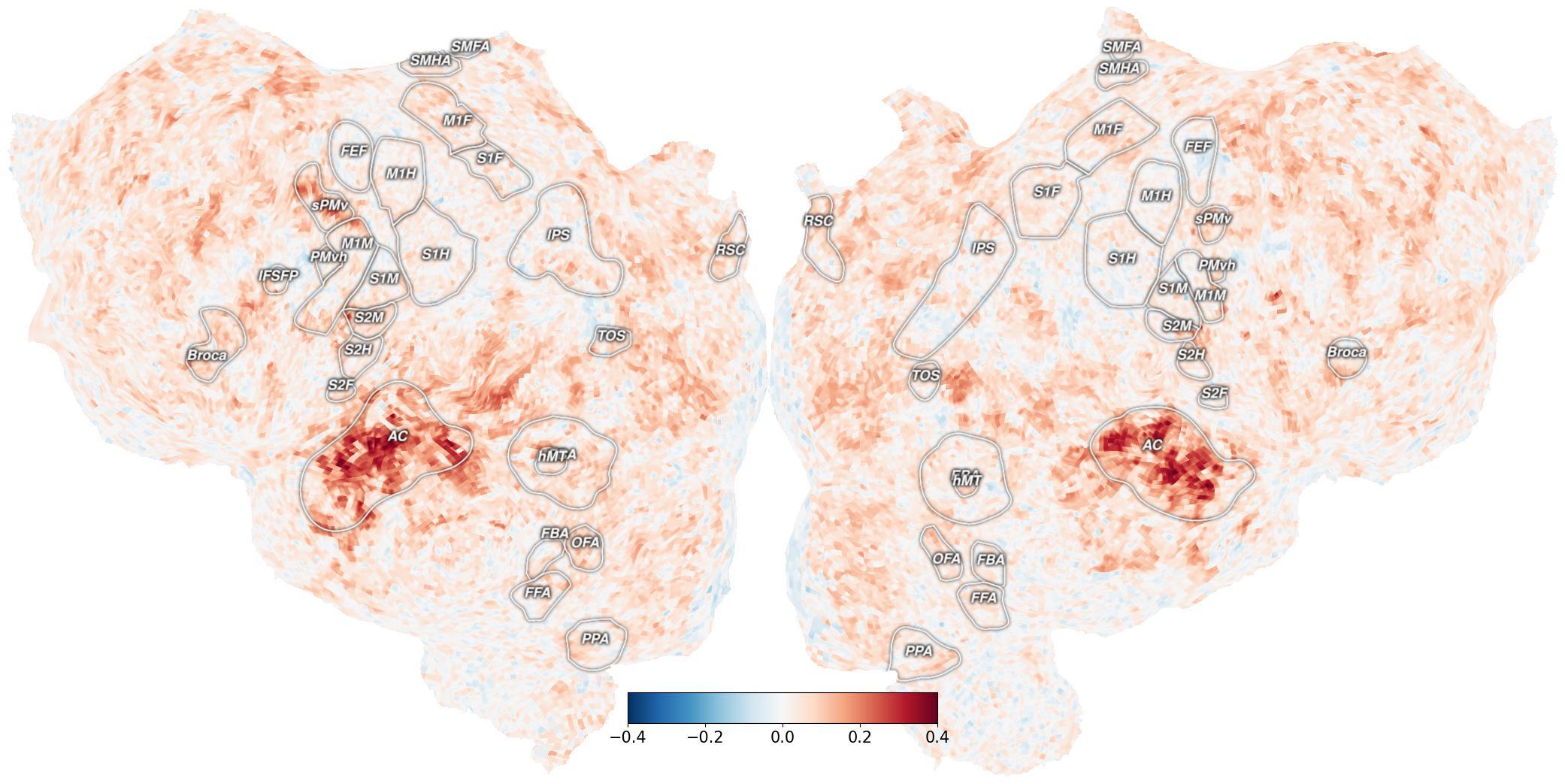} \\
    \end{tabular}
    
    \caption{Voxelwise prediction performance (test correlation) for the two encoding models measured here on naturalistic stories, along with a word-rate baseline.}
    \label{fig:enc_performance}
\end{figure}

\begin{table}[]
    \centering
    \caption{Prediction performance (test correlation) for the two encoding models used here, along with a word-rate baseline.}
    \begin{tabular}{lrrr|r}
    \toprule
    & S01 & S02 & S03 & AVG\\
    \midrule
    LLaMA & 0.08 & 0.15 & 0.17 & 0.13 \\
    OPT & 0.08 & 0.14 & 0.16 & 0.13\\
    Word-rate & 0.04 & 0.03 & 0.03 & 0.03 \\
    \bottomrule
    \end{tabular}
    \label{tab:my_label}
\end{table}

\FloatBarrier

\subsection{Temporal Bleed effect}

\begin{figure}[H]
    \centering
    
    \includegraphics[width=0.5\textwidth]{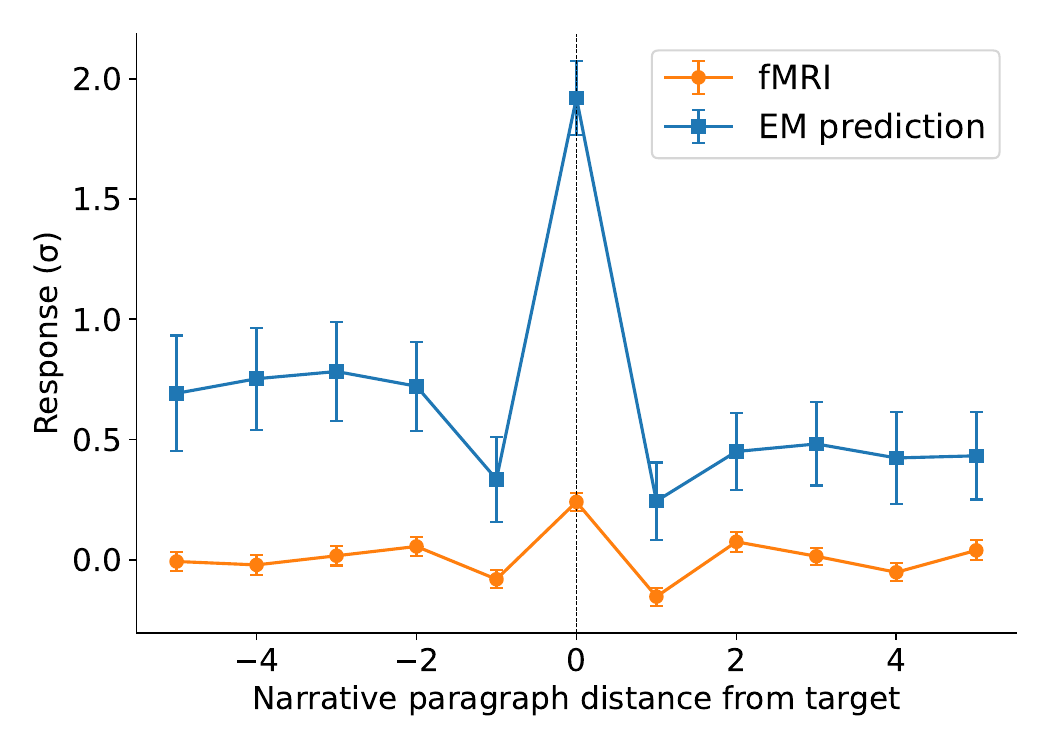}
    \caption{Measurement of driving patterns in adjacent paragraphs. We observe no "temporal bleed" effect where adjacent paragraphs drive response in their corresponding voxel, in fact, there appears to a slight suppressive effect ($-0.119\sigma$). Driving effects largely appear to be temporally localized within paragraph boundaries. Result shown for UTS02. Bars show standard error.}
    \label{fig:tembleed}
\end{figure}

\FloatBarrier

\subsection{Distributional Analysis}

\label{app:distribution}
\begin{figure}[H]
    \centering
    
    \includegraphics[width=1\textwidth]{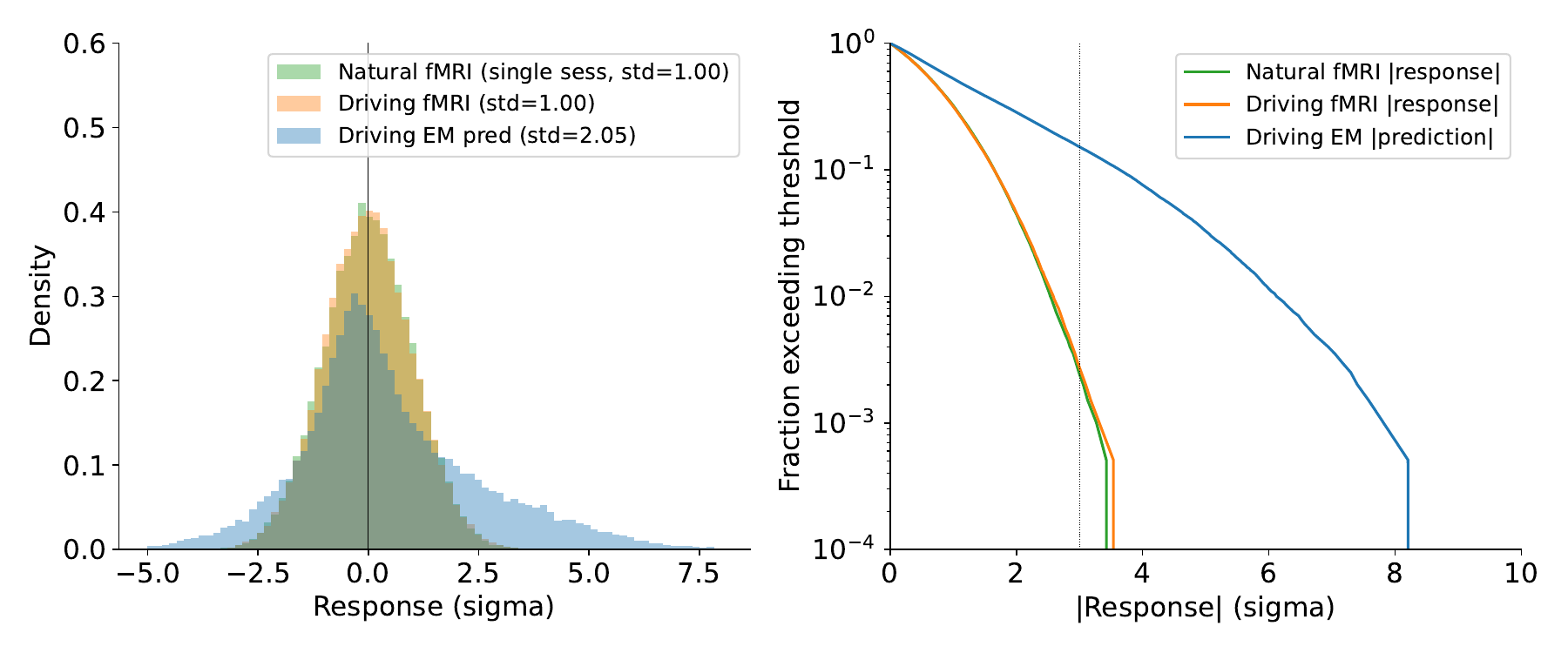}
    \caption{Distributional analysis of encoding model predictions and driving/naturalistic fMRI ground truth. We find that the encoding model Result shown for UTS02 voxels. \textit{Left:} Distribution of responses on ground truth data in both naturalistic and driving stories as well as the predictions for the driving stories. We see that the encoding model systematically overestimates the driving value. \textit{Right:} Survival function for the same distributions. The tail is much longer for driving predictions, justifying the clipping approach in our driving ceiling analysis}
    \label{fig:tembleed}
\end{figure}





\end{appendices}

\end{document}